\documentclass[preprint,12pt]{elsarticle}
\usepackage{lineno,hyperref}
\usepackage{graphics}
\usepackage{graphicx}
\usepackage{epsfig}
\usepackage{float}
\usepackage{multirow}
\usepackage{multicol}
\usepackage{times}
\usepackage{amsmath}
\usepackage{amssymb}
\usepackage{algorithm}
\usepackage{algorithmic}
\usepackage{subcaption}
\usepackage{rotating}
\usepackage{mathrsfs}
\usepackage{threeparttable}
\usepackage{epstopdf}
\usepackage{amssymb}
\usepackage{makecell}
\usepackage[shortcuts]{extdash}
\usepackage{color}
\usepackage{diagbox}
\usepackage{tabu}
\biboptions{numbers,sort&compress}
\begin{document}
\captionsetup[figure]{labelfont={bf},labelformat={default},labelsep=period,name={Fig.}}

\begin{frontmatter}

\title{PaVa: a novel Path-based Valley-seeking clustering algorithm}

\author[1]{Lin Ma}
\address[1]{School of Statistics, Southwestern University of Finance and Economics, China}
\ead{malin@smail.swufe.edu.cn}

\author[2]{Conan Liu}
\address[2]{UNSW Business School, University of New South Wales, Australia}
\ead{conan.liu88@gmail.com}

\author[1]{Tiefeng Ma*}
\ead{matiefeng@swufe.edu.cn}

\author[3]{Shuangzhe Liu}
\address[3]{Faculty of Science and Technology, University of Canberra, Australia}
\ead{shuangzhe.liu@canberra.edu.au}

\begin{abstract}
Clustering methods are being applied to a wider range of scenarios involving more complex datasets, where the shapes of clusters tend to be arbitrary. In this paper, we propose a novel Path-based Valley-seeking clustering algorithm for arbitrarily shaped clusters. This work aims to seek the valleys among clusters and then individually extract clusters. Three vital techniques are used in this algorithm.
First, path distance (minmax distance) is employed to transform the irregular boundaries among clusters, that is density valleys, into perfect spherical shells.
Second, a suitable density measurement, $k$-distance, is employed to make adjustment on Minimum Spanning Tree, by which a robust minmax distance is calculated.
Third, we seek the transformed density valleys by determining their centers and radius.
% Third, the probability density function of distances is established, on which the location of the first valley is set to be radius of the spherical shell.
Based on the vital techniques, the main contributions of the proposed algorithm can be summarized as follows.
First, the clusters are wrapped in spherical shells after the distance transformation, making the extraction process efficient even with clusters of arbitrary shape.
Second, adjusted Minimum Spanning Tree enhances the robustness of minmax distance under different kinds of noise.
Last, the number of clusters does not need to be inputted or decided manually due to the individual extraction process.
After applying the proposed algorithm to several commonly used synthetic datasets, the results indicate that the Path-based Valley-seeking algorithm is accurate and efficient. The algorithm is based on the dissimilarity of objects, so it can be applied to a wide range of fields. Its performance on real-world datasets illustrates its versatility.

\end{abstract}


\end{frontmatter}

%\linenumbers

\section{Introduction}
Clustering algorithms are a useful tool when exploring latent data structures, which aim to partition a group of objects into several subgroups to achieve both high intragroup and low intergroup similarities without knowing any prior information. With the rapid development of information technology, the clustering methods have been applied to a broader range of scenarios involving more complex datasets \cite{ezugwu2022comprehensive, satyanarayana2022identifying, chen2022application, sharma2022eeffa, haseeb2022autoencoder}, in which the clusters tend to be arbitrarily shaped and surrounded by noise. To this end, researchers have been  motivated to design robust clustering methods that perform well on datasets composed of arbitrarily shaped clusters. Among all the clustering methods for arbitrarily shaped clusters, the graph-based methods and density-based methods have been well-studied.

In the graph-based clustering methods, the dataset is first mapped to a graph structure, with objects corresponding to vertices (nodes) and their relationships corresponding to edges. Then some typical techniques are operated on the graph, such as spectral techniques \cite{ng2001spectral, kang2020multi}, graph-cut operations \cite{li2020novel, ma2021multi} and Minimum Spanning Tree (MST) techniques \cite{lv2018ccimst, liu2019global}. These techniques are employed either independently or in combination. On the one hand, the core operation of the spectral techniques is performing Laplacian transformations on a weighted graph and then executing spectral decomposition (SD). The spectral techniques are time-consuming, so it is employed when the size of the dataset is small. On the other hand, graph-cut operations are used to partition the graph into subgraphs according to the inconsistency of edges. However, the inconsistent edges in graph-cut operations are not well defined in different patterns of the dataset.
In addition to spectral techniques and graph cutting operations, the usage of MST in clustering tasks is more practical and diverse. It captures the non-spherical features of data through a tree-type connection structure of objects. However, such connection structures are significantly affected or even ruined by noise. Therefore, self-adaptation operations are necessary in the presence of different types of noise.

\begin{figure}
  \centering
 \subcaptionbox{}{\includegraphics[width=6.5cm]{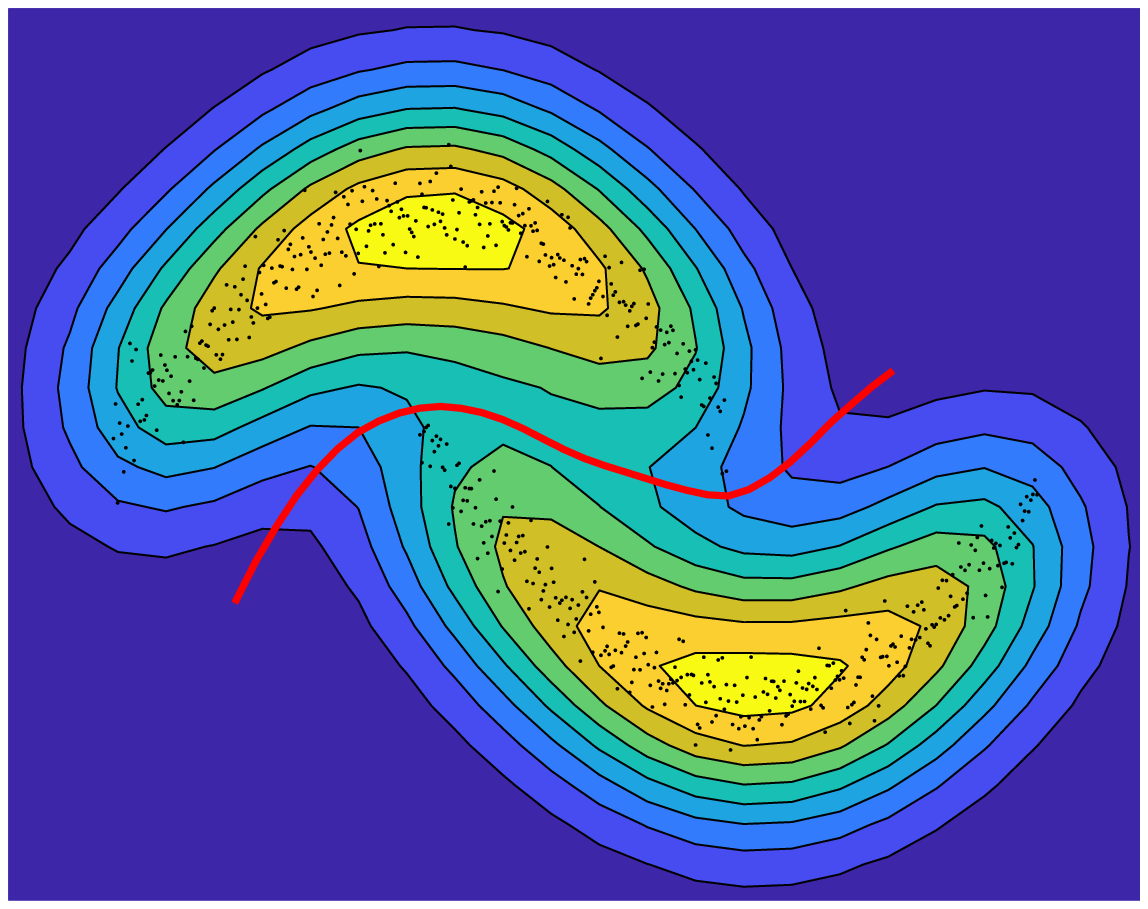}}
 \subcaptionbox{}{\includegraphics[width=6.5cm]{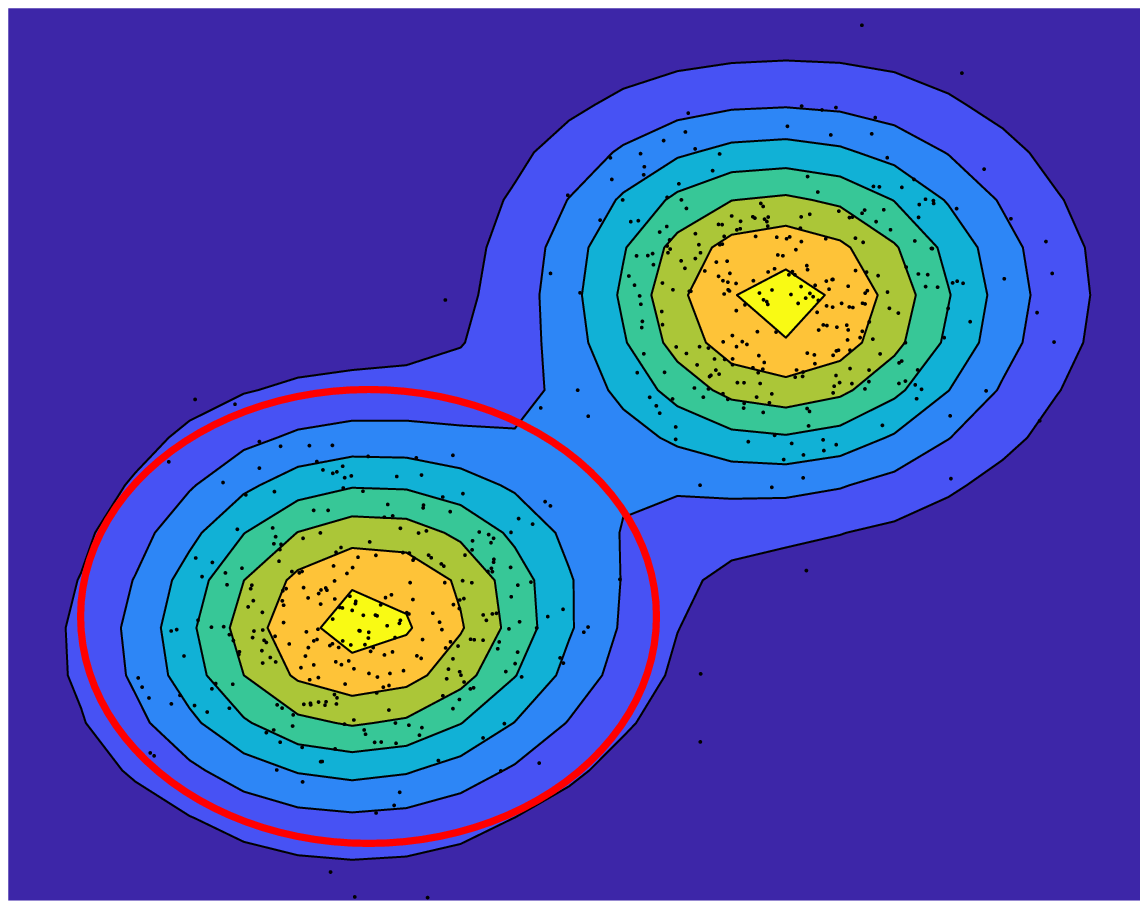}}
  \caption{\small{(a) Valley in dataset \emph{twomoons}. (b) Valley in spherical dataset. Spherical boundary can be uniquely ascertained by a center and a radius, which is much simpler than an irregular boundary. We aim to convert the irregular boundaries to spherical boundaries or hypersphere shells.}}
  \label{figure:valleys}
\end{figure}

Besides the graph-based clustering methods, the density-based clustering methods have the inherent ability to discover arbitrarily shaped clusters and handle noise, such as Density-Based Spatial Clustering Algorithm with Noise (DBSCAN) \cite{ester1996density}, Ordering Points To Identify the Clustering Structure (OPTICS) \cite{ankerst1999optics}, and Mean shift \cite{cheng1995mean}. These algorithms have been widely used and their variants are proposed massively. The above density-based clustering methods are all based on the framework of cluster-expansion. Most of them share the common limitations of sensitivity to parameters and high time complexity.  On the contrary to finding the dense area and local density peaks, the valley-seeking techniques find the density valleys \cite{koontz1976graph}. In this way, clusters can be extracted as long as their boundaries are known. Most of the valley-seeking methods explore the boundaries of the original clusters \cite{zhang2007neighbor}, whether regular or irregular. This rough operation is also time-consuming and not robust. The Density-Based Radar Scanning clustering algorithm (DBRS) \cite{ma2022new} approximates the density valleys between clusters with spherical shells, whose centers and radii are determined efficiently using a data-driven approach. However, DBRS is only capable when the clusters are round in shape.

Based on this review of graph-based clustering algorithms and density-based clustering algorithms, we note there exists an opportunity to develop a hybrid algorithm of graph-based clustering and density-based clustering methods, synthesizing the advantages of both. Motivated by the fact that the irregular boundaries can be transformed to hypersphere shells by applying minmax distance, we first represent the spatial relationship between objects with minmax distance. However, the existence of noise negatively impacts the transformation, which blurs the boundaries between clusters. Thus, we utilize $k$-distance to refine and adjust the MST, by which minmax distance is calculated. After this minmax distance-based transformation, we apply a similar pipeline with DBRS to extract clusters individually by seeking the spherical density valleys among clusters. This operation is much more efficient than handling the arbitrarily shaped clusters directly. As in Fig. \ref{figure:valleys}, a spherical density valley can be uniquely determined by a center and a radius, while modelling the irregular density valley is a tough task.

In this paper, we propose a novel Path-based Valley-seeking (PaVa) clustering algorithm for arbitrarily shaped clusters. Our main contributions can be summarized as follows.

  (i) Minmax distance substitutes Euclidean distance as a dissimilarity measurement to convert the irregular valleys among arbitrarily shaped clusters to spherical boundaries, which can be uniquely ascertained by a center and corresponding radius.

  (ii) The center and radius are determined by a data-driven manner instead of fixed locations and values, which improves the flexibility to handle heterogeneous clusters and enhance the robustness to parameters.

  (iii) Given that minmax distance calculated from MST is not robust to noise, we adjust the weights of edges in MST with the density of objects, $k$-distance. The PaVa clustering algorithm is then shown to perform well on datasets with different types of noise.

  (iv) Due to the individual extraction process, the number of clusters does not need to be given by the user.
  Moreover the only parameter $k$, the number of neighbors involved to calculate the $k$-distance, can be set to $\log(N)$, where $N$ is the size of the dataset. Therefore, the proposal is very accessible to non-expert users.

The rest of the paper is organized as follows. In Section \ref{section_related_work}, we introduce related work, that is, graph-based clustering methods and density-based clustering methods for arbitrarily shaped clusters. In Section \ref{section_preliminaries} and \ref{section_algorithm}, we present some preliminaries and describe the PaVa clustering algorithm. Then, in Section \ref{section_experiment}, we use synthetic and real-world datasets to evaluate our algorithm's performance and compare it with some existing methods. Finally, we provide some concluding remarks on the obtained results and provide ideas for possible future work in Section \ref{section_conclusion}.

\section{Related Work} \label{section_related_work}

In this section, we introduce the clustering methods from two perspectives, graph-based clustering methods and density-based clustering methods, which are closely related to our proposal.

\subsection{Graph-based clustering methods}

In the graph-based clustering methods, what is prioritized is mapping the dataset to its corresponding graph. Based on the constructed graph structure, typical techniques such as graph-cut operations, spectral, or MST-based techniques, are employed either independently or in combination.

Spectral graph theory is a common tool to separate arbitrarily shaped clusters. After using spectral characteristics of the Laplacian of a weighted graph, the vertices corresponding to objects are partitioned into natural clusters according to some post-clustering process. For instance, k-means is commonly applied to the top-$k$ eigenvectors as a post-clustering process, that is, spectral k-means \cite{ng2001spectral} and its similar algorithm, spectral weighted k-means, perform a row-segmentation scheme \cite{liu2017spectral}. It is proved to have the same mathematical foundations with kernel k-means \cite{dhillon2004kernel}; that is because the core of both approaches lies in their efforts to construct adjacency structures between objects.

Graph-cut operations are used to partition the graph into subgraphs according to the inconsistency of edges. For example, an inconsistent edge can be confirmed by the ratio between its edge weight and the average of other nearby edge weights \cite{zahn1971graph}. In this way, the edges corresponding to the large dissimilarities between objects are recognized and set to be breakpoints between two arbitrarily shaped clusters.

The MST techniques are more practical and diverse than the spectral techniques and graph-cut operations in clustering tasks. For instance, they can be used to produce a clustering validation index \cite{xie2020new}, guidance for split-and-merge processes \cite{zhong2011minimum} or as a tool for density measurement \cite{Mishra2022RDMN}. More importantly, they serve as a basis of transformation from Euclidean to Non-Euclidean spaces. For example, the length of the shortest path between two vertices is defined as the geodesic distance, which represents the cost of `walking' from one position to another \cite{gattone2018geodesic, du2018density}. Additionally, minmax distance (explained in Concept \ref{definition:minmax distance} in Section \ref{section_preliminaries}) shows the conductive similarity between two objects. After applying minmax distance, objects belonging to the same cluster are close together, while the objects belonging to different clusters are far apart, even when the clusters are of arbitrary shapes. This transformation can also explain the ability of hierarchical clustering methods to handle arbitrarily shaped clusters; if we use the single-linkage in the hierarchical clustering methods, the subclusters would gradually merge into elongated structures and clusters with arbitrary shapes would be found. Different as they are, the single-linkage and minmax distance can be explained with the same philosophy, namely, Minimum Spanning Tree (MST). The single-linkage hierarchical clustering method is also called clustering of maximum-spacing, and its procedure is precisely Kruskal's Minimum Spanning Tree algorithm. More precisely, the single-linkage is running Kruskal's algorithm but stopping it right before it adds its last $k$-$1$ edges \cite{kleinberg2006algorithm}. Furthermore, a range of works have indicated the power of the MST-based methods in the context of detecting and describing inherent clustering structures with arbitrary shapes \cite{zahn1971graph, ma2021multi}. For example, MST techniques are embedded into existing clustering methods such as density-based clustering methods \cite{cheng2019clustering} \cite{qiu2022fast} and partition-based clustering methods \cite{liu2019global}.

Although the MST-based methods naturally handle clustering problems with arbitrary shapes, they are not robust to noise. The impact of uniformly distributed outliers can be relieved by finding and deleting inconsistent edges, while the trickiest condition is the connection structure between clusters, namely "neck" or "bridge", which is to blame for failing to recognize two adjacent clusters separately. Considering the low-density feature of noise and connection structures, some literature have employed the density of vertices (objects) in the MST to reduce the impact of noise. In \cite{zahn1971graph, hwang2019optimized}, the neck is defined to be a small localized subset of the objects whose deletion generates inconsistent edges between clusters. After recognizing and removing the neck, the inconsistent edges are removed from the MST to separate the clusters. Additionally, a two-step robust MST-based clustering algorithm is proposed in \cite{liu2018robust}. In the first phase, connected objects are separated by applying a density-based coarsening process. Based on the intermediate gain in the first phase, a greedy method is presented to get the final result using minmax distance calculated by MST. Besides, the robustness to noise is improved by a symmetry-favored kNN accroding to the sparsity of noise area, which increases the reliability of results  \cite{khan2022fast}. Another approach is to adjust the weight of the MST using the M-estimator as a density measure. In this way, a robust path-based similarity measure is established \cite{ chang2008robust}.

%Similarly, in this paper we use $k$-distance as density measurement to adjust MST for the reasons mentioned in the previous section.

\subsection{Density-based clustering methods}

The density-based clustering methods are generic for its ability to detect arbitrarily shaped clusters
and determine the number of clusters while filtering out noise. The majority of density-based clustering algorithms aim to find clusters located in dense regions. Moreover, the valley-seeking technique is used to act oppositely.

The density based clustering methods, such as DBSCAN, Mean shift and Density peak clustering (DPC)  \cite{rodriguez2014clustering} are intuitive in finding dense regions. DBSCAN has been extensively used, but it is parameter-sensitive and time-consuming. In order to address the above limitations, various improved variants of DBSCAN have been proposed. For example, the parameter $eps$ is adapted for every object instead of being a fixed value \cite{ZHANG2021344}. For efficiency, an operational set is established for DBSCAN to decrease the time for computing density \cite{hanafi2022fast}. The Mean shift algorithm detects clusters using a hill-climbing strategy. However, the shifting process does not work perfectly, especially when the density measurement is not suitable. Thus, a refined weighted Mean shift \cite{chakraborty2021automated} and more robust density measurement \cite{cariou2022novel} is employed in one of its variants. Moreover, DPC adopts a subtle strategy for determining cluster centers. Many techniques, such as adaptive core fusion \cite{fang2020adaptive} and KNN \cite{chen2020fast}, are used to refine its assign process for arbitrarily shaped clusters.

Contrary to the above mentioned methods, valley-seeking techniques explore the regions with less data \cite{gokcay2002information}. For example, a modified valley-seeking algorithm has been used to adaptively choose the outlier thresholds for generating the final clusters \cite{laohakiat2021incremental}. In general, objects are iteratively moved toward the path of maximum gradient and assigned to its nearest mode \cite{koontz1976graph}. The estimation of the density gradient is the most important factor, on which most of the valley-seeking algorithms are based. Therefore, the majority of valley-seeking algorithms study the estimation of the density gradient. In \cite{laohakiat2021incremental}, the variant is supposed to follow a Gaussian distribution, where the area outside two standard deviations is set to be a valley. Moreover, the nonparametric methodologies such as Parzen Window Method (PWM) take full advantage of the properties of the Gaussian kernel to estimate the density gradient \cite{parzen1962estimation}. To overcome the oscillation and over-smoothness of the kernel estimation, Cluster Intensity Function (CIF), an analog of the density function, has been introduced to capture the cluster centers and boundaries \cite{yip2006dynamic}. Moreover, a clustering evaluation function (CEF) \cite{zhang2007neighbor} and Normalized Density Derivative (NDD) \cite{fukunaga2013introduction} are proposed to measure the divergence as another measurement of density gradient. In addition to exploiting the measurements of the density gradient, a graph-theoretic approach is presented in \cite{koontz1976graph} and confirmed to behave as a valley-seeking algorithm to avoid the iterative operation.

%In this paper, we represent the spatial relationship between objects in terms of the aforementioned minmax distance before the valley-seeking process, which ensures that each cluster can be wrapped in a spherical hull without containing objects belonging to different clusters. This operation eliminates the estimation of density estimation, thus it is efficient and robust.

\section{Preliminaries} \label{section_preliminaries}
In this section, we present two classical concepts, minmax distance and $k$-distance. On the one hand, minmax distance is critical to the PaVa clustering algorithm. It shows the conductive similarity between objects, which is necessary when the clusters are non-convex. On the other hand, noise has great impact on the MST, so we employ the $k$-distance to enhance the stability of MST.

\newtheorem{definition}{Concept}
\begin{definition}[minmax distance \cite{zahn1971graph}]
Suppose $X=\{x_1, x_2, .., x_N\}$ is a dataset with $N$ objects. Its corresponding graph $G=(V, E)$ is constructed by mapping its objects to vertices $V=\{v_1, v_2, ..., v_N\}$ and relationships between objects to edges $E=\{e_1, e_2, ..., e_{N_E}\}$.
In graph $G$, let $P_{v_o, v_i}$ denote the set of all paths from vertex $v_o$ to vertex
$v_i$. For each path $p \in P_{v_o, v_i}$, if the edge weights are calculated by a dissimilarity measure (e.g., Euclidean distance), then the effective distance with respect to $p$, $d^p(v_o, v_i)$, is the maximum edge weight along the path $p$, i.e.,

\begin{equation}
d^p (v_o, v_i)=max_{1 \leq h < |p|} weight(p[h], p[h+1]).
\end{equation}
where $|p|$ is the length of path $p$, $p[h]$ denotes the $h$th vertex along path $p$ and $weight(p[h], p[h+1])$
denotes the edge weight between $p[h]$ and $p[h+1]$. The minmax
distance $d^{mm}(v_o, v_i)$ is the minimum $d^p(v_o, v_i)$ of all paths, i.e.,

\begin{equation}
d^{mm}(v_o, v_i)=min_{p \in P_{v_o, v_i}} d^p(v_o, v_i).
\end{equation}
\label{definition:minmax distance}
\end{definition}

\begin{figure}
  \centering
 \subcaptionbox{}{\includegraphics[width=6.5cm]{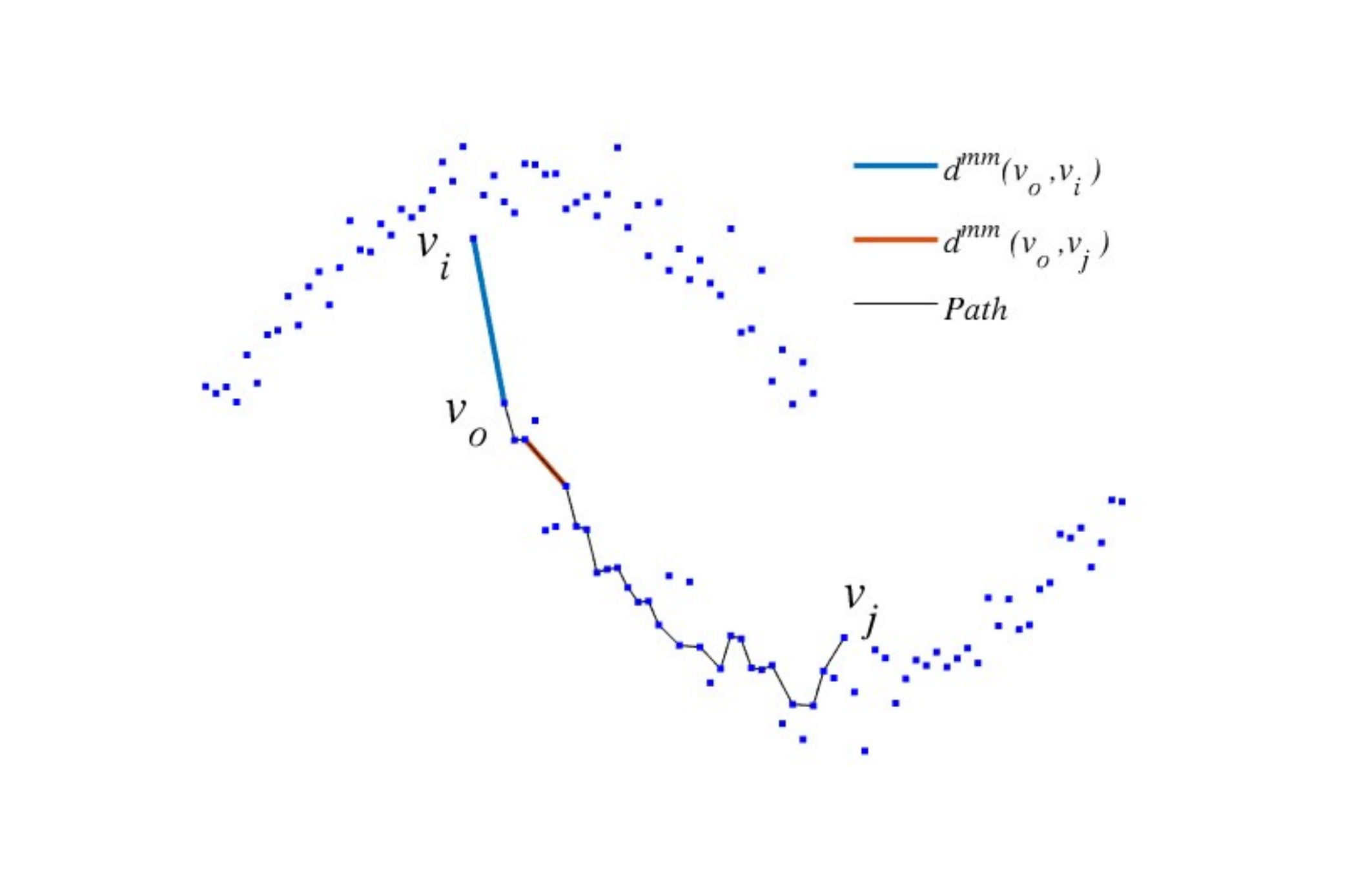}}
 \subcaptionbox{}{\includegraphics[width=6.5cm]{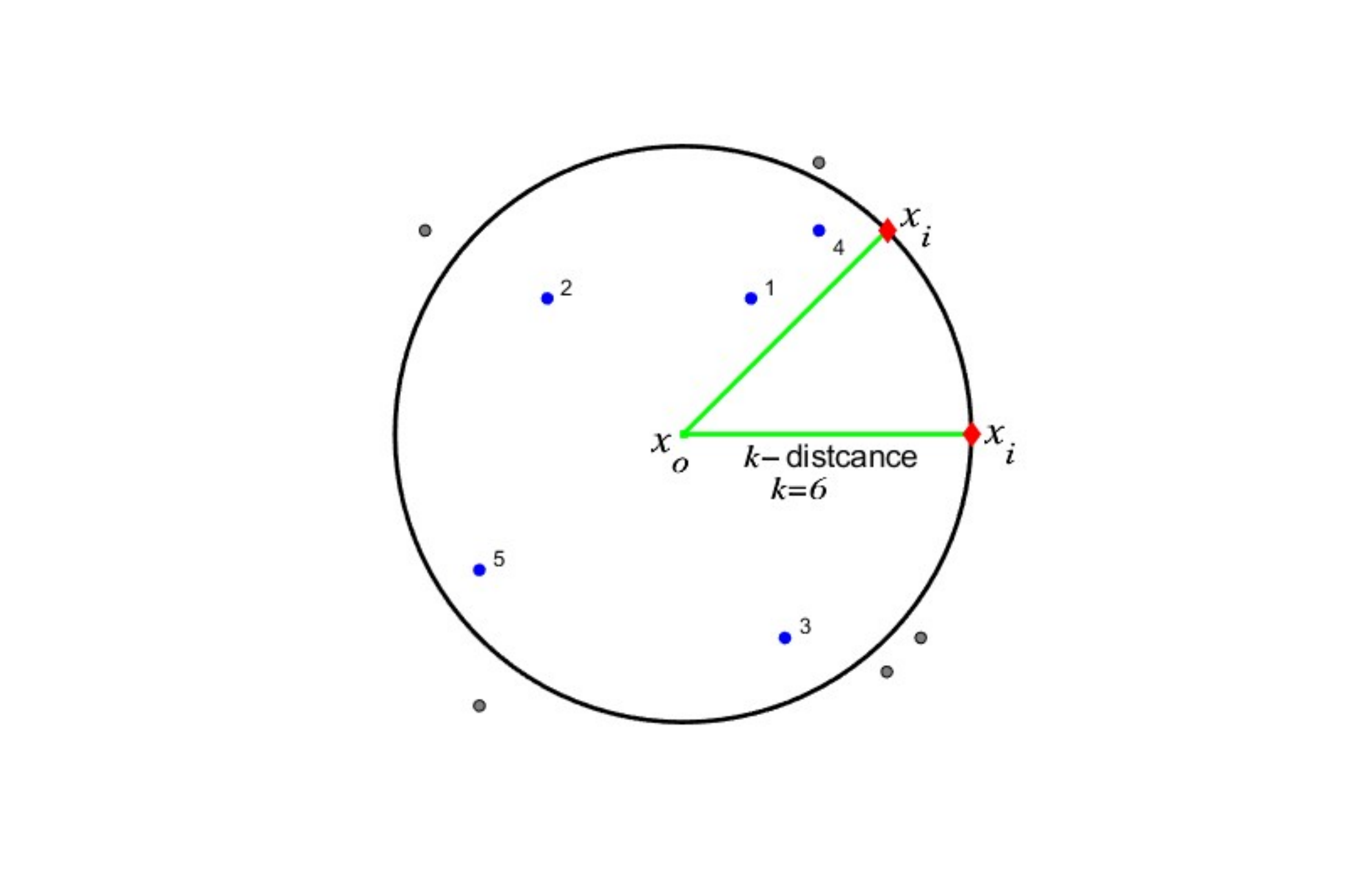}}
  \caption{\small{(a) Minmax distance. The paths from $v_o$ to $v_i$ and $v_o$ to $v_j$ are the unique paths in corresponding MST. The lengths of the bold lines are respectively the minmax distances between $v_o$ and $v_i$, $v_o$ and $v_j$. (b) $k$-distance. There are two objects, for simplicity, simultaneously denoted as $x_i$, who fulfill the requirements in Concept \ref{definition:k-distance}. Thus the distance between $x_o$ and $x_i$ is the $k$-distance($x_o$). }}
  \label{figure:preliminaries}
\end{figure}

Minmax distance stresses the connectedness of objects via mediating elements rather than favoring high original similarity. Therefore, minmax distance reflects the similarity of objects more precisely when the shapes of clusters are arbitrary. Moreover, it has been proven that the wanted path $p$ is exactly the unique path in corresponding MST. In Fig.\ref{figure:preliminaries}.(a), the unique path in the MST from vertex $v_o$ to vertex $v_i$ is the wanted path from vertex $v_o$ to vertex $v_i$, so is the path from $v_o$ to $v_j$. Thus, minmax distance is also called $path \  distance$, which can be computed by the MST.
%----------------------------------------------------------------------------
\begin{definition}[$k$-distance \cite{wang2021relative}]
 For any positive integer $k$ and dataset $X$, the $k$-distance of object $x_o \in X$, denoted as $k$-distance($x_o$), is defined as the distance $d(x_o, x_i)$ between $x_o$ and an object $x_i \in X$ such that: (i) for at least $k$ objects $x_j \in X \setminus \{x_o \}$ it holds that $d(x_o, x_j) \le d(x_o, x_i) $, and (ii) for at most $k \mbox{-} 1$ objects $x_j \in X \setminus \{x_o\}$ it holds that $d(x_o, x_j) < d(x_o, x_i)$ . See Fig.\ref{figure:preliminaries}.(b).
\label{definition:k-distance}
\end{definition}

The $k$-distance of object $x_o$ represents the dispersion degree around a certain object $x_o$. The larger the $k$-distance, the more extensive the range required to contain $k$ objects around object $x_o$, which implies its suitableness as a density measurement. For simplicity, we will exploit $k$-distance($x_o$) and $k$-distance($v_o$) interchangeably, where $v_o$ is the vertex in graph $G$ corresponding to the object $x_o$ in dataset $X$.
%----------------------------------------------------------------------------

\section{The PaVa Algorithm}\label{section_algorithm}

In this section, we provide some definitions and detail the proposed algorithm, PaVa. We begin with the datasets composed of spherical clusters, where clusters' centers and radii are vital factors. Then, inspired by the spherical clustering methods, we extend the algorithm to handle non-spherical conditions by replacing the Euclidean distance with minmax distance.  After all the necessary issues are illustrated, we present the main algorithm. Moreover, to enhance its robustness under noise, we propose an adjusted MST and explain its reasonability. Finally, we discuss its time complexity.

\subsection{Definitions}

In this subsection, we make some definitions to describe the PaVa algorithm more coherently, which facilitate the comprehension of our proposal.

\newtheorem{theorem}{Definition}
\begin{theorem}[Adjusted MST]
Suppose $T=(V, E)$ is the Minimum Spanning Tree generated from dataset $X$ of size $N$, where $ V=\{v_1, v_2, \cdots,  v_N\}$ are vertices corresponding to objects, $ E=\{e_1, e_2, \cdots, e_{N \mbox{-} 1}\} $ are edges corresponding to connection relationship between objects;  $ weight(v_i,  v_j) $, the weight of edge between $v_i$ and $v_j$, is set to be the dissimilarity, such as Euclidean distance, between objects corresponding  to  $ v_i$ and $ v_j$. Keeping the connection relationship, the weights $weight'(v_i,  v_j) $ in the adjusted MST are

\begin{equation}
weight'(v_i,  v_j)=\sqrt[3]{ weight(v_i,  v_j)* k\-/distance(v_i)*k\-/distance(v_j)}.
\end{equation}
\label{definition:adjusted_MST}
\end{theorem}
Adjusted MST aims to adjust the weights of edges according to the density of objects. The sparser the region in which an object is located, the larger the weight of the edges connected to its corresponding vertices, which means that noise objects are set to be far from objects on the main body of the clusters.
\begin{theorem}[PDF of distances]
Suppose that $X$ is a dataset of $N$ objects: $X=\{x_1, x_2, ..., x_N\}$. For a given object $x_o \in X$ and a distance measurement $d(\cdot, \cdot)$ (such as Euclidean distance or minmax distance), the distances between $x_o$ and all data objects in dataset $X$, $\{d(x_o, x_1), d(x_o, x_2), ..., d(x_o, x_N)\}$ are used to estimate a probability density function (PDF). We define it as the PDF of distances.
\label{definition:PDF}
\end{theorem}
An example is shown in Fig.\ref{figure:spherical_distacne_PDF}. In Fig.\ref{figure:spherical_distacne_PDF}.(a), we set the object shown as a red diamond to be a fixed object, noted as $x_o$, and set the distance measurement to be Euclidean distance. The distances $\{d(x_o, x_1), d(x_o, x_2), ..., d(x_o, x_N)\}$ are presented as $z$ axis in Fig.\ref{figure:spherical_distacne_PDF}.(b). Sequentially, these distances are used to form the PDF of distances, which is approximated with a red curve in Fig.\ref{figure:spherical_distacne_PDF}.(c). The same is true in Fig.\ref{figure:nonsperical_distance_PDF}, with distance measurements replaced with minmax distances.
%----------------------------------------------------------------------------
\begin{theorem}[Radius of cluster]
Suppose that the fixed object $x_o$ is the selected center of a cluster. The PDF of distances corresponding to $x_o$ can be presented with a curve, shown as the red curve in Fig.\ref{figure:spherical_distacne_PDF}.(c). We define the location of the first valley on the curve as the radius of the cluster centered at $x_o$.
\label{definition:radius}
\end{theorem}
It is suitable to approximate the curve by the smoothed frequency histogram (red curves in Fig.\ref{figure:spherical_distacne_PDF}.(c) and Fig.\ref{figure:nonsperical_distance_PDF}.(b)) as we only need the location of the first valleys. As shown in Fig.\ref{figure:spherical_distacne_PDF}.(c) and Fig.\ref{figure:nonsperical_distance_PDF}.(b), the locations of the first valleys are suitable radii for clusters, which serve as clear boundaries between the current clusters whose centers are ascertained and other clusters.

\subsection{Key elements}

To better illustrate how our algorithm handles datasets with arbitrarily shaped clusters, we start with part \ref{center_radius_spherical} to demonstrate the center and radius of a spherical cluster, which is similar to the definitions in \cite{ma2022new} but with a slight difference. Next, we link the center and radius to the arbitrarily shaped cluster in parts \ref{center_nonspherical} and \ref{radius_nonspherical}. However, the clustering problem for arbitrarily shaped clusters is more complicated than in the spherical case. On the one hand, the centers are not clearly defined for arbitrarily shaped clusters, so we need a generalized definition of centers. On the other hand, arbitrarily shaped clusters cannot be separated by a hypersphere, and the boundaries between the arbitrarily shaped clusters are always non-convex, which can not be determined by radii.
Therefore, we generalize the spherical clustering problem to the non-spherical clustering problem from two aspects: the cluster's center and radius.

\subsubsection{Center and radius for spherical clusters}\label{center_radius_spherical}
\begin{figure}
  \centering
 \subcaptionbox{}{\includegraphics[width=7cm]{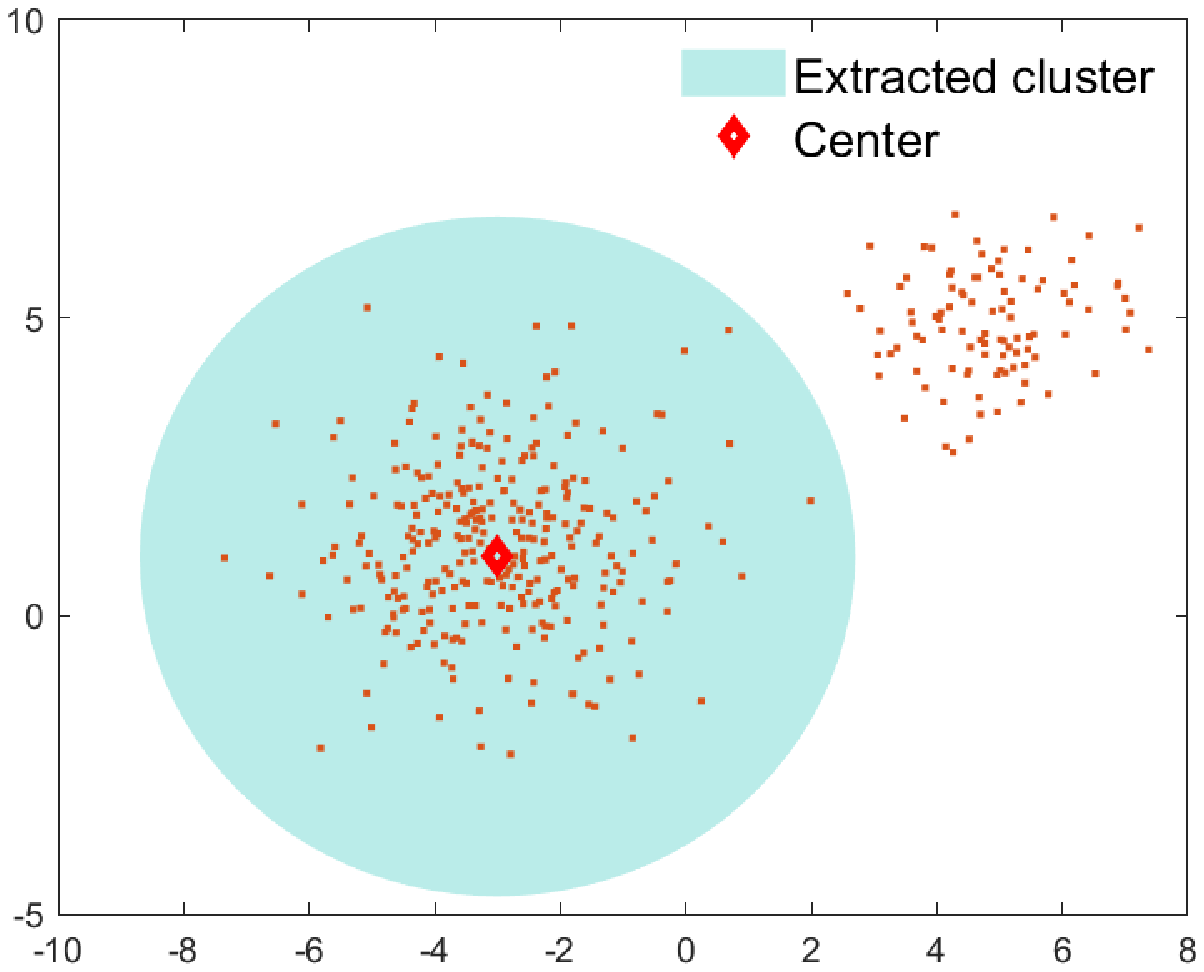}}

 \subcaptionbox{}{\includegraphics[width=6cm]{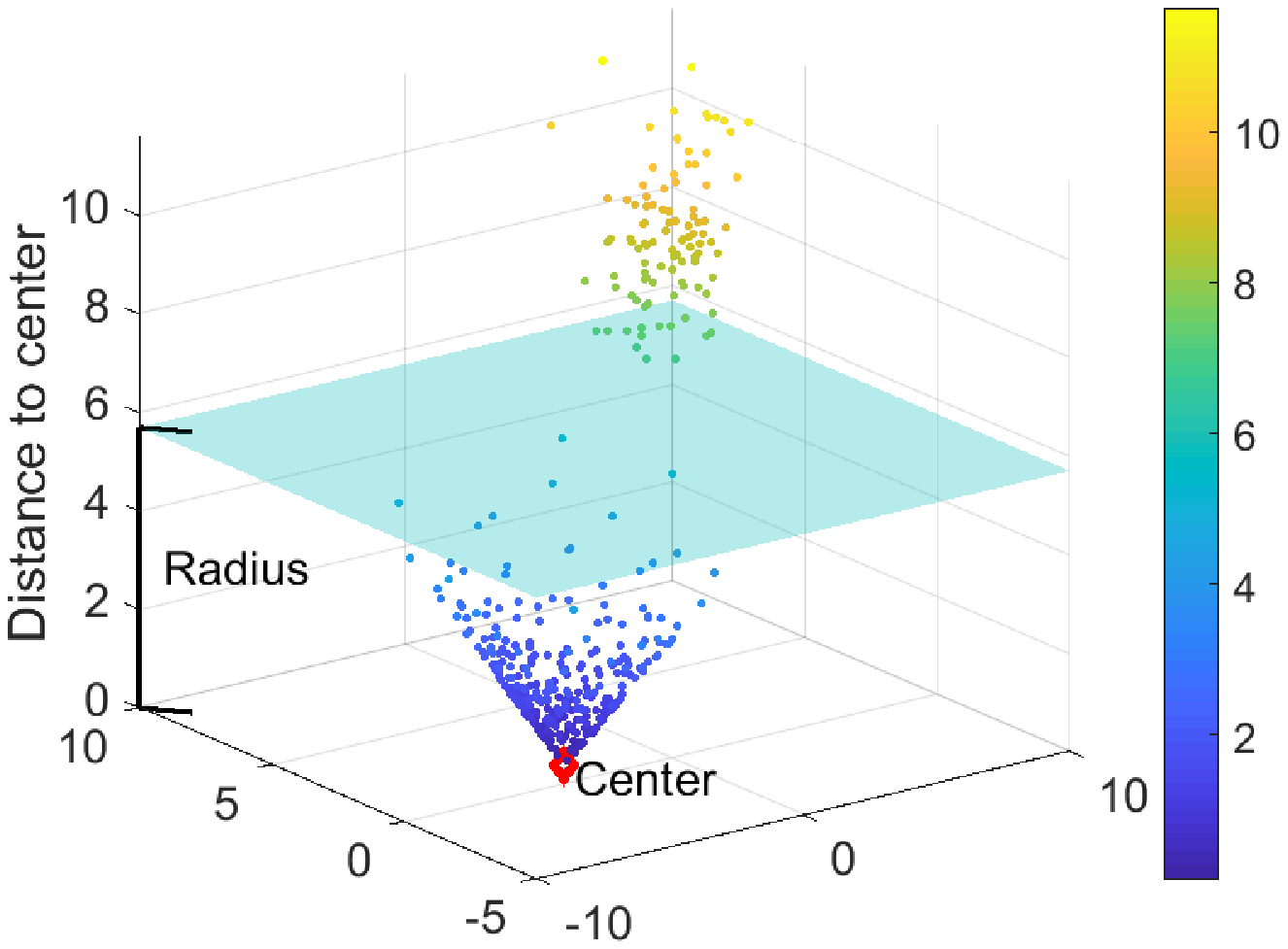}}
 \subcaptionbox{}{\includegraphics[width=6cm]{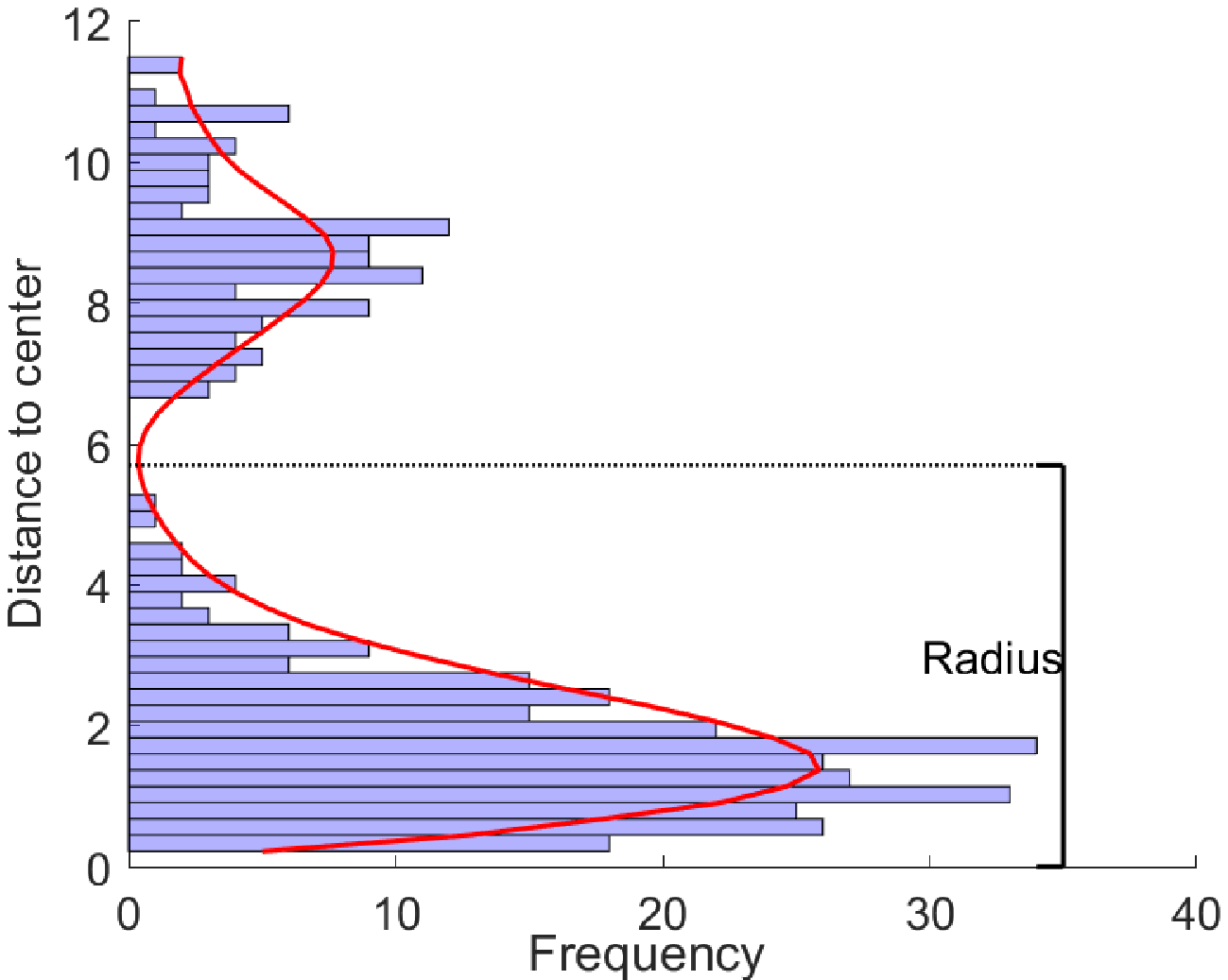}}
  \caption{\small{(a) Extracting spherical cluster. (b) Distances to cluster center. (c) The approximated PDF of distances between the center and all objects in the dataset. A spherical cluster can be extracted if its center and radius are known. The center is the object whose density is maximum and the radius is the location of the first valley on the PDF of distances.}}
  \label{figure:spherical_distacne_PDF}
\end{figure}

 Any spherical cluster would be extracted precisely as long as its center and radius are known. This is shown in Fig.\ref{figure:spherical_distacne_PDF}.(a).

According to the features of the spherical cluster, the object located at the densest area is deservedly set to be the center. As the center is known, the radius can be determined via the PDF of distances, which is established by the distances between the center and the rest of the objects. As shown in Fig.\ref{figure:spherical_distacne_PDF}.(b), keep the $x$ and $y$ coordinates and present their distances to the center as $z$ coordinates. There is a clear boundary separating objects belonging to the current cluster from the others. The location of the boundary corresponds to the first valley in the curve, and the curve represents the PDF of distances. The distances between the center and objects belonging to the current cluster are shorter than the value corresponding to the location of the first valley. We henceforth refer to the location of the first valley as the cluster radius.  For clarity and robustness, the PDF is smoothed from the frequency histogram to the red curve. Thus the location of the first valley on the curve is a reasonable value for the radius. Moreover, the PDF of distances is discussed theoretically in \cite{ma2022new}.

\subsubsection{Center for arbitrarily shaped cluster} \label{center_nonspherical}
\begin{figure}[H]
  \centering
  \subcaptionbox{}{\includegraphics[width=6cm,height=5.0cm]{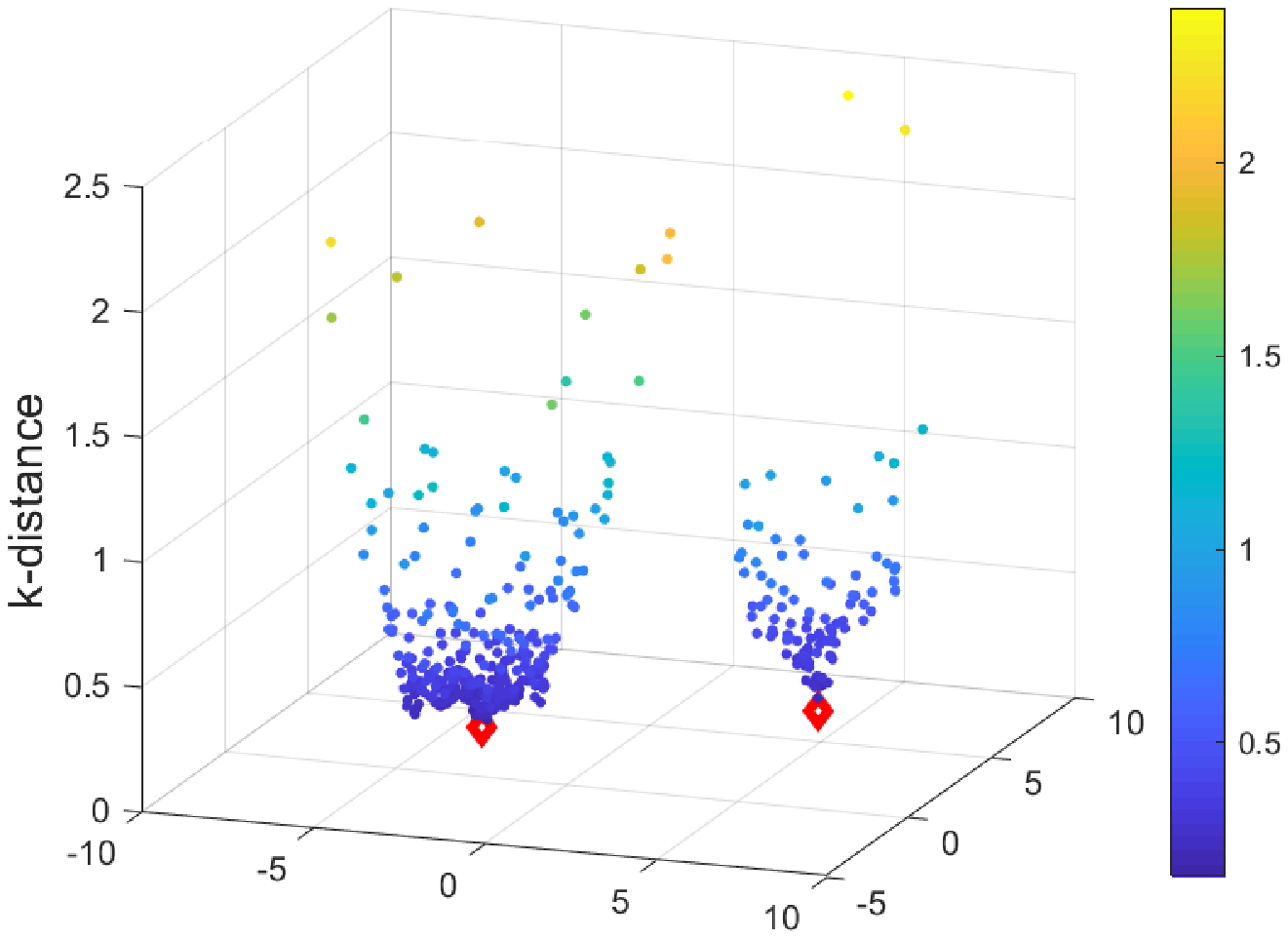}}
  \subcaptionbox{}{\includegraphics[width=6cm,height=5.0cm]{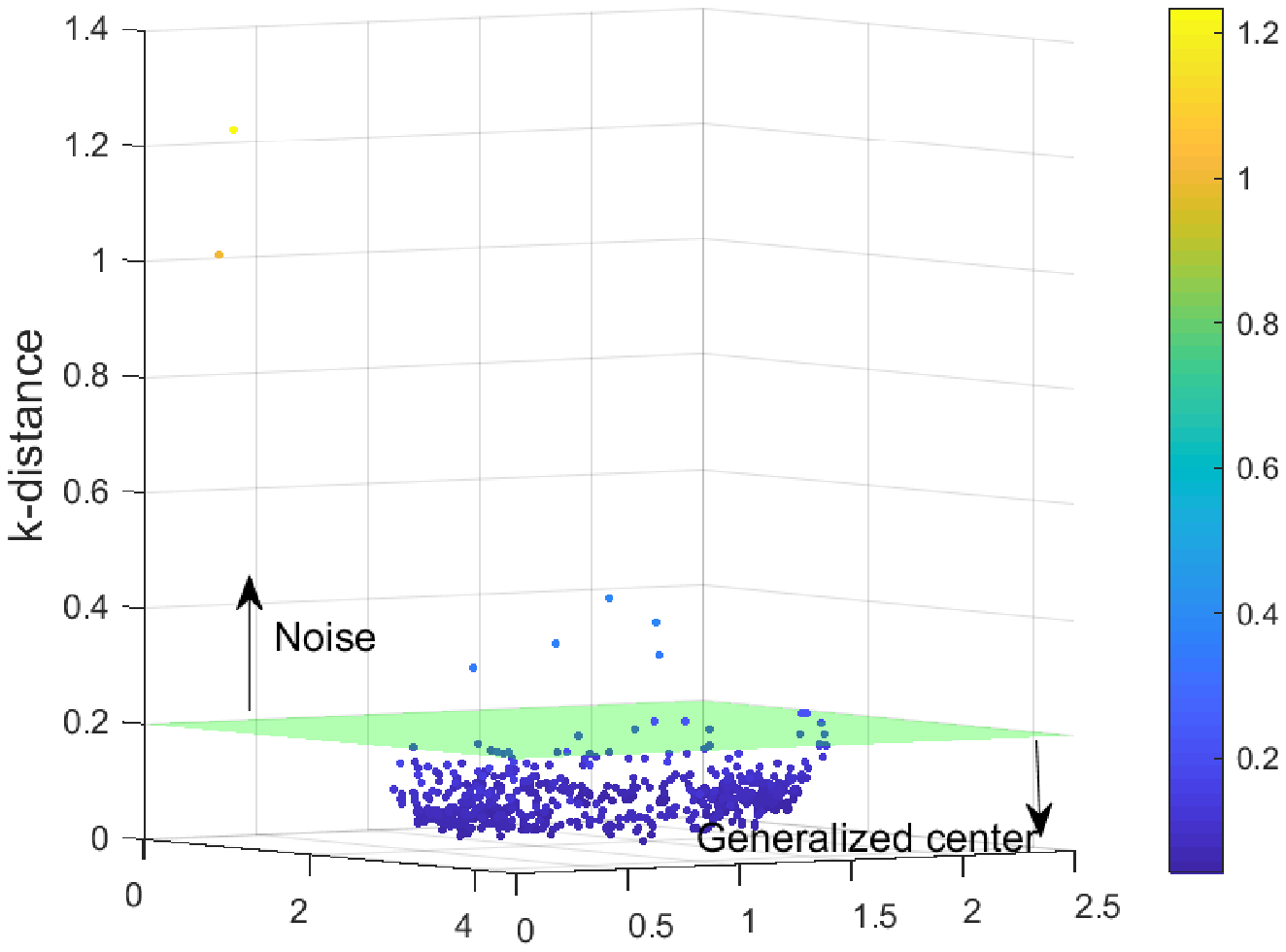}}
  \caption{\small{(a) Centers of spherical cluster found by $k$-distances. (b) Generalized centers of non-spherical clusters found by $k$-distance. Keeping $x$ and $y$ coordinates, the $z$ coordinates are set to be the density measurement, $k$-distance. In non-spherical clusters, we need an object located in the main body of the cluster to rule out noise objects.}}
  \label{figure:knn_center}
\end{figure}

In the spherical clustering problem, we chose the object located at the densest area as the center. As shown in Fig.\ref{figure:knn_center}.(a), we measure the density by $k$-distance.
Similarly, we can also use $k$-distance to ascertain the center of the non-spherical cluster. The center of an arbitrarily shaped cluster is hard to define.  Fortunately, this is unnecessary for our algorithm; the only requirement is to find an object that lies in the main body of the cluster and avoid choosing a noise object as the center. We name it the generalized center; see Fig.\ref{figure:knn_center}.(b). Only for simplicity, we choose the object whose density is highest, that is, with the shortest $k$-distance.
In this way, the ascertained center reserves the same properties as in the spherical clustering problem:

(i) The minmax distances between the generalized center and objects belonging to the current cluster are short.

(ii) The minmax distances between the generalized center and objects belonging to different clusters are long.

(iii) A boundary exists in the two groups of distances.

The above three properties hold when minmax distance is applied, as detailed in the following subsection.

\subsubsection{Radius for arbitrarily shaped clusters} \label{radius_nonspherical}
\begin{figure}[H]
  \centering
  \subcaptionbox{}{\includegraphics[width=6cm,height=5.0cm]{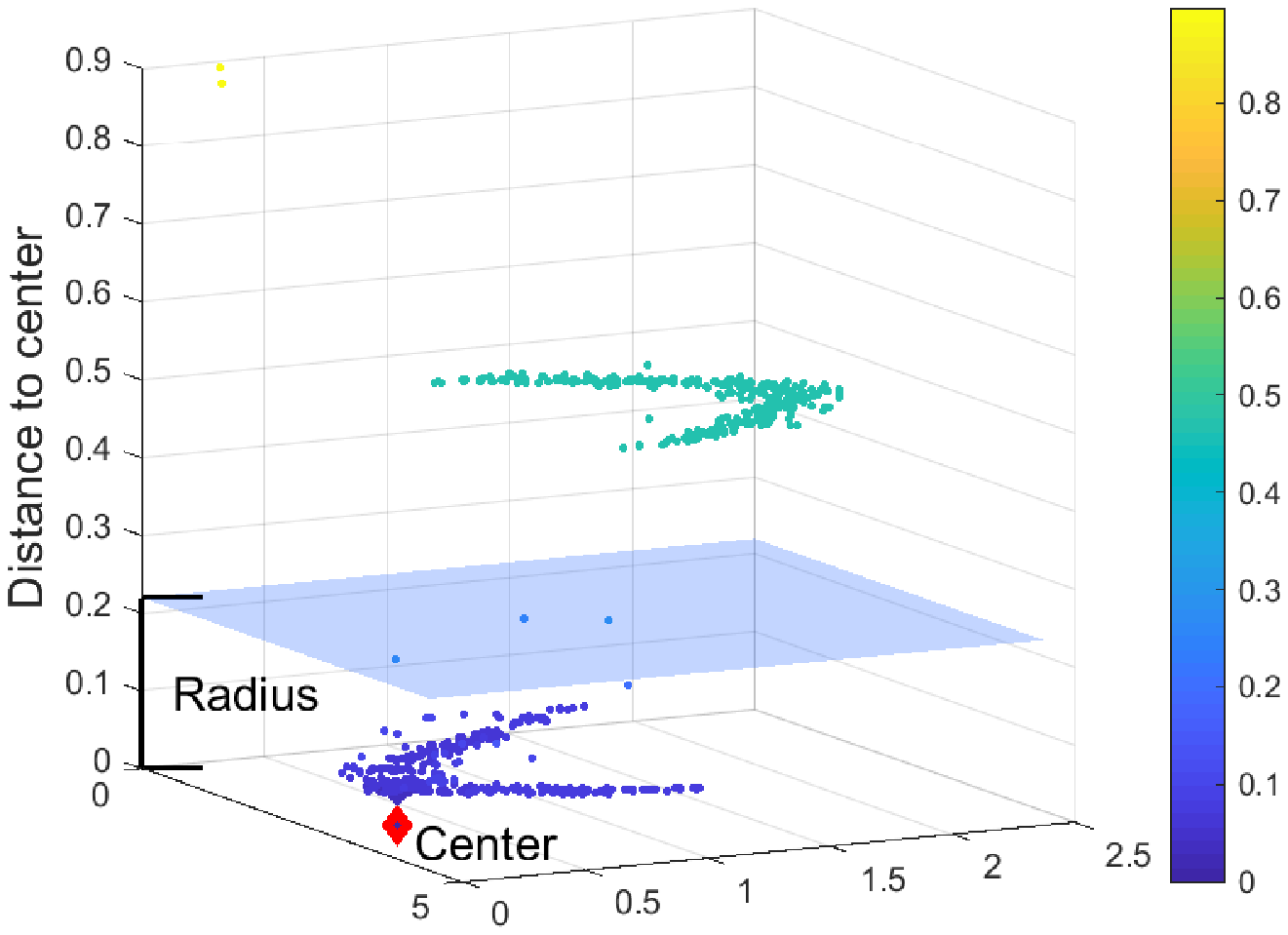}}
  \subcaptionbox{}{\includegraphics[width=6cm,height=5.0cm]{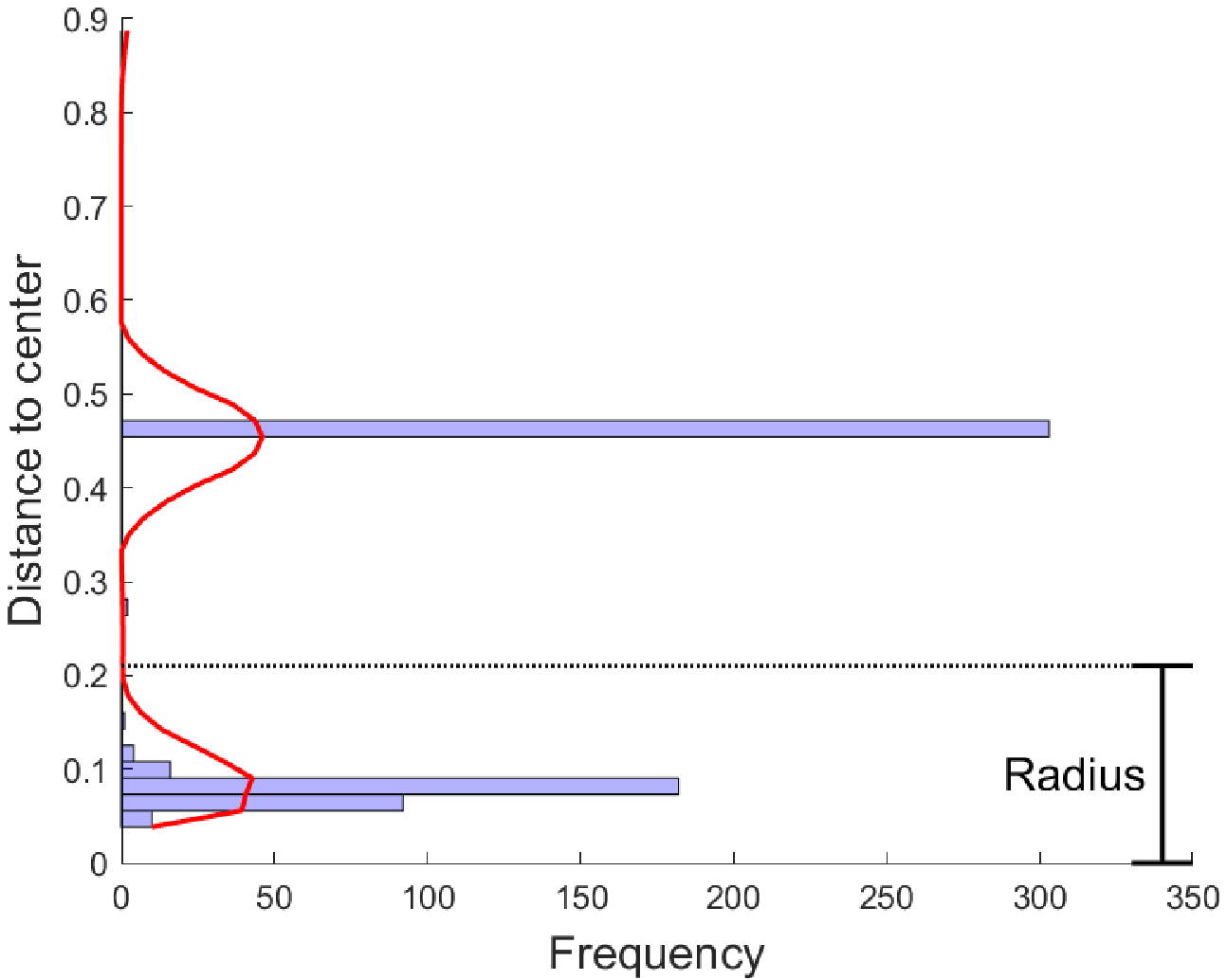}}
  \caption{\small{(a) Minmax distances to a generalized center. (b) The approximated PDF of minmax distances between the center and all objects. Keeping $x$ and $y$ coordinates, the $z$ coordinates are set to be the minmax distances between objects to center. Objects with distances to the center shorter than the radius are labeled as the same cluster with the known center. The distances between the rest of objects and the center are greater than the radius.}}
  \label{figure:nonsperical_distance_PDF}
\end{figure}

Now that we have a generalized center, the next step is to determine the radius of a cluster. Similar to the spherical clustering problem, the PDF of distances between the center and all objects in the dataset is constructed and used. As shown in Fig.\ref{figure:nonsperical_distance_PDF}.(a), keeping the $x$ and $y$ coordinates of objects, minmax distances between these objects to the generalized center are presented in $z$ coordinate. In Fig.\ref{figure:nonsperical_distance_PDF}.(b) the PDF of minmax distances is established. The location of the first valley in the corresponding curve is precisely the location of the boundary between the clusters.  We then obtain the cluster radius for a non-spherical cluster, which is analogous to the spherical case. The only difference is that the Euclidean distance is replaced by minmax distance. As shown in Fig.\ref{figure:nonsperical_distance_PDF}., even though the two clusters cannot be separated linearly, the boundary between the two groups of distances is clear under minmax distance. With this approach, the objects whose minmax distances to the generalized center are smaller than the determined radius are infallibly labeled as the same cluster as the generalized center.

We show the pseudocode of calculating minmax distance in Algorithm \ref{algorithm:minmax_dist} and the pseudocode of calculating radius in Algorithm \ref{algorithm:radius}.

\begin{algorithm}
\small
	\renewcommand{\algorithmicrequire}{{\bf Input:}}
	\renewcommand{\algorithmicensure}{{\bf Output:}}
	\caption{\small {Minmax Distance}}
	\begin{algorithmic}[1]
		\REQUIRE  $\texttt{center}$, tree $T$
		\ENSURE  $\texttt{minmax\_dist\_center}$: the minmax distances between $\texttt{center}$ all objects
		\STATE  $\texttt{minmax\_dist\_center} \gets $ zeros(n,1)
        \STATE  $\texttt{visited} \gets$ zeros(n,1)
        \STATE  $\texttt{queue} \gets \texttt{center}$
        \WHILE  {$\texttt{queue}$ is nonempty}
        \STATE  $\texttt{head\_neighbors}  \gets$ unvisited neighbors of $\texttt{queue}(1)$ in $T$
        \STATE  $\texttt{queue} \gets  [\texttt{queue}; \texttt{head\_neighbors} ]$
        \STATE  $\texttt{visited}(\texttt{head\_neighbors}) \gets 1$
        \WHILE  {$\texttt{head\_neighbors} $ is nonempty}
        \STATE  $\texttt{weight\_new} \gets $ weight(\texttt{head\_neighbors}(1), \texttt{queue}(1))
        \STATE  $\texttt{minmax\_dist\_center}(\texttt{head\_neighbors}(1)) \gets$ max\{\texttt{weight\_new}, \texttt{minmax\_dist\_center}(\texttt{queue}(1))\}
        \STATE   delete $\texttt{head\_neighbors}(1)$ from $\texttt{head\_neighbors}$
        \ENDWHILE
        \STATE  delete $\texttt{queue}(1)$ from $\texttt{queue}$
        \ENDWHILE
	\end{algorithmic}
    \label{algorithm:minmax_dist}
\end{algorithm}

\begin{algorithm}
\small
	\renewcommand{\algorithmicrequire}{{\bf Input:}}
	\renewcommand{\algorithmicensure}{{\bf Output:}}
	\caption{\small {Radius}}
	\begin{algorithmic}[1]
		\REQUIRE  $\texttt{minmax\_dist\_center}$ \textcolor{green}{\% calculated by Algorithm \ref{algorithm:minmax_dist} }
		\ENSURE  $\texttt{radius}$
		\STATE  $\texttt{minmax\_dist\_center}>prctile(\texttt{minmax\_dist\_center},99) \gets$ NaN
        \STATE  [$\texttt{frequency}$,$\texttt{dist}$] $\gets$ hist($\texttt{minmax\_dist\_center}$,200);
        \STATE  $\texttt{frequency} \gets \texttt{frequency}$-1
        \STATE  $\texttt{sm\_frequency} \gets$ smooth($\texttt{frequency}$)
        \FOR{$i \gets 2$ to $n$}
        \IF{$\texttt{sm\_frequency}$(i) $\leq$ min\{$\texttt{sm\_frequency}$(i-1),$\texttt{sm\_frequency}$(i+1)\}}
        \STATE break
        \ENDIF
        \ENDFOR
        \STATE $\texttt{radius} \gets  $\texttt{dist}$(i)$
	\end{algorithmic}
    \label{algorithm:radius}
\end{algorithm}

\subsection{Framework}

The PaVa clustering algorithm is comprised of three steps and is shown in Algorithm \ref{algorithm:main}. The first step is preparation work, including the calculation of $k$-distance and MST (line:1-2). The second step is the main body of the algorithm (line:4-10) which will be iterated $M$ times, where $M$ is the number of clusters. Each loop consists of locating a center (line:6), determining the corresponding radius (line:7-8), extracting this cluster based on the previously obtained center and radius (line:9), and finally, making a judgment on whether all clusters are extracted (line:4). In the last step, we assign the unlabeled objects with the help of the tree structure (line:12).

\begin{algorithm}
\small
	\renewcommand{\algorithmicrequire}{{\bf Input:}}
	\renewcommand{\algorithmicensure}{{\bf Output:}}
	\caption{\small {Main Function}}
	\begin{algorithmic}[1]
		\REQUIRE dataset $\texttt{X}$, parameter $k$ for $k$-distance
		\ENSURE  $\texttt{labels}$ of all objects in the dataset
		\STATE  calculate $\texttt{k\-/distance}$s of all objects in $\texttt{X}$
        \STATE  generate the $\texttt{MST}$ from $\texttt{X}$
        % \STATE  construct $\texttt{adjusted MST}$ by $\texttt{k\-/distance}$s and $\texttt{MST}$
        \STATE  $M \gets 0$
        \WHILE  {number of unlabeled objects $> 90\%$}
        \STATE  $M \gets M+1$  \ \ \ \ \textcolor{green}{\% the number of found clusters}
        \STATE  $\texttt{center} \gets$ unlabeled object whose $\texttt{k\-/distance}$ is minimum among all the unlabeled objects
        \STATE  $\texttt{minmax\_dist\_center} \gets $ minmax distances($\texttt{center}$,$\texttt{MST}$) \textcolor{green}{\% Algorithm \ref{algorithm:minmax_dist}}
        \STATE  $\texttt{radius} \gets$ radius$\texttt{(minmax\_dist\_center)}$ \textcolor{green}{\% Algorithm \ref{algorithm:radius}}
        \STATE  $\texttt{cluster\_extracted} \gets$ unlabeled objects whose minmax distances to $\texttt{center}$ are less than $\texttt{radius}$
        \STATE  $\texttt{labels(cluster\_extracted)} \gets M$
        \ENDWHILE
        \STATE  $\texttt{labels}$(unlabeled objects) $\gets \texttt{labels}$(nearest labeled neighbors) on the $\texttt{adjusted MST}$
	\end{algorithmic}
    \label{algorithm:main}
\end{algorithm}

\subsection{A refinement: adjusted MST}

\begin{figure}[H]
  \centering
  \subcaptionbox{}{\includegraphics[width=6cm,height=5.5cm]{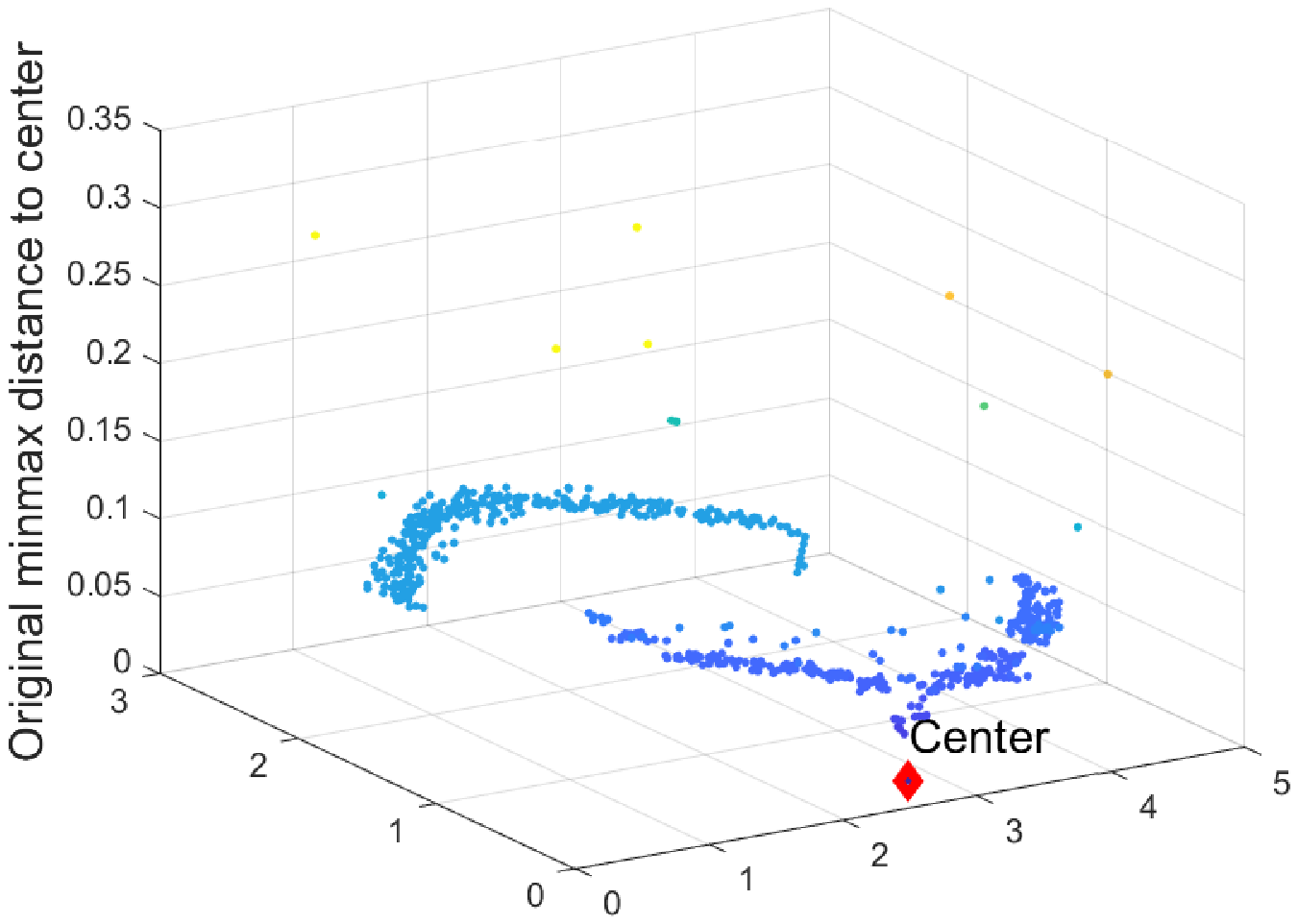}}
  \subcaptionbox{}{\includegraphics[width=6cm,height=5.5cm]{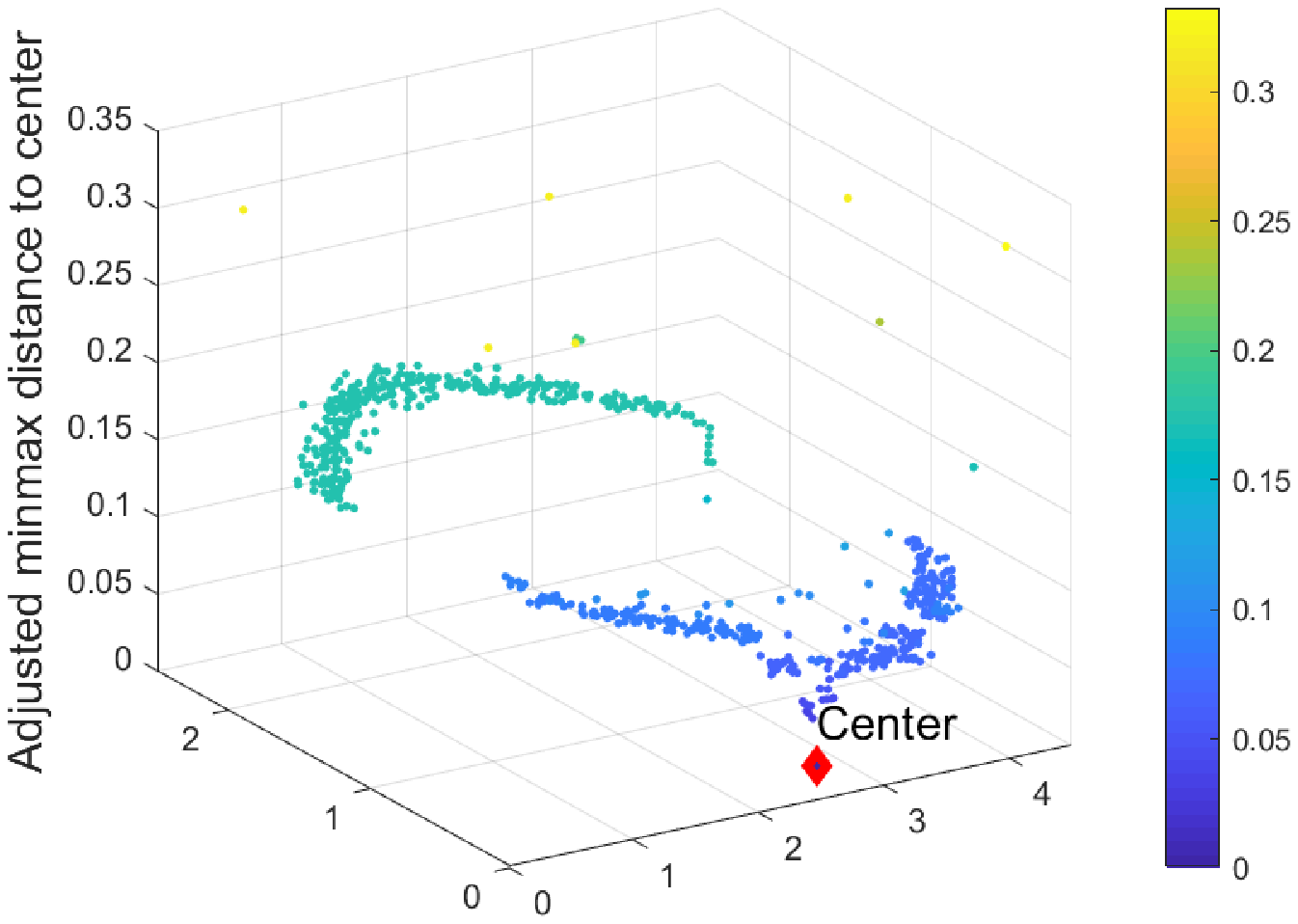}}
  \caption{\small{(a) Minmax distances between the center and all objects when `noise bridge' exists, calculated from vanilla MST. (b) Minmax distances between center and all objects when `noise bridge' exists calculated from adjusted MST. The boundary on minmax distances calculated from vanilla MST are not clear when noise exists. After adjusting the MST, the boundary is clear and easy to get in PDF of minmax distances.}}
  \label{figure:noise_bridge}
\end{figure}

According to Concept \ref{definition:minmax distance} in Section \ref{section_preliminaries}, minmax distance is not stable when noise structure exists. As shown in Fig.\ref{figure:noise_bridge}.(a), and dataset \emph{twomoons\_bridge} in Table \ref{Table non-spherical}, a `noise bridge' connects two clusters. The distances between the center and objects belonging to different clusters are close, making it harder to separate these two clusters (the comparison between Fig.\ref{figure:nonsperical_distance_PDF}.(a) and  Fig.\ref{figure:noise_bridge}.(a)). That is, the first valley on the PDF of distances becomes ill-defined and even disappears due to the ambiguous distribution of the minmax distances in the presence of  `noise bridge' . Therefore, the proper radius of the current cluster is hard to determine.

From the perspective of our ultimate goal, we expect a situation where objects belonging to the same cluster are close to each other and objects belonging to different clusters are far from each other. It is the presence of noise that makes minmax distance between objects belonging to different clusters shorter. To alleviate the impact of noise, we should increase the distances shortened by the `noise bridge'. As mentioned before, the noise objects are located at the low-density regions. To this end, $k$-distance could be used to adjust minmax distance to alleviate the impact of noise. The larger the $k$-distance, the sparser the area, and the greater the probability the object to be noise.  Therefore, the distances between it and other objects should be adjusted farther. So we make an adjustment on the MST, which only needs a slight modification based on Algorithm \ref{algorithm:main} (line:1-2).

According to Definition \ref{definition:adjusted_MST}, we include that the adjustments from $weight(v_i, v_j)$ to $weight'(v_i, v_j)$  are slight when the values of $k$-distance($v_i$) and $k$-distance($v_j$) are small; on the contrary, the adjustment on an edge is drastic if at least one vertex (object) corresponding to this edge is a noise object with a large $k$-distance.  In this way, the distances between objects within a cluster are maintained to a great extent; the distances between noise and the other objects increases substantially in the adjusted MST. Replacing the original MST with the adjusted MST, the minmax distances between objects belonging to different clusters increases due to the adjustment on the noise objects in its path. As a result, the distribution of minmax distances is more scattered, which implies sharper boundaries between clusters and a more accurate radius. As is shown in Fig.\ref{figure:noise_bridge}., the boundary between clusters under the minmax distances calculated by adjusted MST (Fig.\ref{figure:noise_bridge}.(b)) is clearer than the boundary under the minmax distances calculated by the original MST (Fig.\ref{figure:noise_bridge}.(a)). Moreover, $k$-distance is used to find the general center and construct adjusted MST, separately. Therefore, using $k$-distance as a density measurement is highly efficient and does not increase the computation load.

\subsection{Analysis on time complexity}

The main computational burden arises from two modules.

In the first module, we construct the vanilla MST and establish the adjusted MST on the basis of $k$-distance. On the one hand, constructing an MST in Euclidean space needs $O(N\log N)$, (In 2D Euclidean space, it needs $O(N\log N)$; in 3-dimensional Euclidean space, it needs $O((N\log N)^{\frac 4 3})$ to construct an exact MST; it remains to be studied when the dimension is higher than 2. However, we do not need an exact one, and an approximation is good enough; we only need it to keep a connected structure and maintain reasonable minmax distance). On the other hand, multidimensional binary search tree (MBTS) is used to calculate $k$-distance. Building the MBTS needs $O(N\log N)$, finding the nearest neighbor on MBST needs $O(\log N)$, and deleting a vertex on MBST needs $O(\log N)$, so calculating $k$-distances for all objects needs  $O(kN\log N)$.

In the second module, the first step ascertains a center in $O(N)$ by finding the object with a minimum $k$-distance. The second step extracts the corresponding cluster in $O(N)$, including the calculation of the minmax distances between the center and all objects, which equals traversing the tree in $O(N)$ and constructing PDF while finding its first valley in $O(N)$. Finding and extracting a cluster are repeated $M$ times ($M$ is the number of clusters). So the time complexity for the second module is $O(MN)$. Moreover, the time to assign unlabeled objects is negligible thanks to the constructed tree structure.

In conclusion, the time complexity is $O(kN\log N +MN)$. When the number of clusters $M < k\log N$, it is reduced to $O(kN\log N)$, where $k$ is set to be an integer around $\log N$ as a rule of thumb.

\section{Experiments }\label{section_experiment}

In this section, the accuracy of the PaVa clustering algorithm is evaluated with three metrics, namely, Rand index (RI), Adjusted Rand index (ARI), and F score (FS). Meanwhile, its efficiency is evaluated by the running time (RT). To show the universality of the PaVa clustering algorithm, we apply it to both synthetic datasets and real-world datasets. Moreover, we analyze the robustness of our algorithm on different values of parameter $k$.

\subsection{Synthetic datasets}
This subsection is comprised of two parts. In the first part, we apply the PaVa clustering algorithm to two datasets composed of spherical clusters.  In the second part, several popular non-spherical datasets are used to validate the ability of the PaVa clustering algorithm under more complex conditions. Meanwhile, clustering methods for non-spherical datasets, hierarchical clustering algorithm (single-linkage), kernel k-means, DBSCAN \cite{ester1996density} and its variant DC-SKCG \cite{ZHANG2021344}, spectral clustering algotithm \cite{ng2001spectral} with ideal parameter setting are used as benchmarks; additionally, another minmax path-based clustering algorithm named Global Optimal Path-based Clustering algorithm (GOPC) \cite{liu2019global} and Cut-Point Clustering algorithm (CutPC) \cite{li2020novel} is also applied.

\subsubsection{Datasets composed of spherical clusters}

The radii of clusters are determined in a data-driven manner, so it is flexible in handling heterogeneous clusters. We collected two datasets; the first is a two-dimensional dataset (dataset$1$) with six heterogeneous clusters with different sizes and volumes; the second is a three-dimensional dataset named \emph{hapta} (dataset$2$) with six homogenous clusters and one denser cluster at the origin.

Using the given datasets, we compare the PaVa clustering algorithm with classical clustering algorithms for spherical dataset, k-means, x-means, GMM, and hierarchical clustering algorithm (average-linkage). All of the mentioned clustering methods require users to input parameters. In a certain range, the larger the parameter $k$, the more accurate the center is in the PaVa clustering algorithm. Meanwhile, the larger the parameter $k$, the larger the computation cost. So there is a trade-off between accuracy and efficiency. Nevertheless, in the last part of this section, we will show that the value of parameter $k$ doesn't have a significant impact on running time, so ignoring the efficiency, it is common to set the value of $k$ to be an integer around $\log N$ (we set  $k=10$ for these two spherical datasets). This setting is more user-friendly than any of the other four clustering methods for the following reason: the number of clusters needs to be inputted by users in the other four clustering methods, which is difficult without prior knowledge. In this part, the number of clusters in k-means, GMM and hierarchical clustering algorithm is set according to the truth-value. In addition, x-means requires two parameters, an estimated lower bound and an upper bound for the number of clusters. We set them to be 2 and the truth-value, respectively.

As shown in Table \ref{Table spherical}, our algorithm is at least as good as the classical clustering algorithms, even though the actual number of clusters is given for classical clustering methods. The hierarchical clustering algorithm and the PaVa clustering algorithm get the most reasonable clustering result. Thanks to the equal volume of clusters and clear boundaries between clusters, k-means performs well on dataset 2. Nevertheless, its performance is poor on dataset$1$, in which the clusters are heterogeneous. K-means tends to partition datasets into equal volume and has a problem of local convergence, so a huge cluster could be partitioned into two tiny clusters, and a tiny cluster could be incorrectly assigned to its adjacent cluster. Similar to k-means, GMM is also an EM-type algorithm that may get stuck in local convergence. The results of x-means on both datasets do not match the actual conditions. This is because the framework of x-means is the dichotomy.  It may stop early even if the number of clusters is slightly large due to the BIC-based terminal condition.

\begin{table}[H]
\caption{Accuracy of the indicated algorithms and spherical datasets.}\label{Table spherical}
\centering
\resizebox{\textwidth}{!}{%
\begin{tabular}{c|ccc|ccc|ccc|ccc|ccc}
\hline
\diagbox{Methods}{Performance}{Dataset}
                          & \multicolumn{3}{c|}{PaVa}     & \multicolumn{3}{c|}{k-means}       & \multicolumn{3}{c|}{x-means}       & \multicolumn{3}{c|}{GMM}         & \multicolumn{3}{c}{Hierarchical}                   \\ \hline
\multirow{3}{*}{\makecell{Dataset1 \\ N=1500 }}
                          & \multicolumn{3}{c|}{\includegraphics[width=4.5cm,height=4cm]{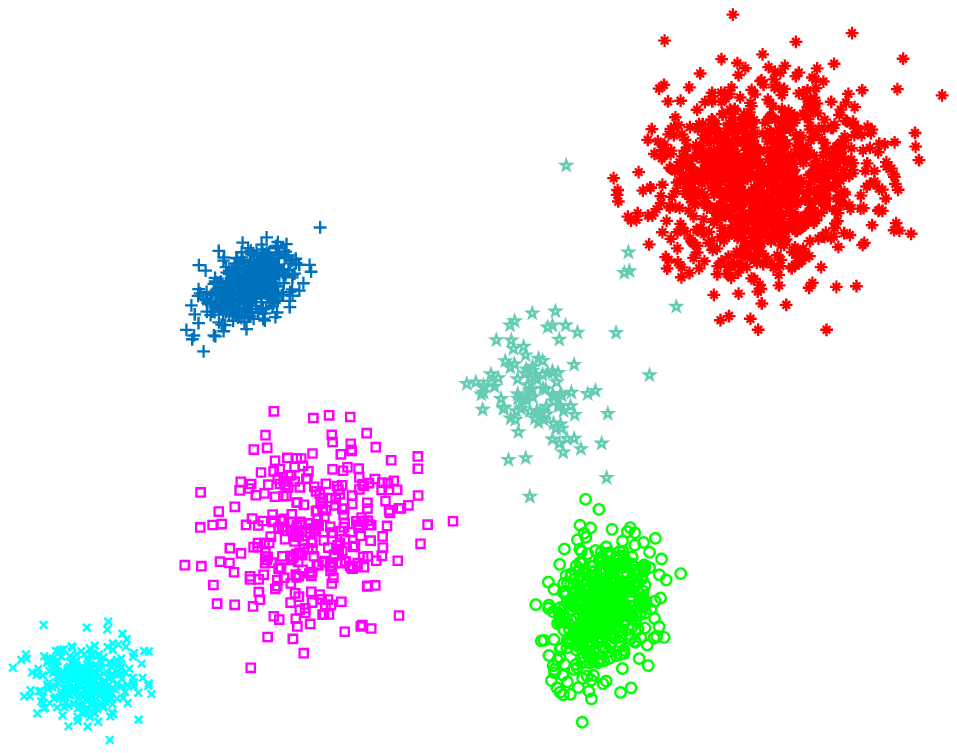}}
                          & \multicolumn{3}{c|}{\includegraphics[width=4.5cm,height=4cm]{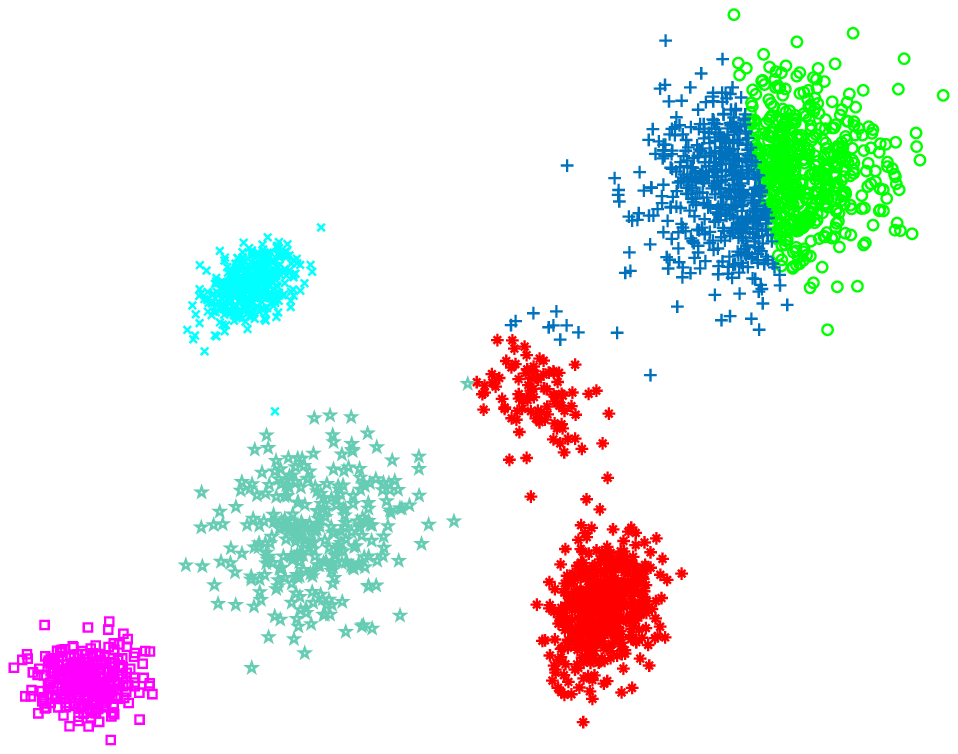}}
                          & \multicolumn{3}{c|}{\includegraphics[width=4.5cm,height=4cm]{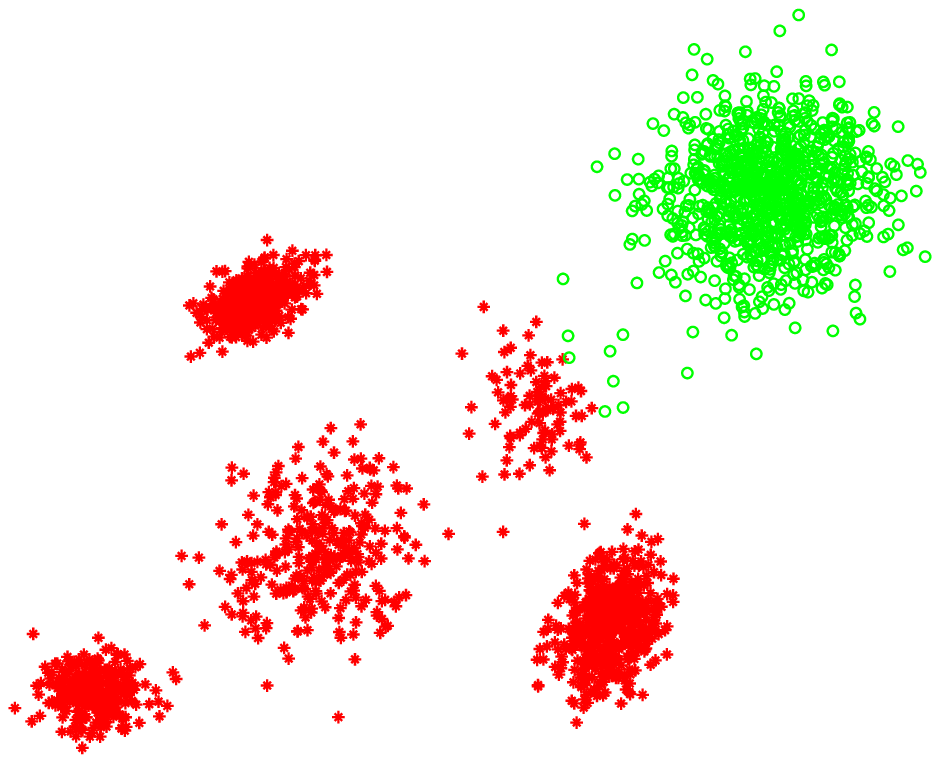}}
                          & \multicolumn{3}{c|}{\includegraphics[width=4.5cm,height=4cm]{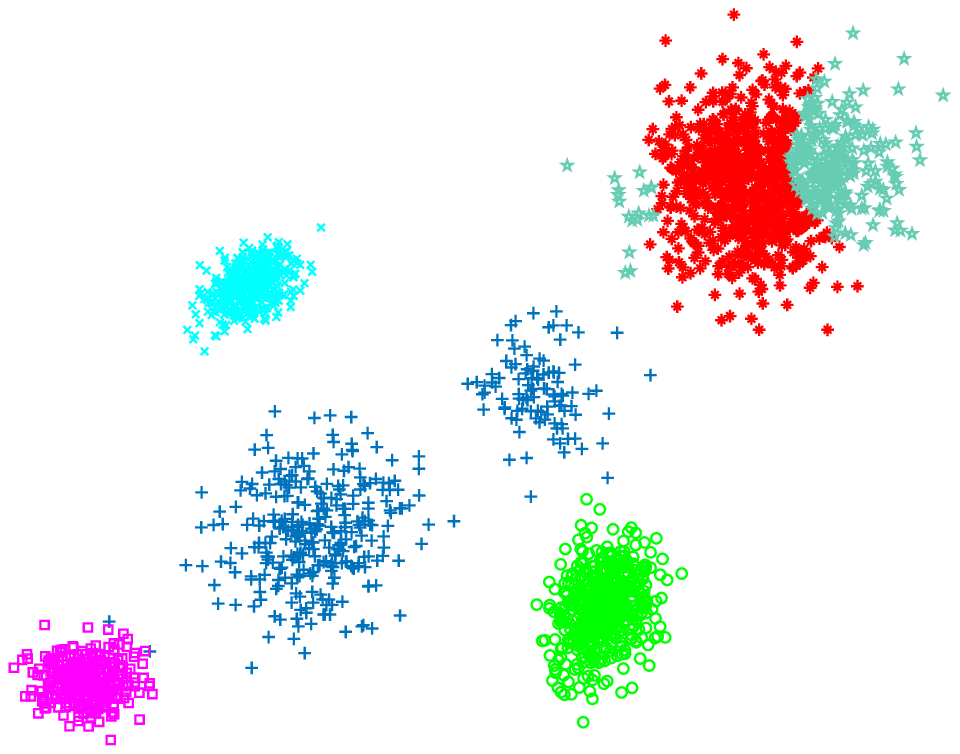}}
                          & \multicolumn{3}{c} {\includegraphics[width=4.5cm,height=4cm]{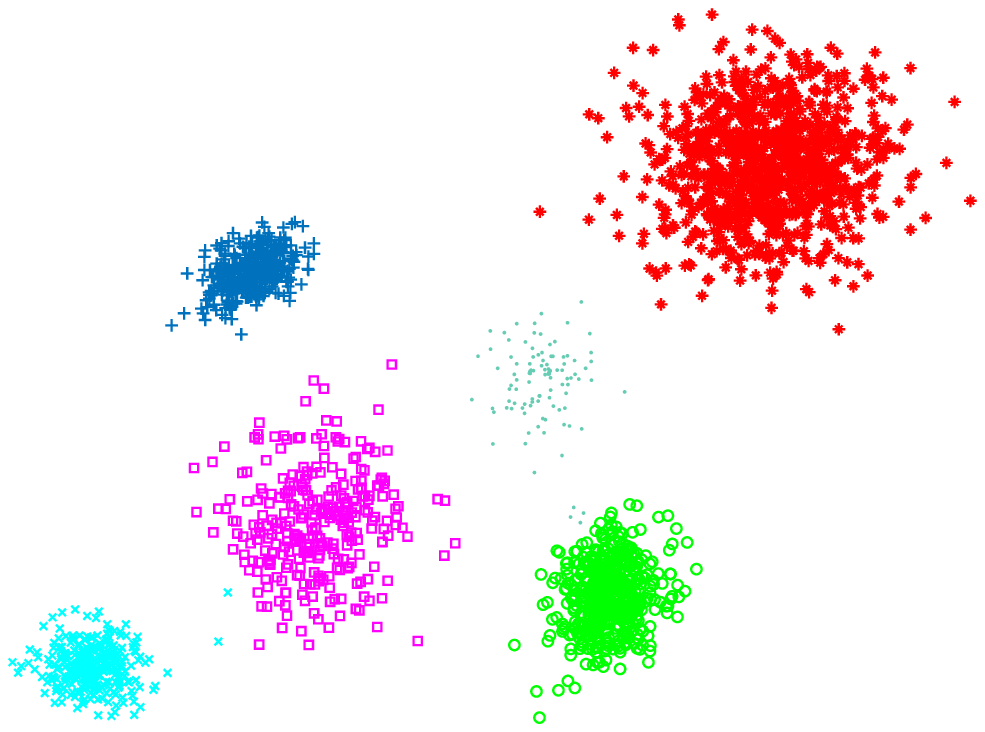}}                        \\ \cline{2-16}
                          & RI     & ARI    & FS           & RI     & ARI    & FS           & RI     & ARI    & FS          & RI     & ARI    & FS         & RI     & ARI    & FS                           \\ \cline{2-16}
                          & 0.9951 & 0.9951 & 0.9897      & 0.9323 & 0.9323 & 0.8378      & 0.8399 & 0.8399 & 0.7591     & 0.9926 & 0.9926 & 0.984     & 0.9989 & 0.9989 & 0.9978                    \\ \cline {2-16}
                          & \multicolumn{3}{c|}{RT}       & \multicolumn{3}{c|}{RT}       & \multicolumn{3}{c|}{RT}      & \multicolumn{3}{c|}{RT}     & \multicolumn{3}{c}{RT}  \\ \cline{2-16}
                          & \multicolumn{3}{c|}{0.3602}       & \multicolumn{3}{c|}{0.0273}       & \multicolumn{3}{c|}{0.1219}      & \multicolumn{3}{c|}{0.2070}     & \multicolumn{3}{c}{0.0680}  \\ \hline
\multirow{3}{*}{\makecell{Dataset2 \\ N=800 }}
                          & \multicolumn{3}{c|}{\includegraphics[width=4.5cm,height=4cm]{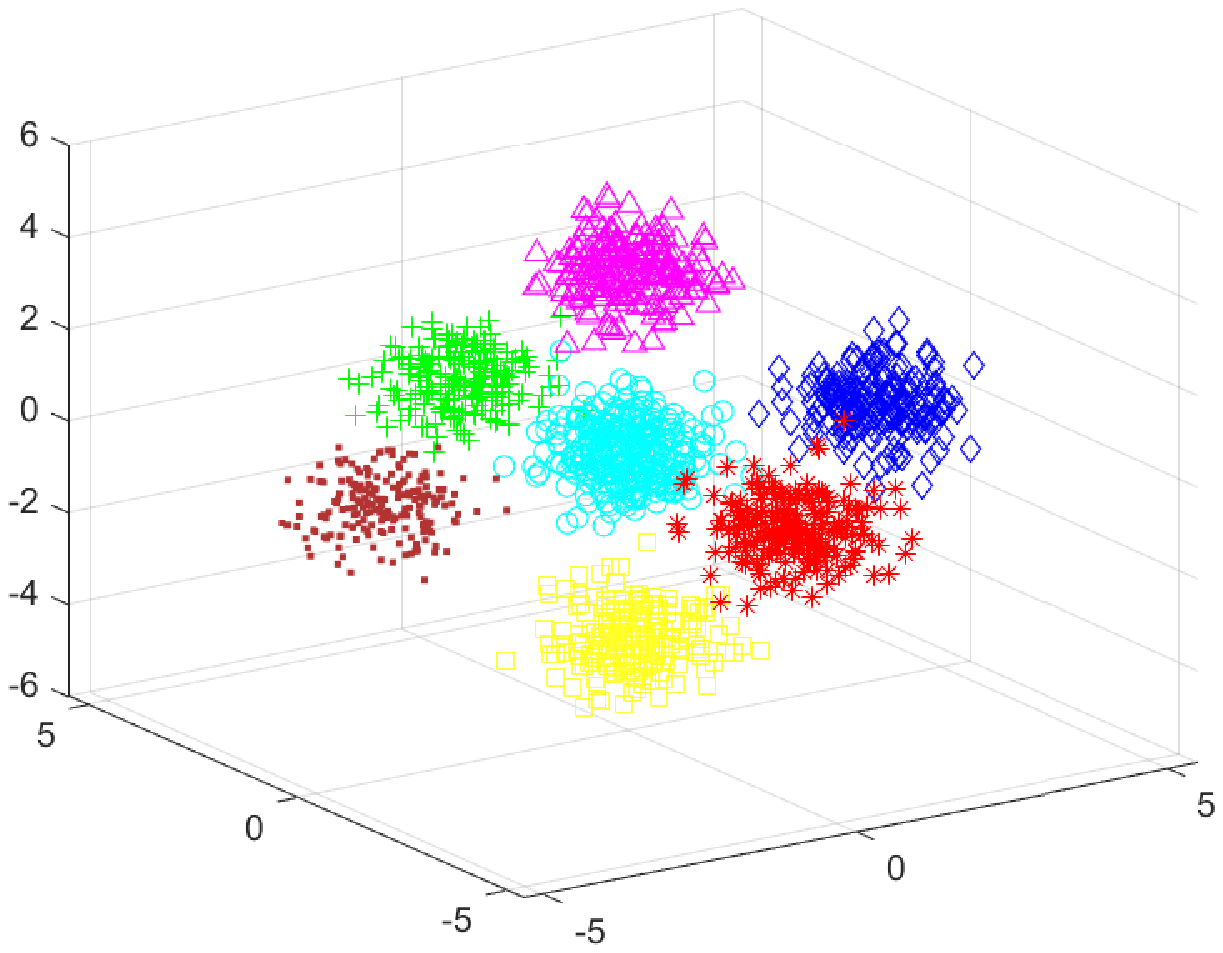}}
                          & \multicolumn{3}{c|}{\includegraphics[width=4.5cm,height=4cm]{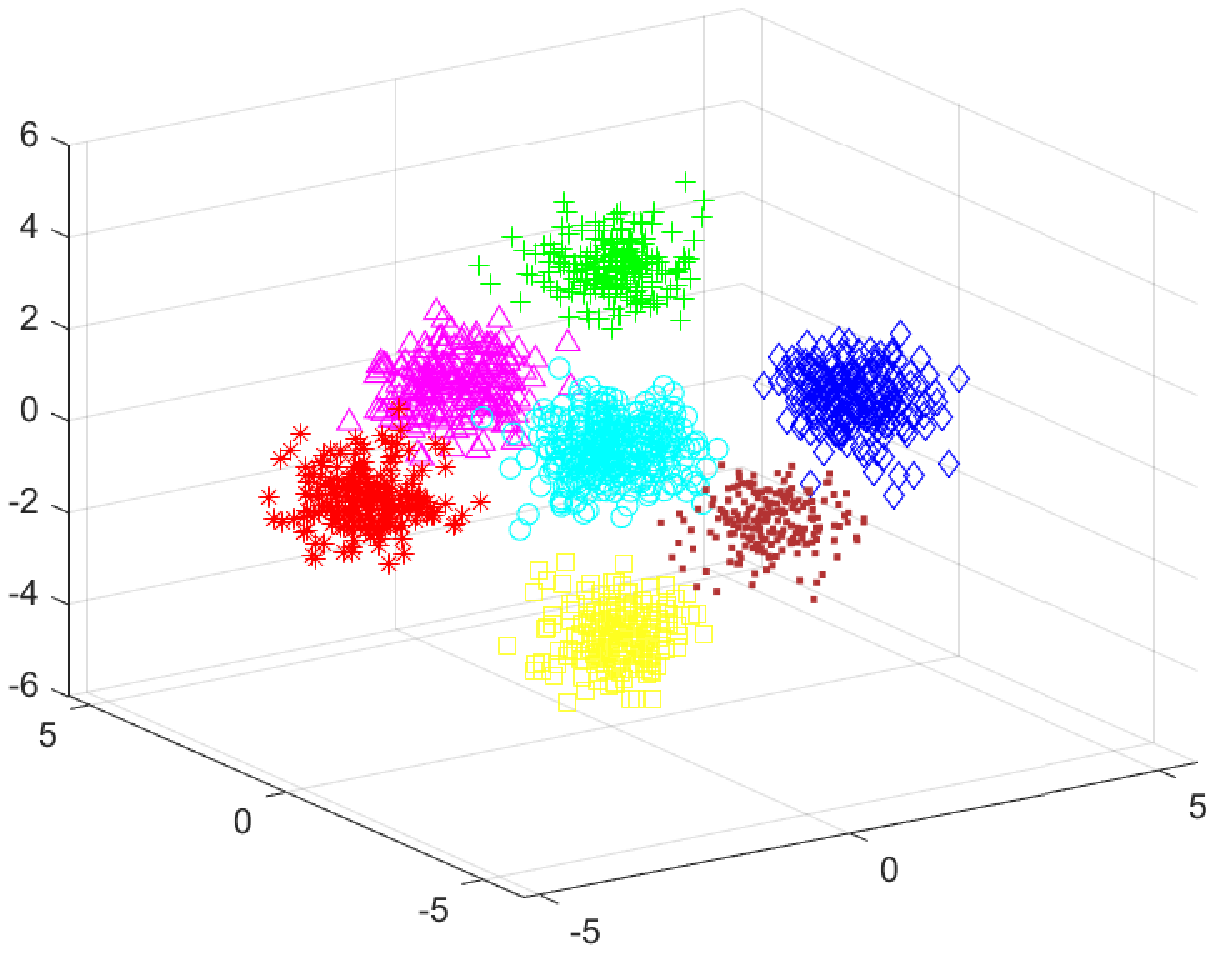}}
                          & \multicolumn{3}{c|}{\includegraphics[width=4.5cm,height=4cm]{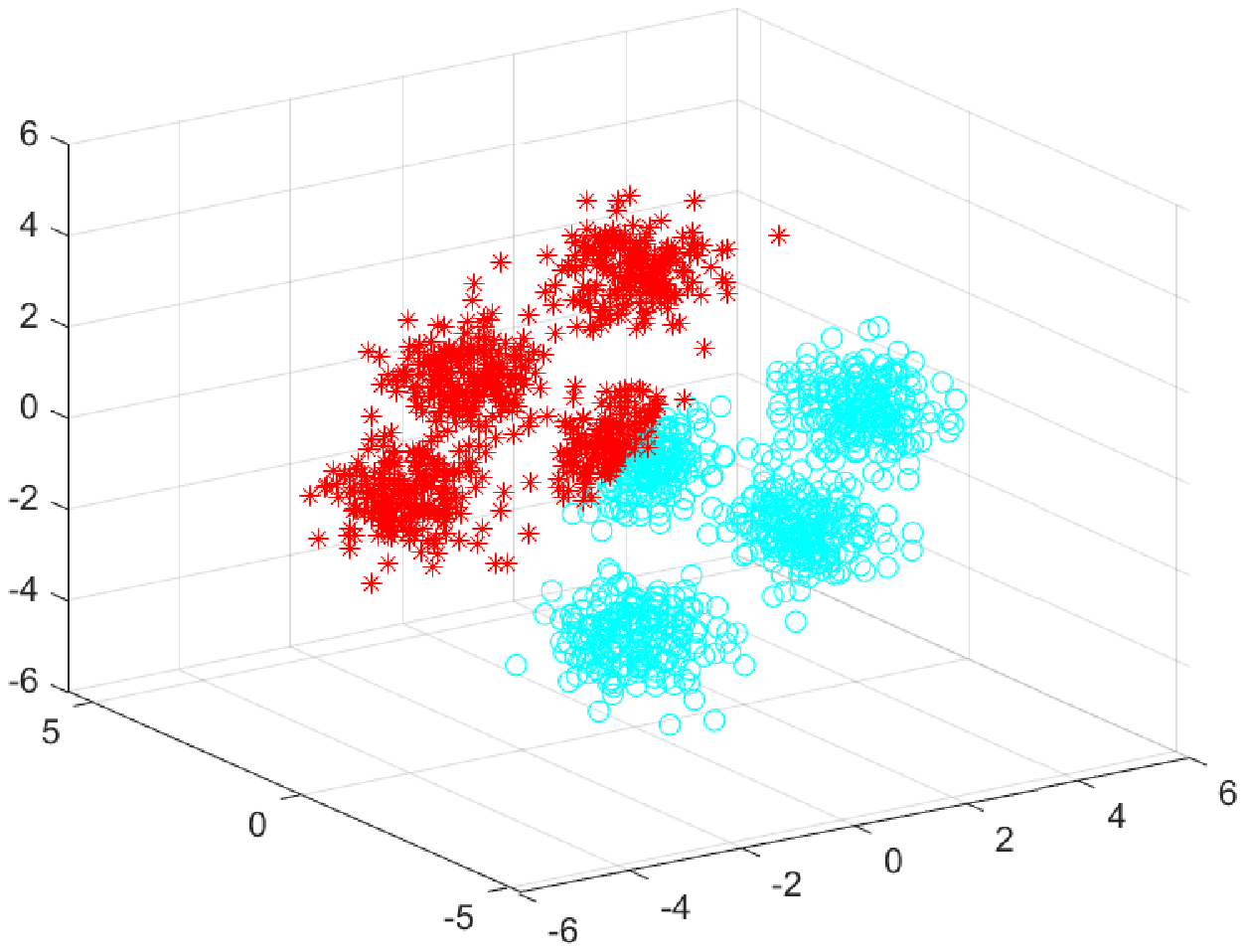}}
                          & \multicolumn{3}{c|}{\includegraphics[width=4.5cm,height=4cm]{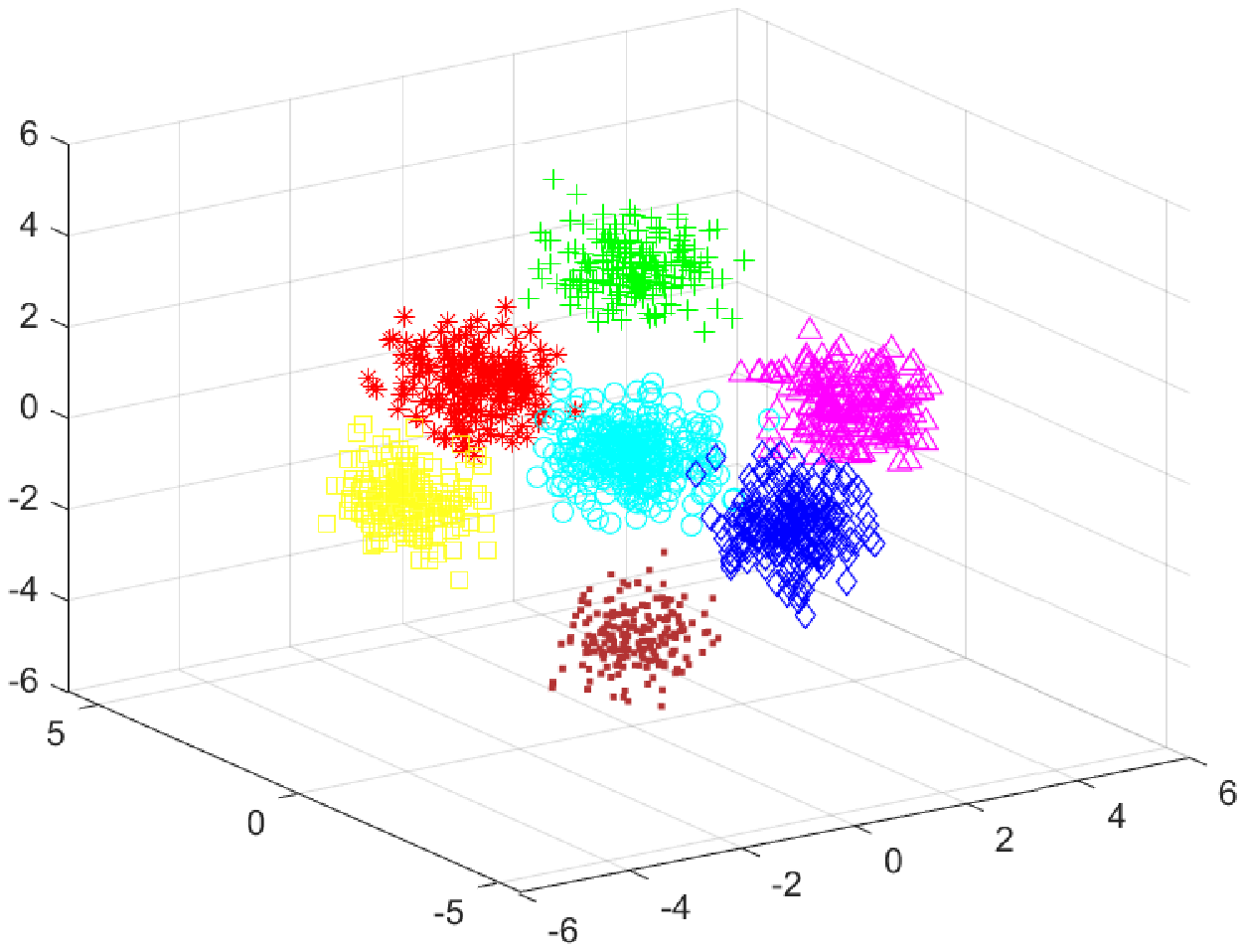}}
                          & \multicolumn{3}{c}{\includegraphics[width=4.5cm,height=4cm]{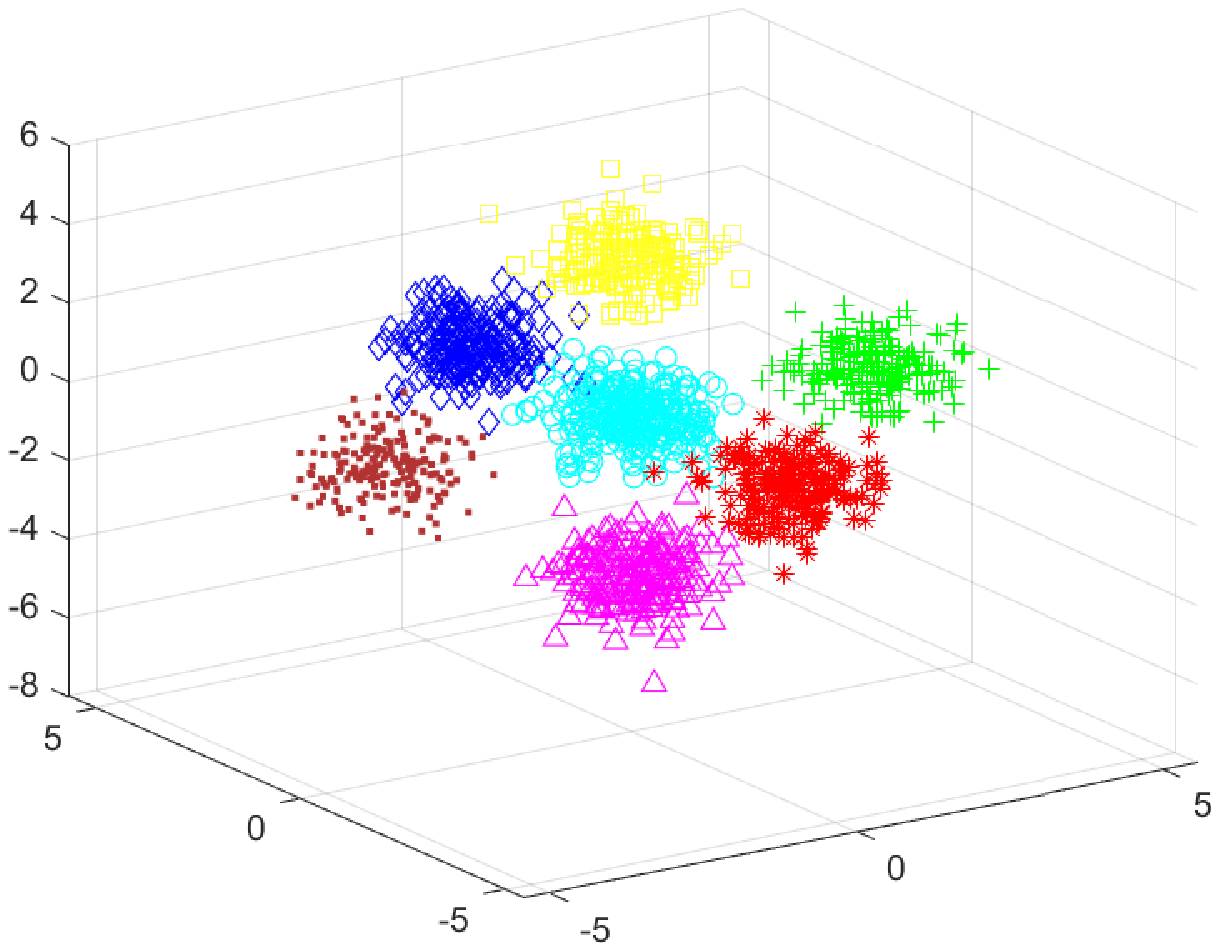}}                        \\ \cline{2-16}
                          & RI     & ARI    & FS           & RI     & ARI    & FS       & RI     & ARI    & FS          & RI     & ARI    & FS      & RI     & ARI    & FS       \\ \cline{2-16}
                          & 0.9974 & 0.9974 & 0.9909      & 0.9872 & 0.9872 & 0.9619  & 0.5498 & 0.5498 & 0.3739     & 0.9991 & 0.9991 & 0.997  & 0.9991 & 0.9991 & 0.997                  \\ \cline{2-16}
                          & \multicolumn{3}{c|}{RT}       & \multicolumn{3}{c|}{RT}       & \multicolumn{3}{c|}{RT}      & \multicolumn{3}{c|}{RT}     & \multicolumn{3}{c}{RT}  \\ \cline{2-16}
                          & \multicolumn{3}{c|}{0.3148}       & \multicolumn{3}{c|}{0.0109}   & \multicolumn{3}{c|}{0.0359}    & \multicolumn{3}{c|}{0.1398}      & \multicolumn{3}{c}{0.0250}       \\ \hline
\end{tabular}%
}
\end{table}

\subsubsection{Datasets composed of arbitrarily shaped clusters}
In this part, the Euclidean distance in the first part is replaced by minmax distance calculated with adjusted MST in the PaVa clustering algorithm. In this way, the PaVa can be applied to datasets composed of arbitrarily shaped clusters; accordingly, benchmark datasets composed of arbitrarily shaped clusters, such as \emph{twomoons}, \emph{concentric rings} (\emph{ccrings} for short), \emph{spiral}, \emph{t4.8k}, \emph{atom}, and \emph{chains}, are presented. To verify the robustness of the PaVa clustering algorithm under different types of noise, we remould the \emph{twomoons} dataset with $100$ uniformly distributed noise objects (\emph{twomoons\_noise}) and bridge-type noise (\emph{twomoons\_bridge}).

There are sufficient objects in all of these datasets, meanwhile the main bodys of clusters are dense enough.  In this way, the established PDFs of minmax distances are robust.  However, when there are less objects in a dataset, the PDF of minmax distance may suffer from a discrete pathology, and thus we do not show small datasets with no distinct dense regions, such as the dataset named \emph{flame}.

The widely used non-spherical clustering algorithms, hierarchical clustering algorithm (single-linkage), kernel k-means, DBSCAN and its recent variant DC-SKCG, spectral clustering algorithm, as well as a minmax path-based algorithm GOPC are applied to these datasets. The only parameter $k$ in the PaVa clustering algorithm is $\lceil \log N \rceil$ for all the dataset except for \emph{spiral}. Considering the particularity of the dataset \emph{spiral}, algorithm adjusting the MST by $k$-distance decreases the weight of edges between clusters in MST, so we use the original MST instead of adjusted MST. From the perspective of parameter setting, the PaVa clustering algorithm is more straightforward and more friendly to non-professional users. In addition to DBSCAN, DC-SKCG, CutPC and PaVa, the remaining algorithms require the number of clusters to be inputted as a parameter, which is directly given by the truth-value. Furthermore, the appropriate parameters, such as $\beta$ in kernel k-means, $\sigma$ in the spectral clustering algorithm, \{$minpts$, $eps$\} in DBSCAN, \{$eps$, $k$ for KNN, $l$ for the number of representatives\} in DC-SKCG, $\alpha$ in CutPC are chosen by trial-and-error, which is tedious work.

Table \ref{Table non-spherical} shows the performance of the indicated algorithms on non-spherical datasets. Most of the algorithms perform well on these non-spherical datasets, but the hierarchical clustering algorithm with single-linkage is not robust to noise. For example, the hierarchical clustering algorithm recognizes some outlier cliques as clusters, and the clusters linked by noise are recognized as a giant cluster in datasets \emph{twomoons\_noise}, \emph{twomoons\_bridge}, and \emph{t4.8k}. We increase the given number of clusters in the hierarchical clustering algorithm so that the impact of noise could be eliminated and the clusters can be divided correctly. In dataset \emph{twomoons\_noise}, the result is not reasonable until the number of clusters is set to be $17$. Similarly, the number of clusters is $12$ in dataset \emph{twomoons\_bridge} and $340$ in dataset \emph{t4.8k}. In short, the parameter in the hierarchical clustering algorithm is difficult to determine due to its significant dependence on the deconcentration of objects.
Kernel k-means performs poorly in some complex datasets, even if we tried to find the most suitable parameters for $\beta$.
DBSCAN achieves correct clustering results in most of the dataset except \emph{spiral}. This is because the farther away from the center, the sparser the objects scatter. However, the global parameters $minpts$ and $eps$ cannot match heterogeneous local density; for this reason, the tails are recognized as noise. Albeit with adaptive neighbor region for every object, DC-SKCG similarly fails to extend the main body of clusters to their tails. Moreover, it also fails on the most complex dataset \emph{t4.8k}. The spectral clustering algorithm recognizes all the main bodies of clusters. However, a flaw occurs on uniformly distributed noise, such as twonoons\_noise and \emph{t4.8k}; the affiliations of noise are disordered. The path-based algorithm, GOPC, also gets good clustering results on all the datasets due to its suitable distance measurement. When appropriate parameters are selected, the CutPC algorithm produces reasonable results. However, it has a problem of recognizing points on the class body as noise points.

As for the running time (RT) shown in Table \ref{Table non-spherical}, the PaVa clustering algorithm runs in an acceptable amount of time. The PaVa clustering algorithm is slower than the hierarchical clustering algorithm in dataset \emph{twomoons}, \emph{Concentric ring}, \emph{t4.8k}, and \emph{atom}. On the remaining datasets, the PaVa clustering algorithm is only slightly slower than the hierarchical clustering algorithm, DBSCAN and DC-SKCG. That is, our algorithm is much more efficient than DBSCAN and DC-SKCG when the datasets increase in size.
Kernel k-means updates the affiliation of objects in every iteration with a time complexity $O(N^2)$; in other words, for the update of one object, all objects are involved. Eigenvalue decomposition is the most time-consuming part of the spectral clustering algorithm with the time complexity of $O(N^2 \log N)$, with running times increasing dramatically the datasets increase in size. Similarly, the GOPC calculates all the pairwise minmax distances; namely, the adjacent matrix is necessary, with total time complexity $O(N^2* N)$. The running time of CutPC is highly correlated with the distribution of data items in the space. If there is a high degree of separation between clusters, the time taken to construct the natural adjacent graph will be shorter. However, if the separation between clusters is not obvious, or if there is noise throughout the space (as in dataset \emph{t4.8k}), the time taken to construct the natural adjacent graph will be longer.
Consequently, its runtime is similar to the PaVa clustering algorithm on small-size datasets, but considerably larger on large-size datasets.

Considering both the accuracy and time, our algorithm can obtain accurate results in an acceptable time while maintaining consistent parameter input. To this end, our algorithm is effective, efficient and user-friendly.

% \usepackage{multirow}  %{0.97\linewidth}{\textheight-2.7cm}
\begin{table}
\caption{Accuracy of the indicated algorithms and non-spherical datasets.}\label{Table non-spherical}
\centering
\ContinuedFloat
\Huge
\resizebox{0.9\linewidth}{0.65\linewidth}{%
\Huge
\begin{tabular}{c|cccc|cccc|cccc|cccc}
\hline
\diagbox{Methods}{Performance}{Dataset}    & \multicolumn{4}{c|}{\emph{twomoons} N=600} & \multicolumn{4}{c|}{\emph{twomoons\_noise} N=700} & \multicolumn{4}{c|}{\emph{twomoons\_bridge} N=620} & \multicolumn{4}{c}{\emph{ccrings} N=6000} \\ \hline
\multirow{3}{*}{PaVa}           & \multicolumn{4}{c|}{\includegraphics[width=8cm,height=7cm]{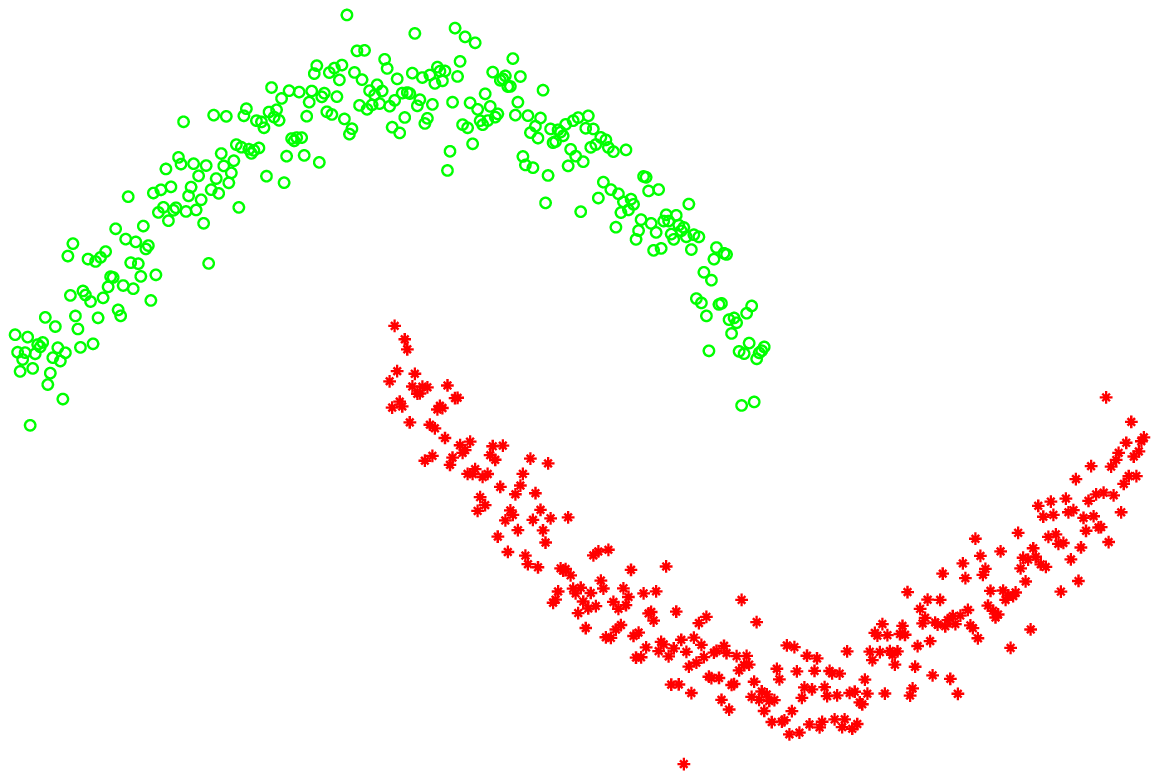}}
                                & \multicolumn{4}{c|}{\includegraphics[width=8cm,height=7cm]{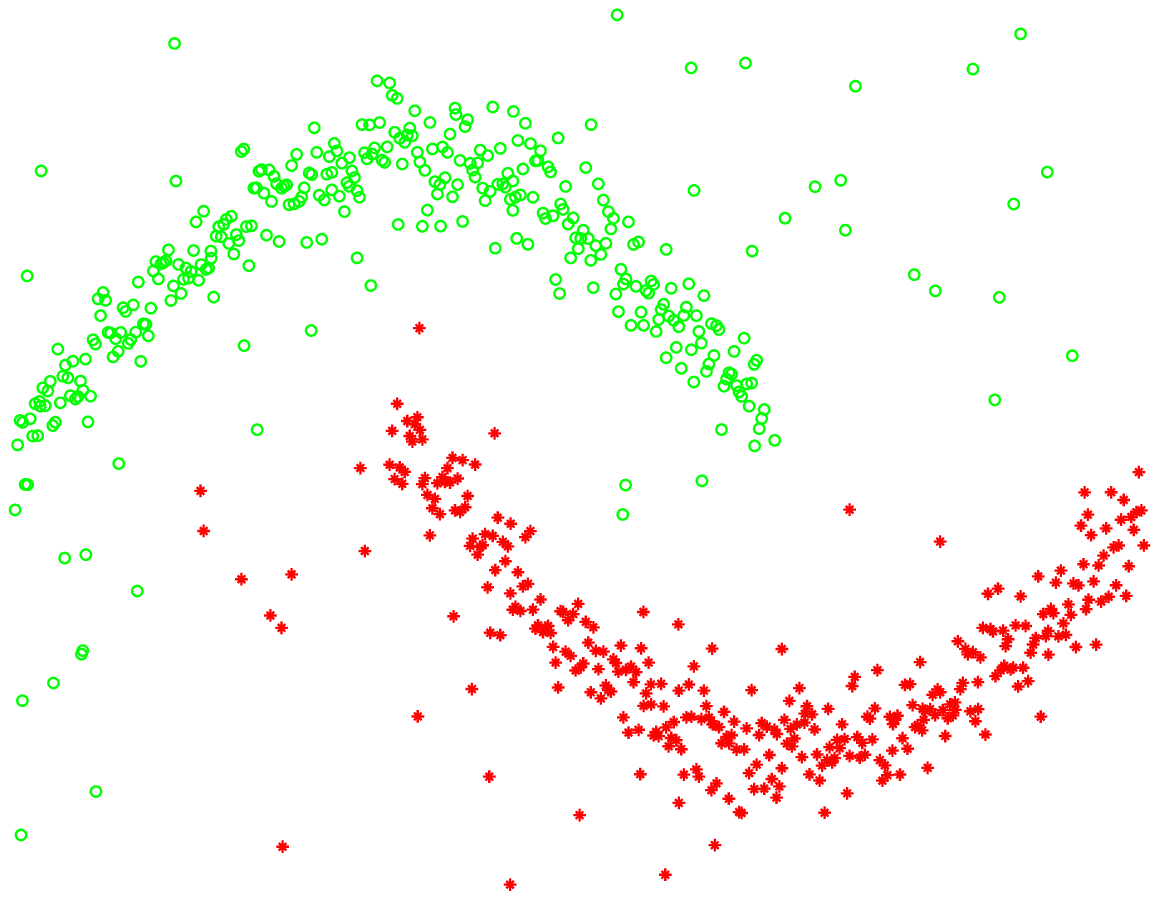}}
                                & \multicolumn{4}{c|}{\includegraphics[width=8cm,height=7cm]{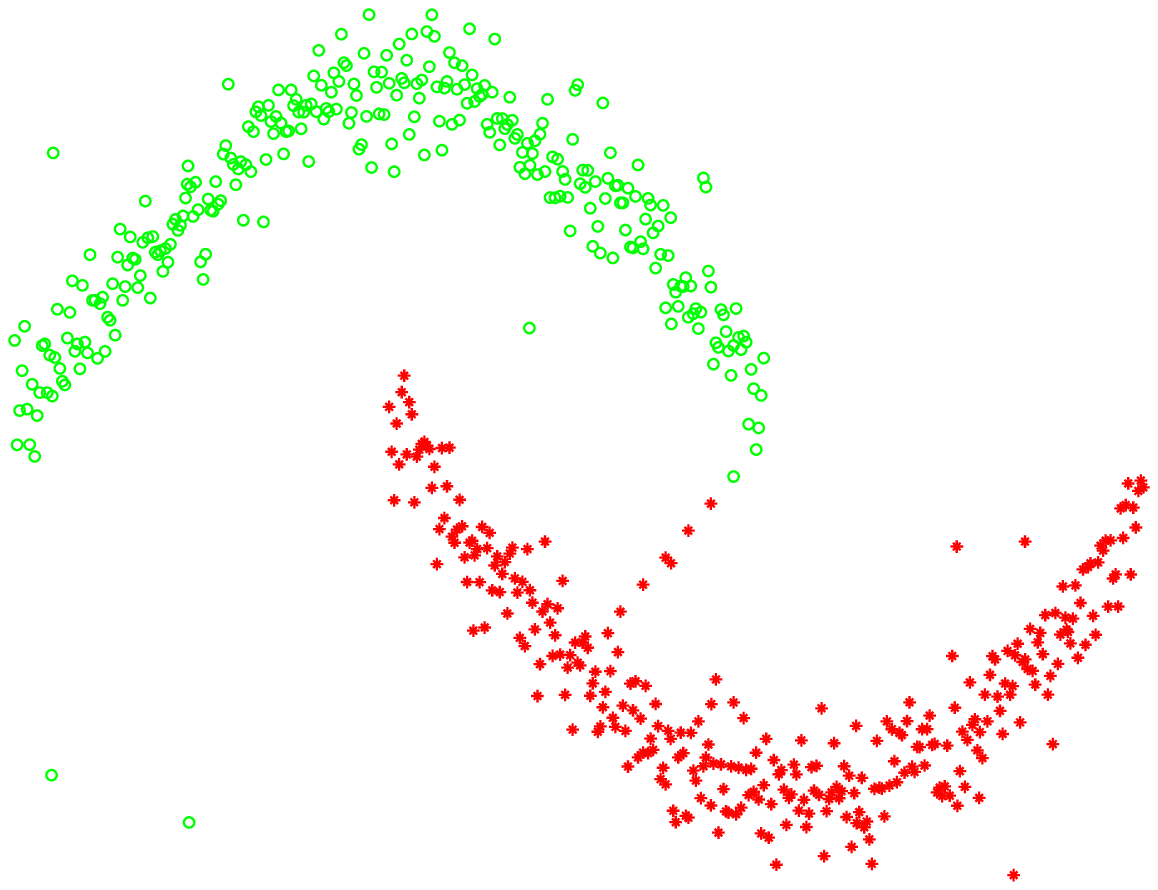}}
                                & \multicolumn{4}{c}{\includegraphics[width=8cm,height=7cm]{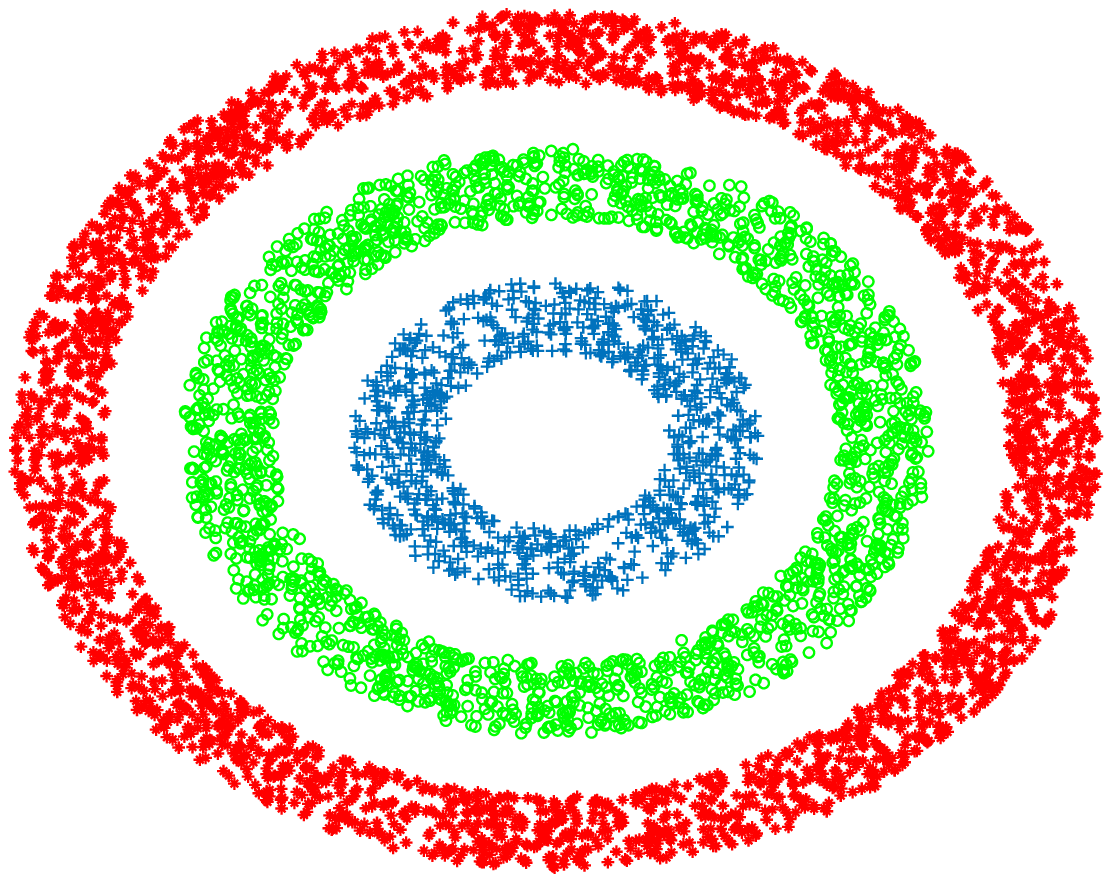}}                                \\ \cline{2-17}
                                & RI     & ARI     & \multicolumn{1}{c|}{FS}            &RT       & RI        & ARI      & \multicolumn{1}{c|}{FS}       & RT
                                & RI     & ARI     & \multicolumn{1}{c|}{FS}            &RT       & RI        & ARI      & \multicolumn{1}{c|}{FS}        & RT       \\ \cline{2-17}
                                & 1      & 1       &  \multicolumn{1}{c|}{1}           &0.0984   & 0.8672    & 1        & \multicolumn{1}{c|}{0.8501}  & 0.1102
                                & 0.9651 & 0.9969  & \multicolumn{1}{c|}{0.9639}       & 0.1172   & 1         & 1       & \multicolumn{1}{c|}{1}       & 0.8641    \\ \hline
\multirow{3}{*}{Hierarchical}   & \multicolumn{4}{c|}{\includegraphics[width=8cm,height=7cm]{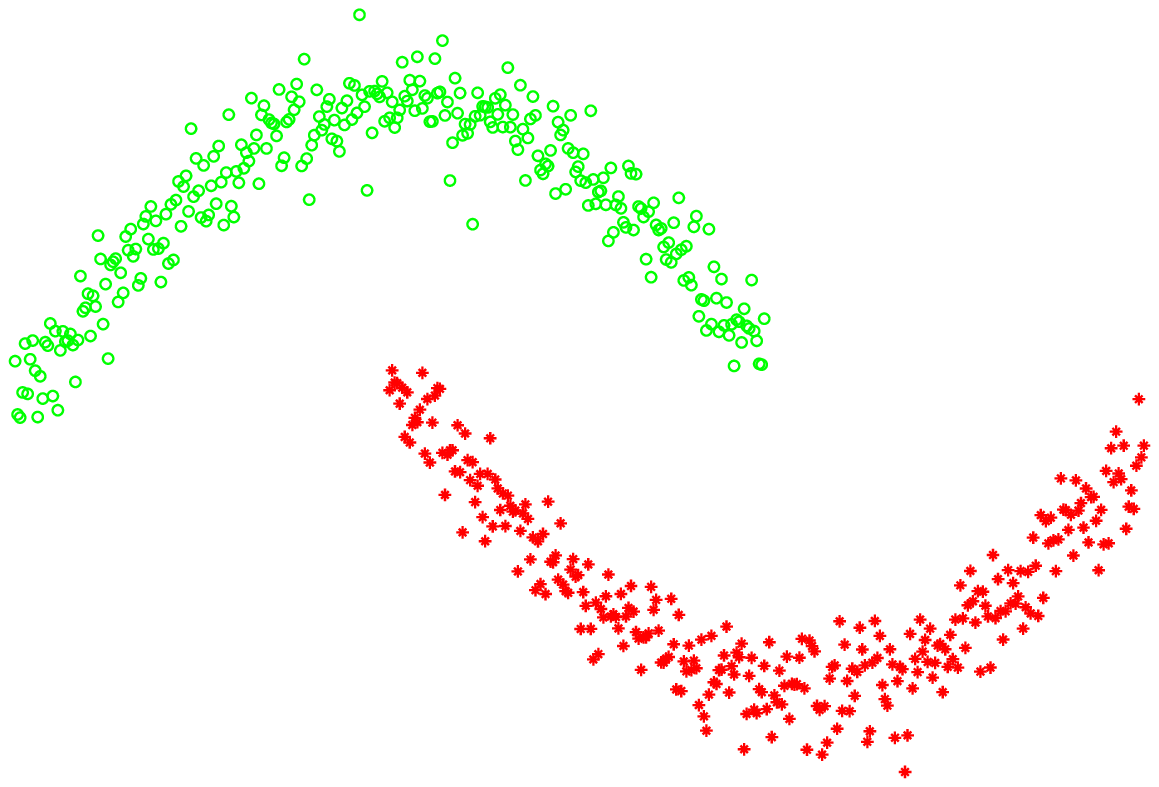}}
                                & \multicolumn{4}{c|}{\includegraphics[width=8cm,height=7cm]{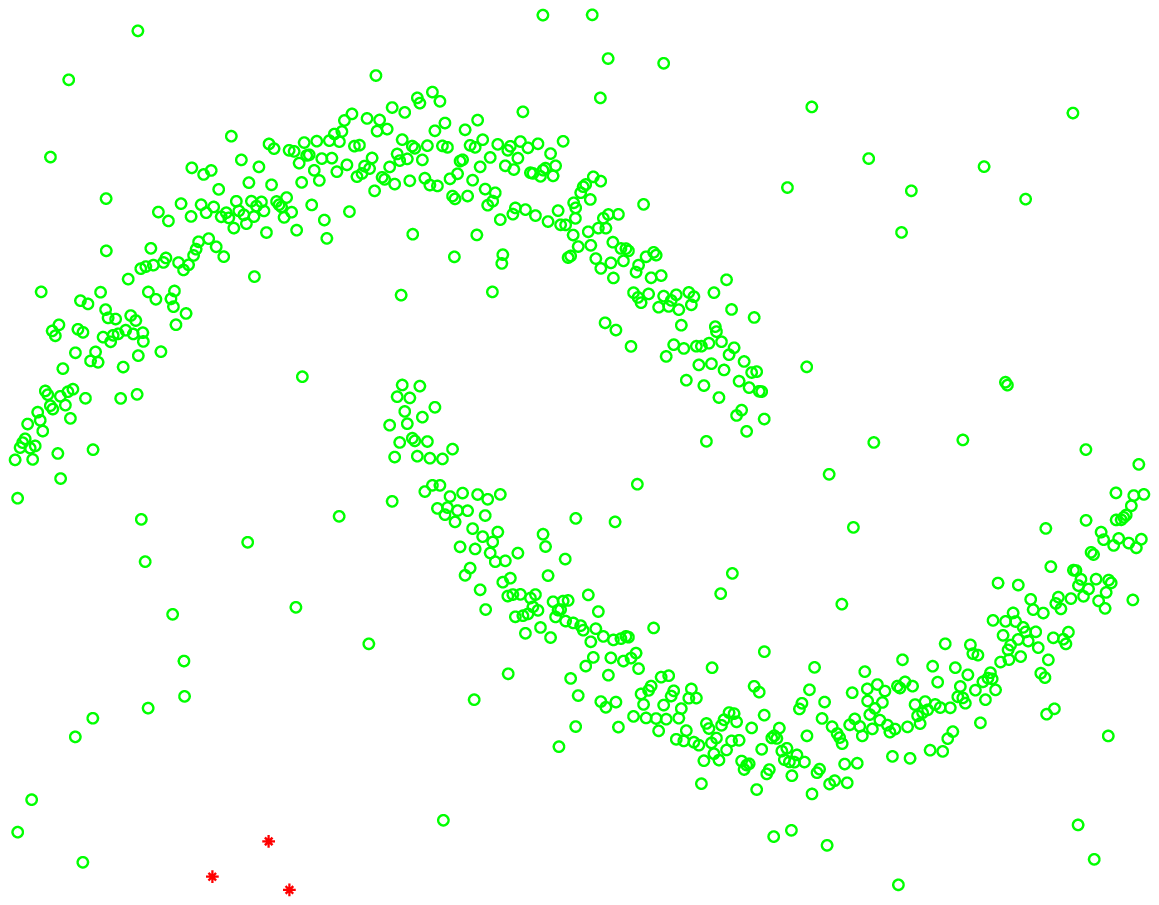}}
                                & \multicolumn{4}{c|}{\includegraphics[width=8cm,height=7cm]{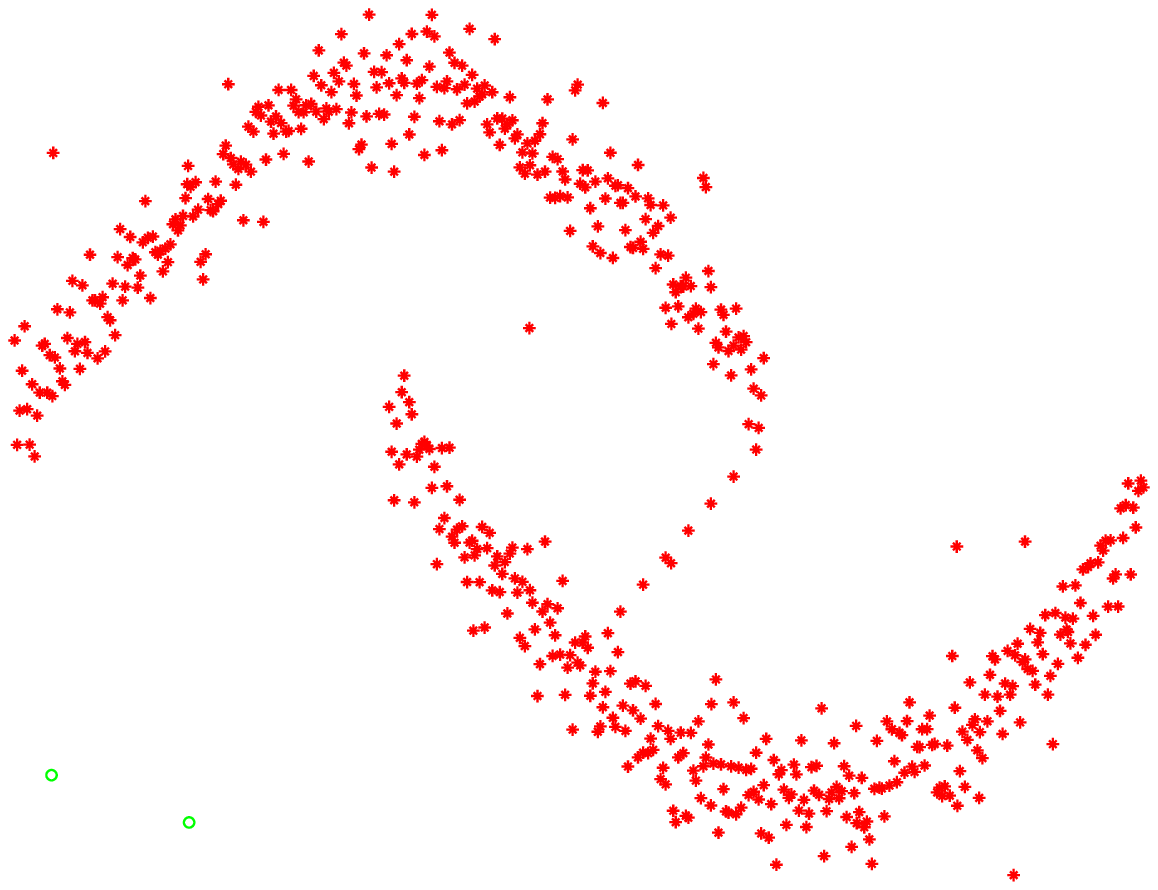}}
                                & \multicolumn{4}{c}{\includegraphics[width=8cm,height=7cm]{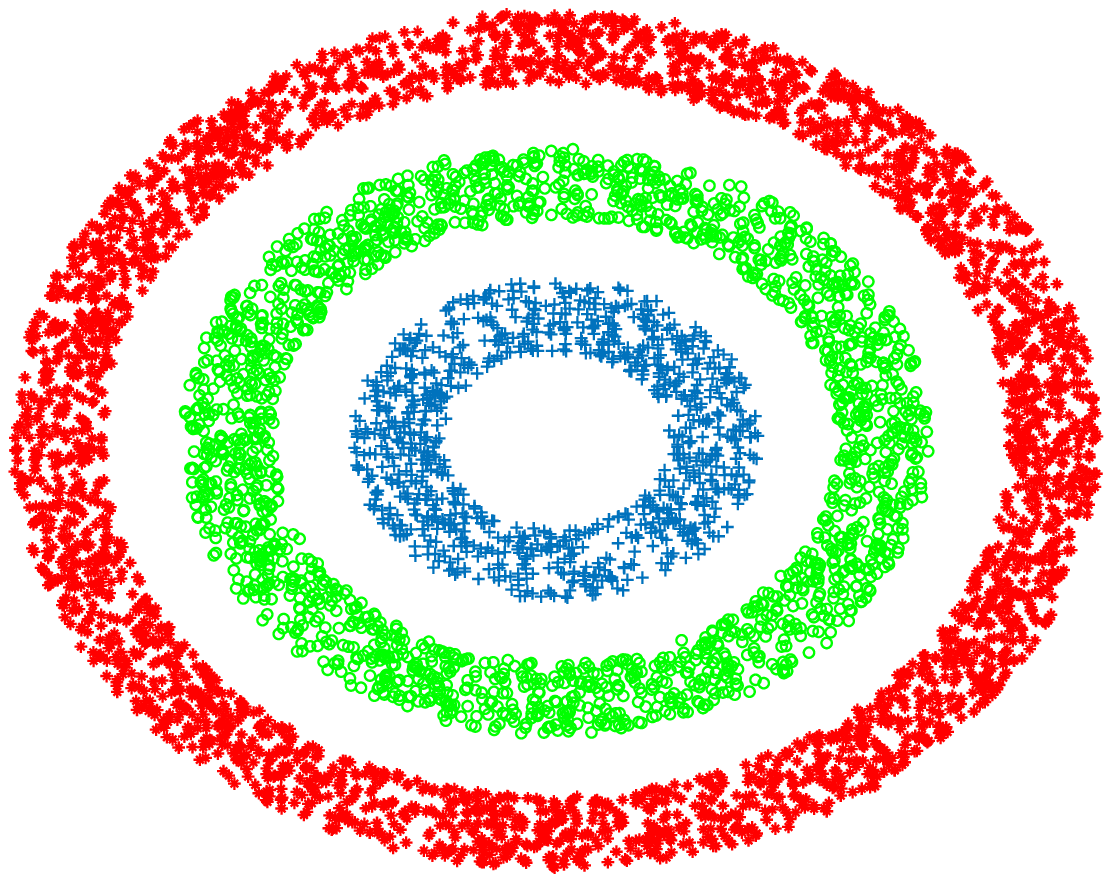}}                                    \\ \cline{2-17}
                                & RI      & ARI     & \multicolumn{1}{c|}{FS}           & RT        & RI        & ARI      & \multicolumn{1}{c|}{FS}        & RT
                                & RI      & ARI     & \multicolumn{1}{c|}{FS}           & RT        & RI        & ARI      & \multicolumn{1}{c|}{FS}        & RT         \\ \cline{2-17}
                                & 1       & 1       & \multicolumn{1}{c|}{1}          & 0.0117    & 0.3889    & 0.6322   & \multicolumn{1}{c|}{0.5585}  & 0.0109
                                & 0.4745  & 0.5310  & \multicolumn{1}{c|}{0.6406}     & 0.0070    & 1         & 1        & \multicolumn{1}{c|}{1}       & 0.8109      \\ \hline
\multirow{3}{*}{kernel k-means} & \multicolumn{4}{c|}{\includegraphics[width=8cm,height=7cm]{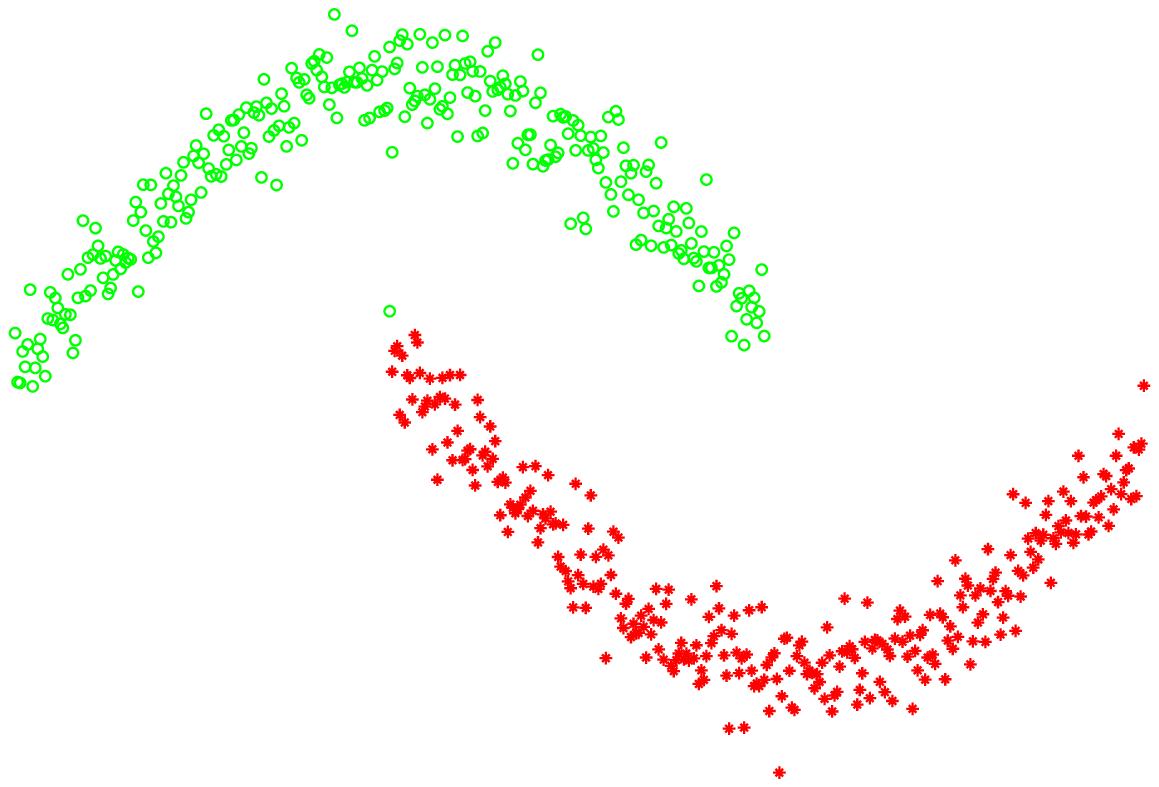}}
                                & \multicolumn{4}{c|}{\includegraphics[width=8cm,height=7cm]{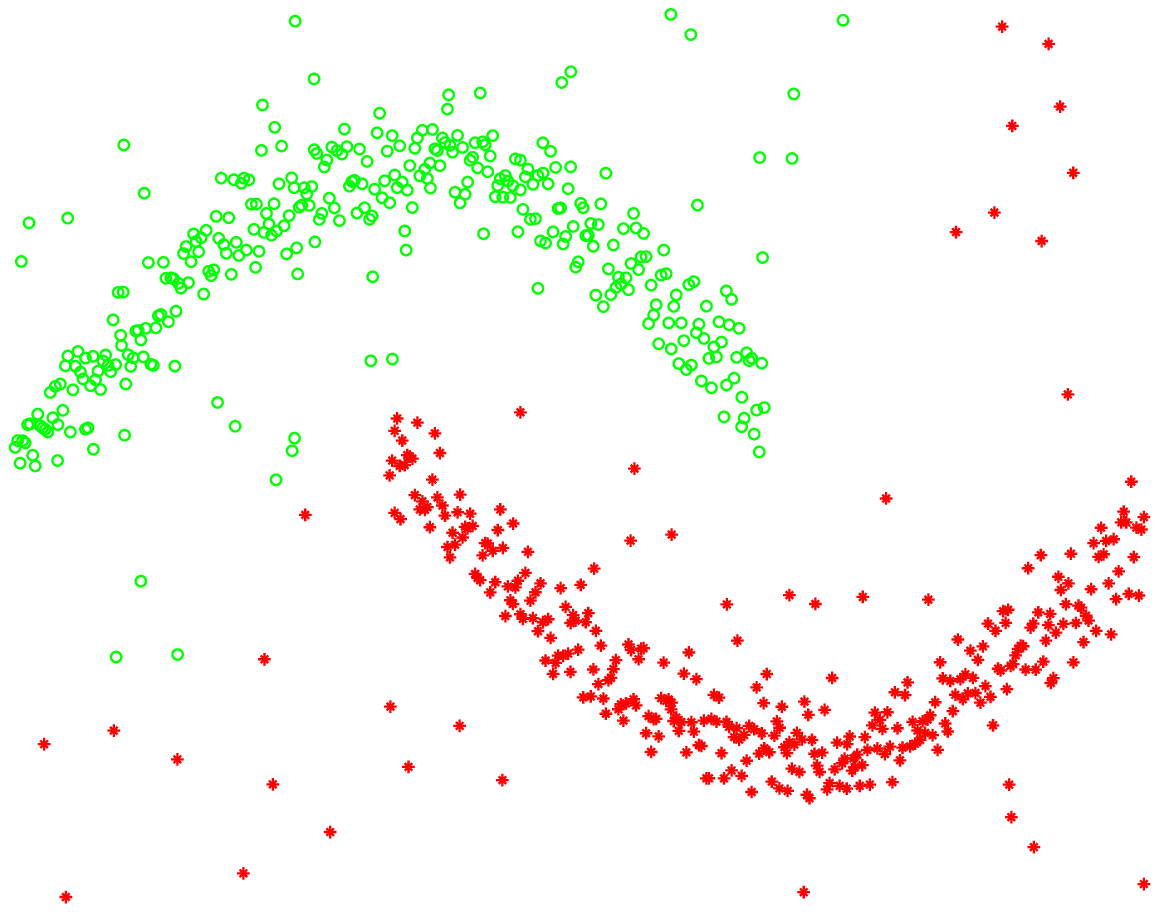}}
                                & \multicolumn{4}{c|}{\includegraphics[width=8cm,height=7cm]{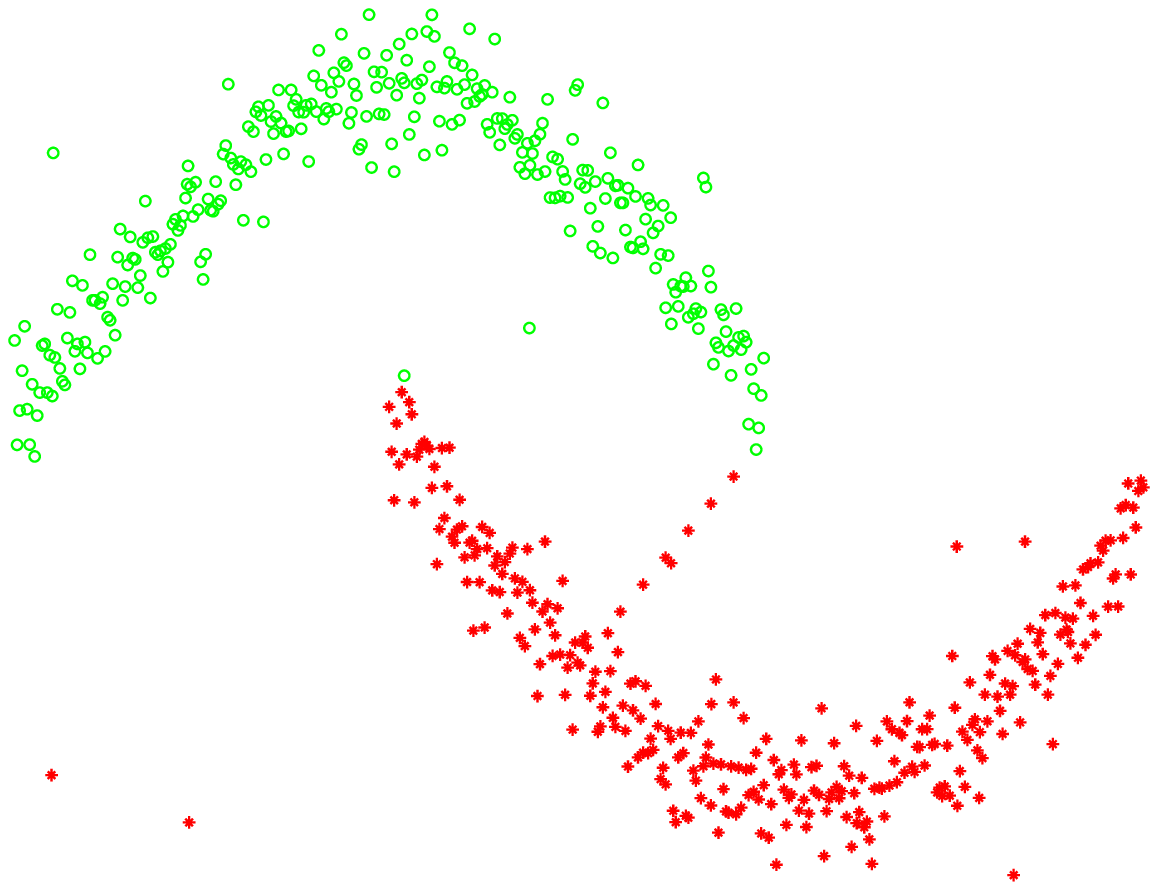}}
                                & \multicolumn{4}{c}{\includegraphics[width=8cm,height=7cm]{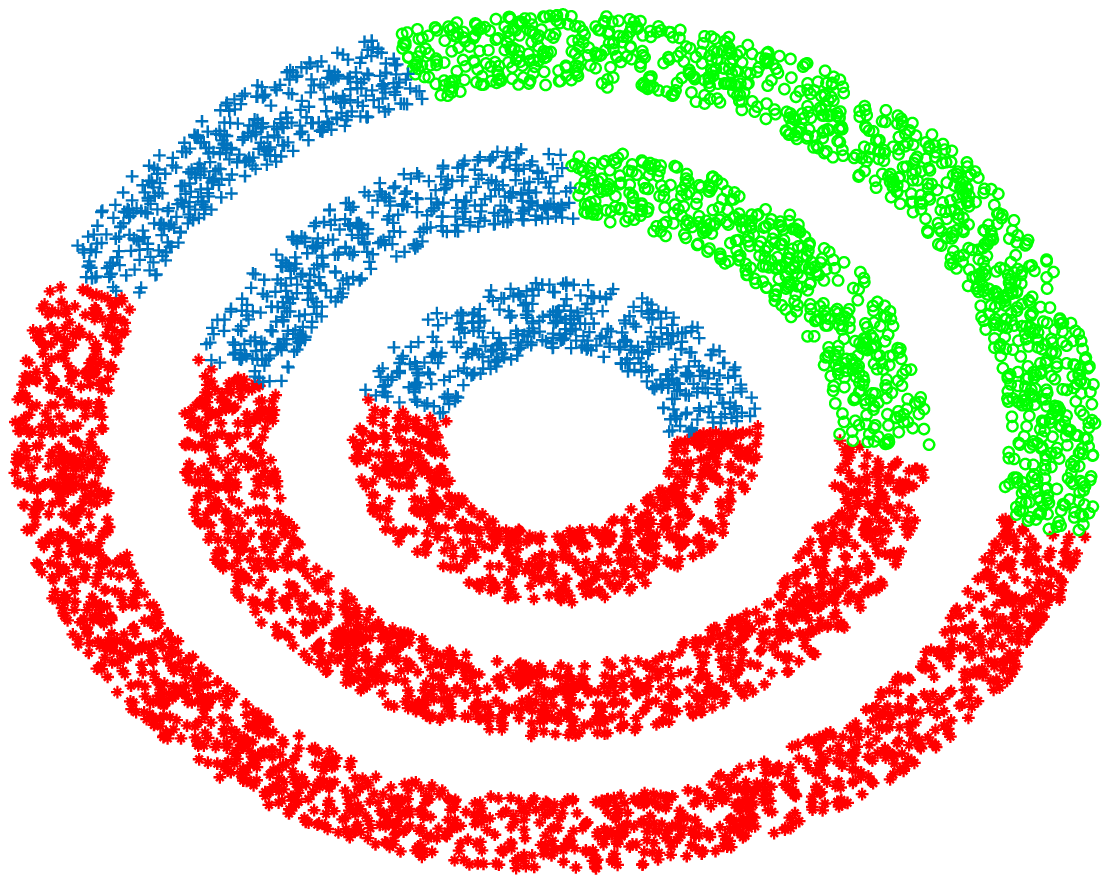}}               \\ \cline{2-17}
                                & RI      & ARI     & \multicolumn{1}{c|}{FS}          & RT        & RI         & ARI      & \multicolumn{1}{c|}{FS}        & RT
                                & RI      & ARI     & \multicolumn{1}{c|}{FS}          & RT        & RI         & ARI      & \multicolumn{1}{c|}{FS}        & RT    \\ \cline{2-17}
                                & 0.8961  & 0.8961  & \multicolumn{1}{c|}{0.8980}    & 0.3570    & 0.7783     & 0.9170   & \multicolumn{1}{c|}{0.7566}  & 0.4789
                                & 0.8417  & 0.8742  & \multicolumn{1}{c|}{0.8409}    & 0.4016    & 0.5311     & 0.5311   & \multicolumn{1}{c|}{0.3869}  & 38.363  \\ \hline
\multirow{3}{*}{DBSCAN}         & \multicolumn{4}{c|}{\includegraphics[width=8cm,height=7cm]{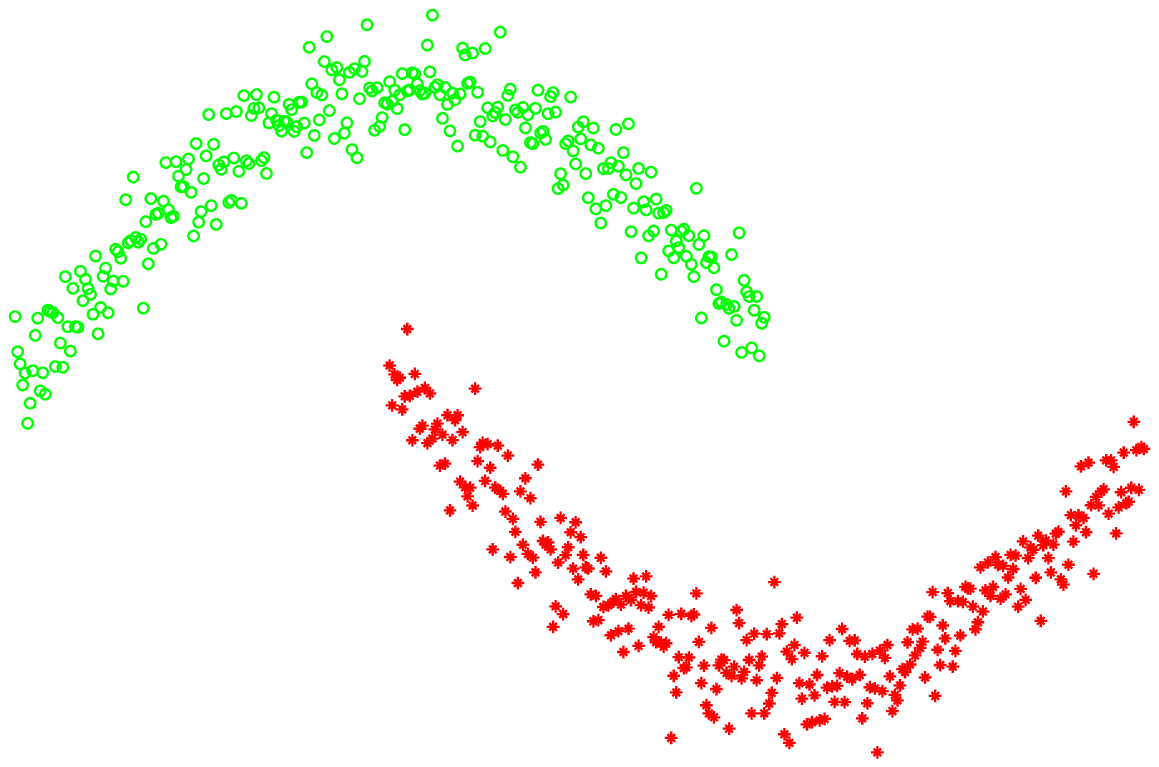}}
                                & \multicolumn{4}{c|}{\includegraphics[width=8cm,height=7cm]{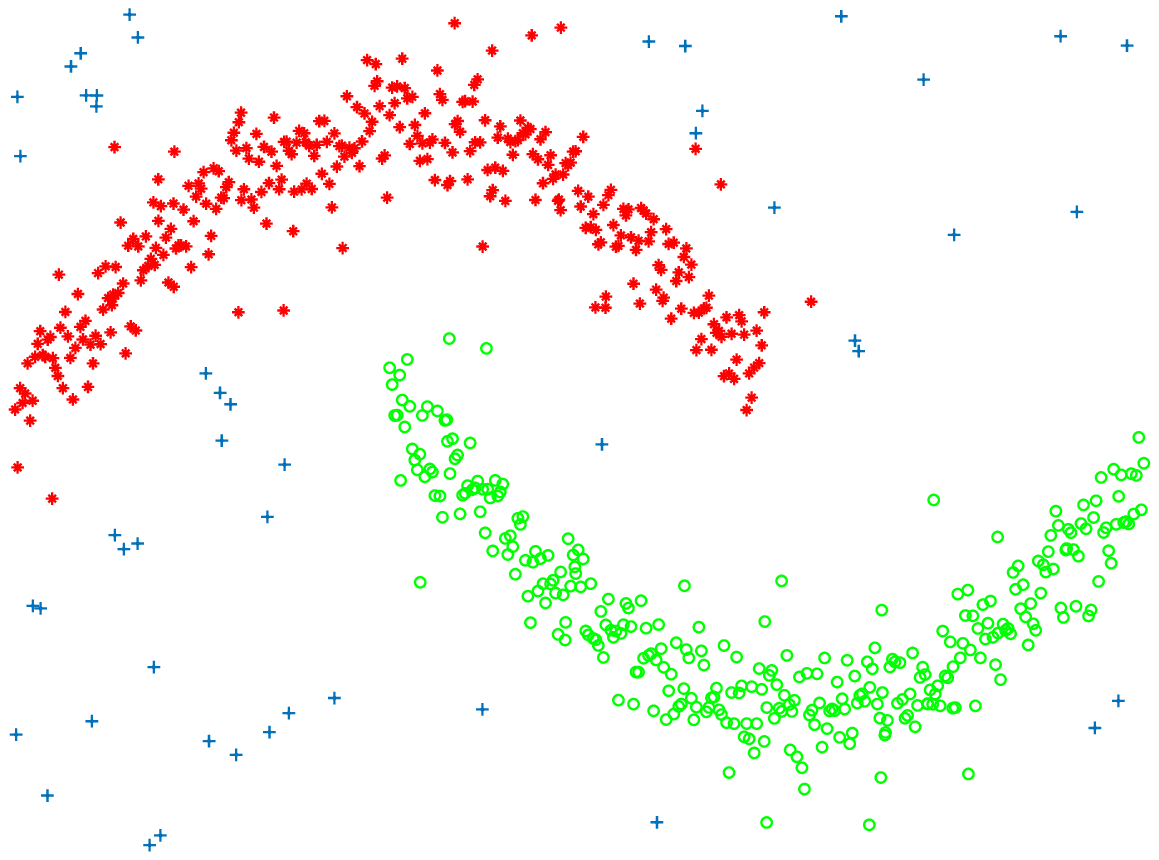}}
                                & \multicolumn{4}{c|}{\includegraphics[width=8cm,height=7cm]{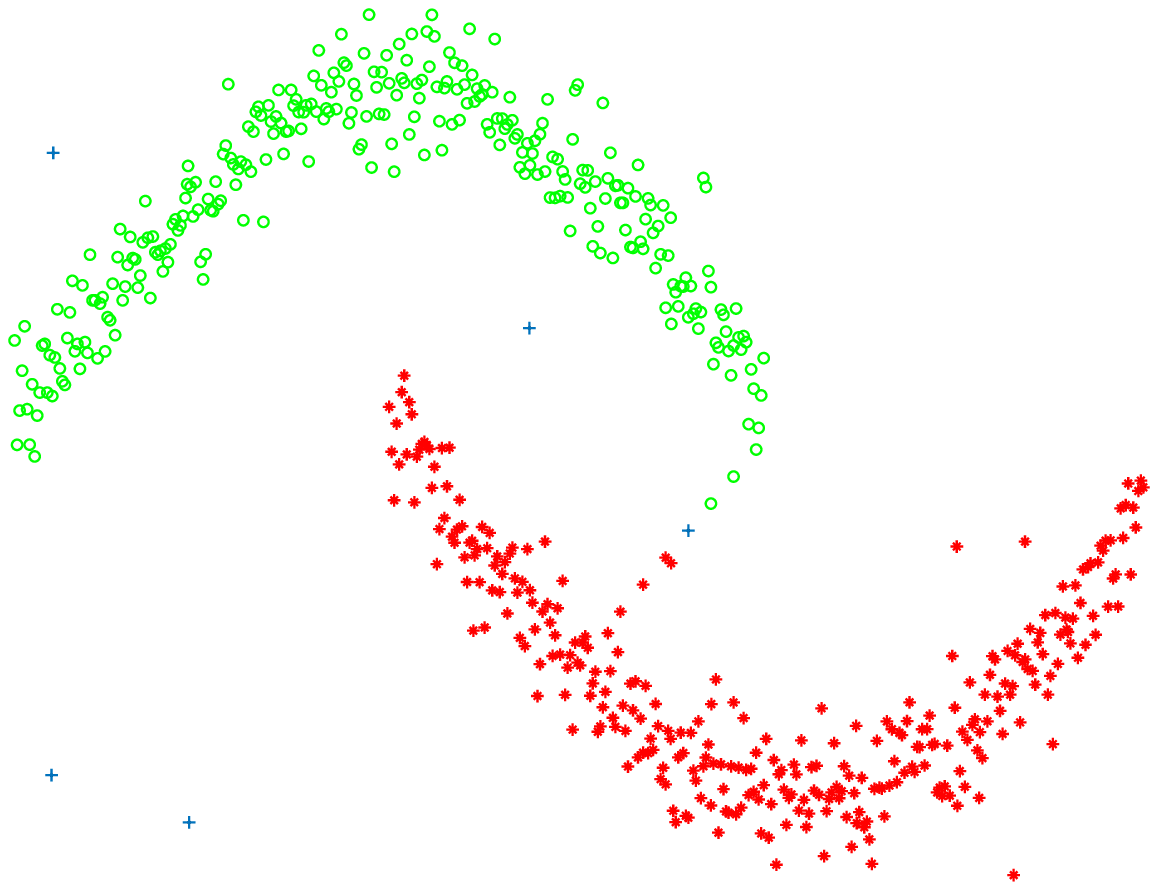}}
                                & \multicolumn{4}{c}{\includegraphics[width=8cm,height=7cm]{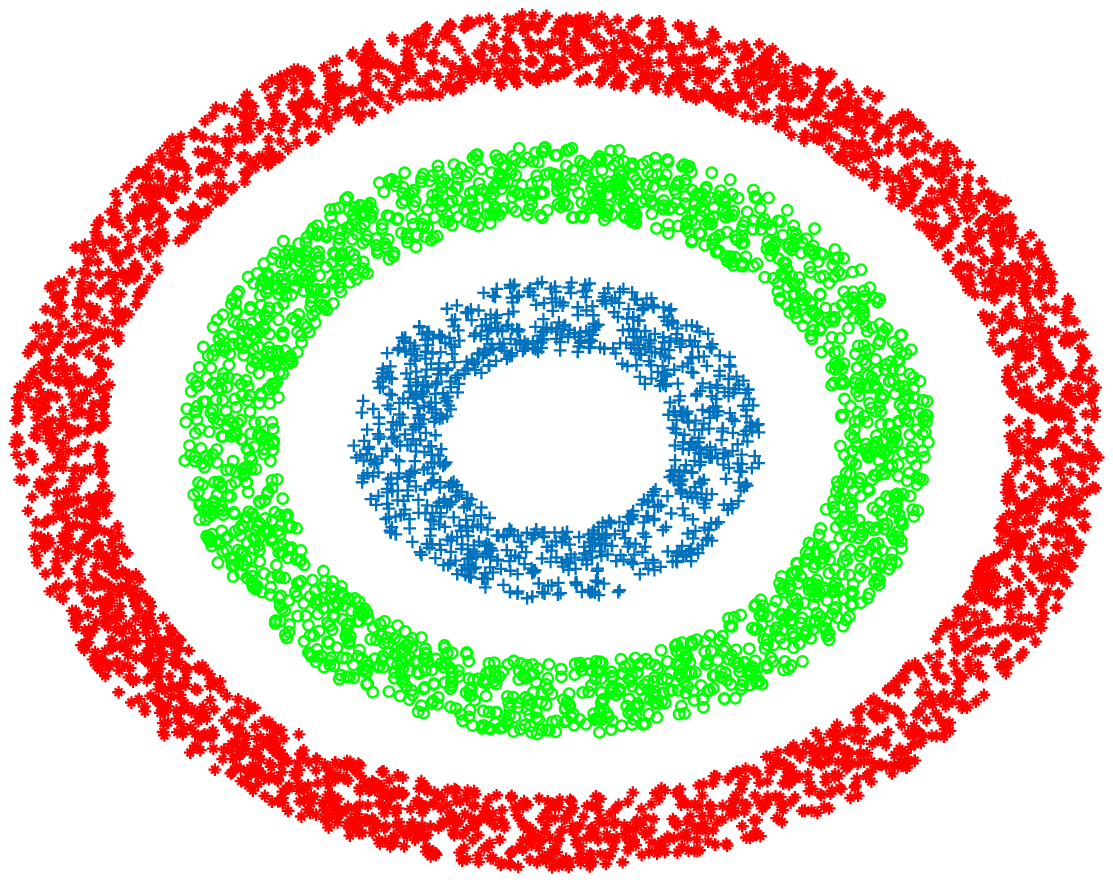}}               \\ \cline{2-17}
                                & RI      & ARI     & \multicolumn{1}{c|}{FS}          & RT        & RI         & ARI      & \multicolumn{1}{c|}{FS}        & RT
                                & RI      & ARI     & \multicolumn{1}{c|}{FS}          & RT        & RI         & ARI      & \multicolumn{1}{c|}{FS}        & RT      \\ \cline{2-17}
                                & 1       & 1       & \multicolumn{1}{c|}{1}         & 0.1000    & 0.9247     & 1        & \multicolumn{1}{c|}{0.9085}  & 0.0594
                                & 0.9728  & 0.9969  & \multicolumn{1}{c|}{0.9716}    & 0.0445    & 1          & 1        & \multicolumn{1}{c|}{1}       & 3.9617   \\ \hline
\multirow{3}{*}{DC-SKCG}        & \multicolumn{4}{c|}{\includegraphics[width=8cm,height=7cm]{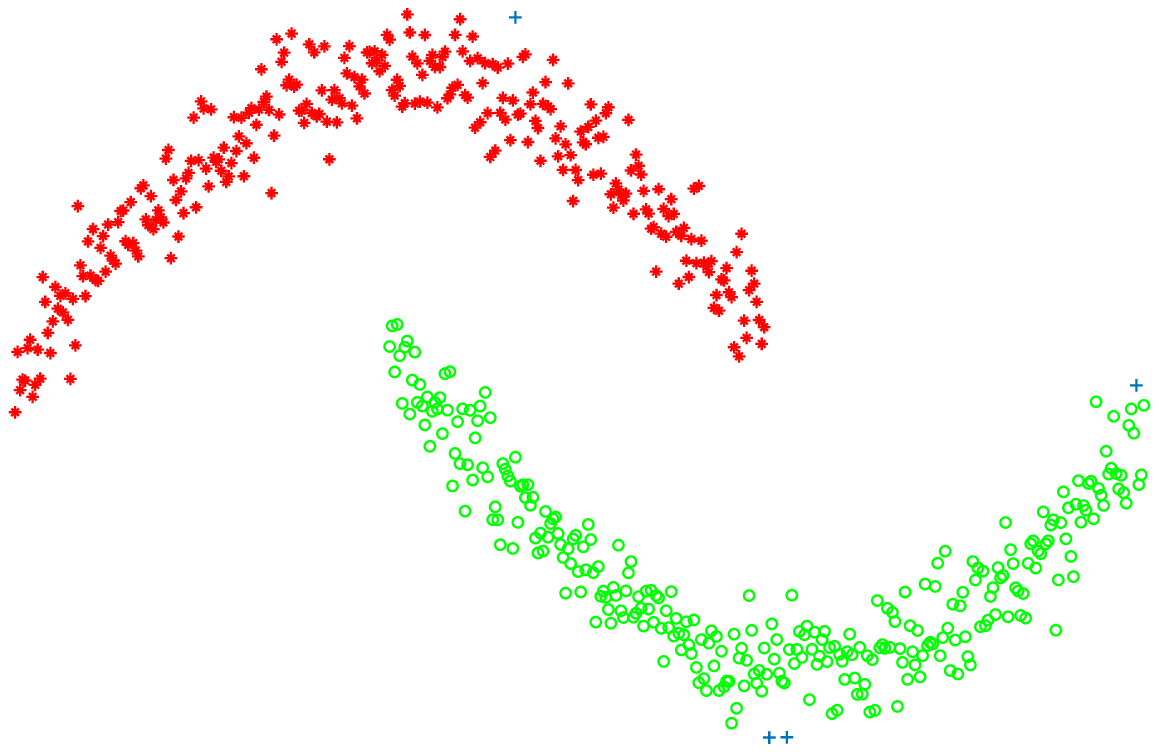}}
                                & \multicolumn{4}{c|}{\includegraphics[width=8cm,height=7cm]{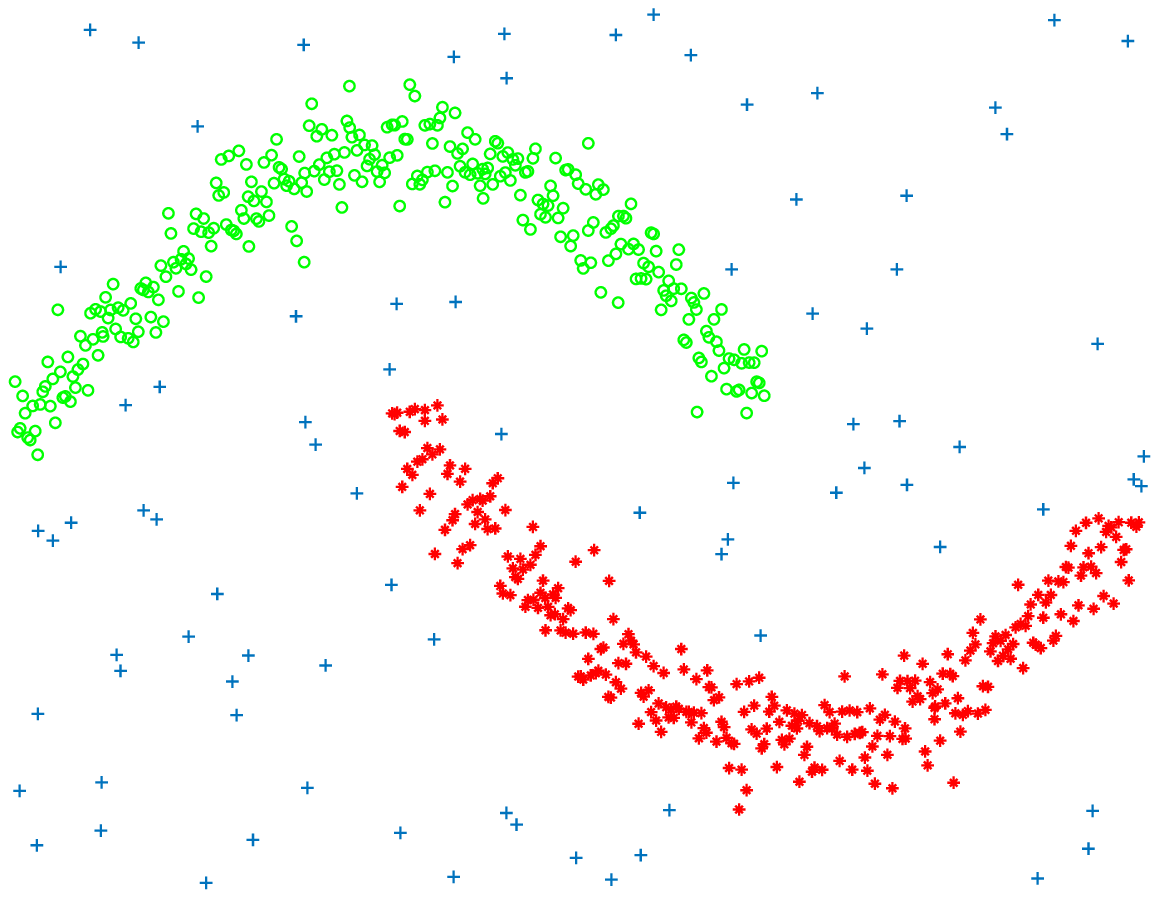}}
                                & \multicolumn{4}{c|}{\includegraphics[width=8cm,height=7cm]{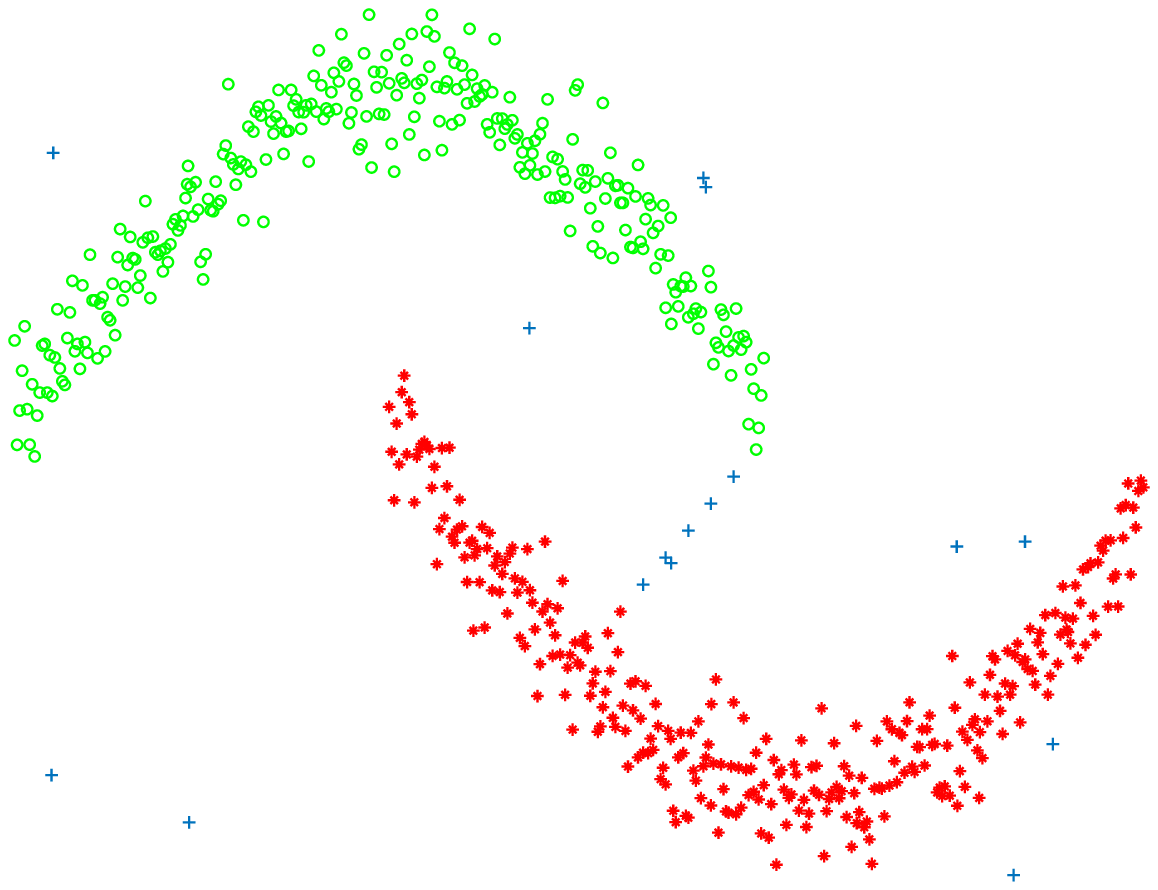}}
                                & \multicolumn{4}{c}{\includegraphics[width=8cm,height=7cm]{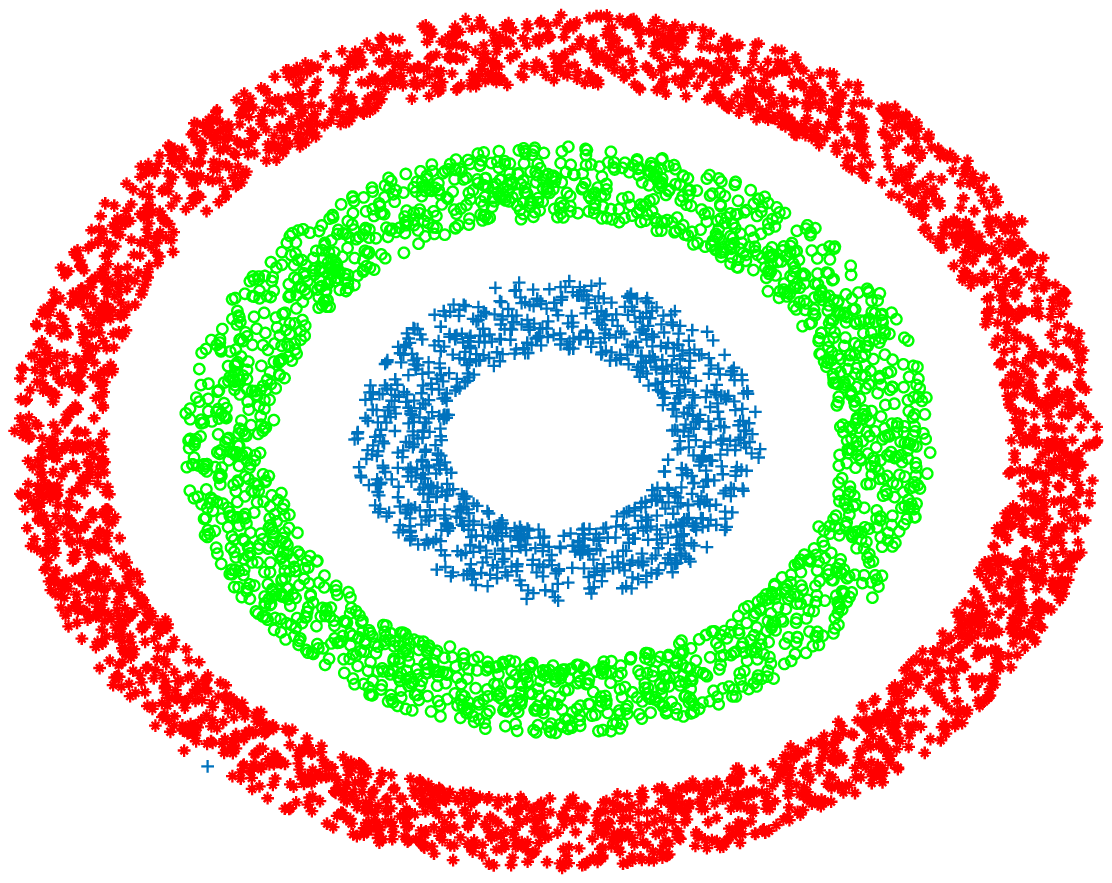}}               \\ \cline{2-17}
                                & RI      & ARI     & \multicolumn{1}{c|}{FS}         & RT         & RI         & ARI      & \multicolumn{1}{c|}{FS}        & RT
                                & RI      & ARI     & \multicolumn{1}{c|}{FS}         & RT         & RI         & ARI      & \multicolumn{1}{c|}{FS}        & RT    \\ \cline{2-17}
                                & 0.9918  & 1       & \multicolumn{1}{c|}{0.9917}   & 0.0625     & 0.9536     & 1        & \multicolumn{1}{c|}{0.9411}  & 0.0609
                                 & 0.9806 & 0.9614  & \multicolumn{1}{c|}{0.9794}   & 0.0641     & 1          & 1        & \multicolumn{1}{c|}{1}       & 5.4352    \\ \hline
\multirow{3}{*}{Spectral}       & \multicolumn{4}{c|}{\includegraphics[width=8cm,height=7cm]{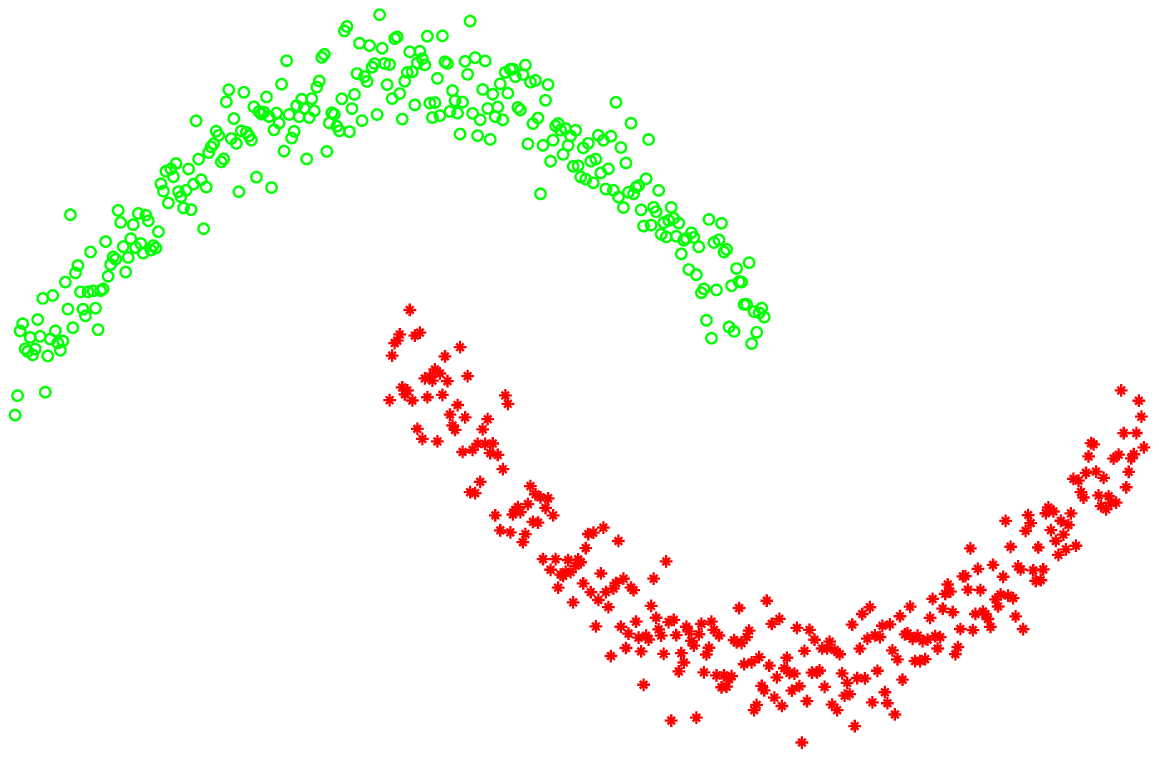}}
                                & \multicolumn{4}{c|}{\includegraphics[width=8cm,height=7cm]{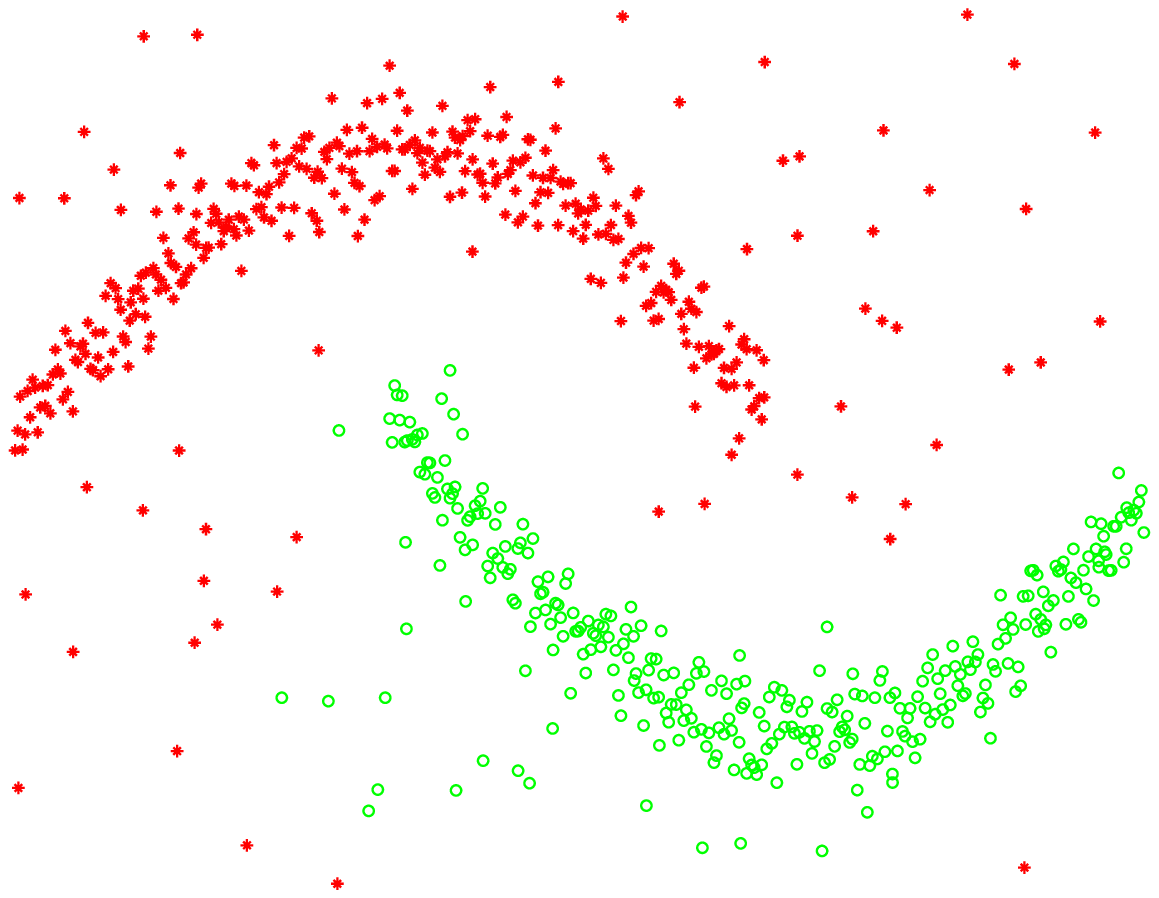}}
                                & \multicolumn{4}{c|}{\includegraphics[width=8cm,height=7cm]{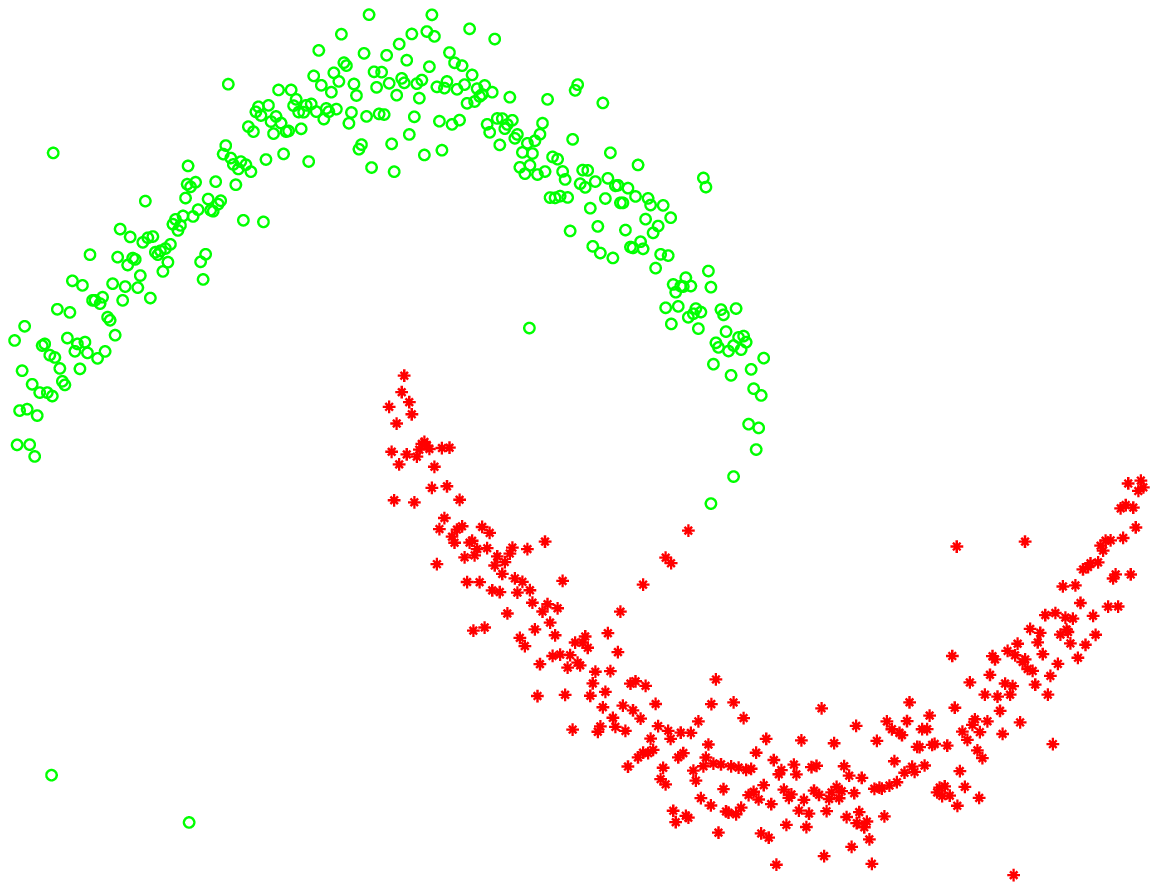}}
                                & \multicolumn{4}{c}{\includegraphics[width=8cm,height=7cm]{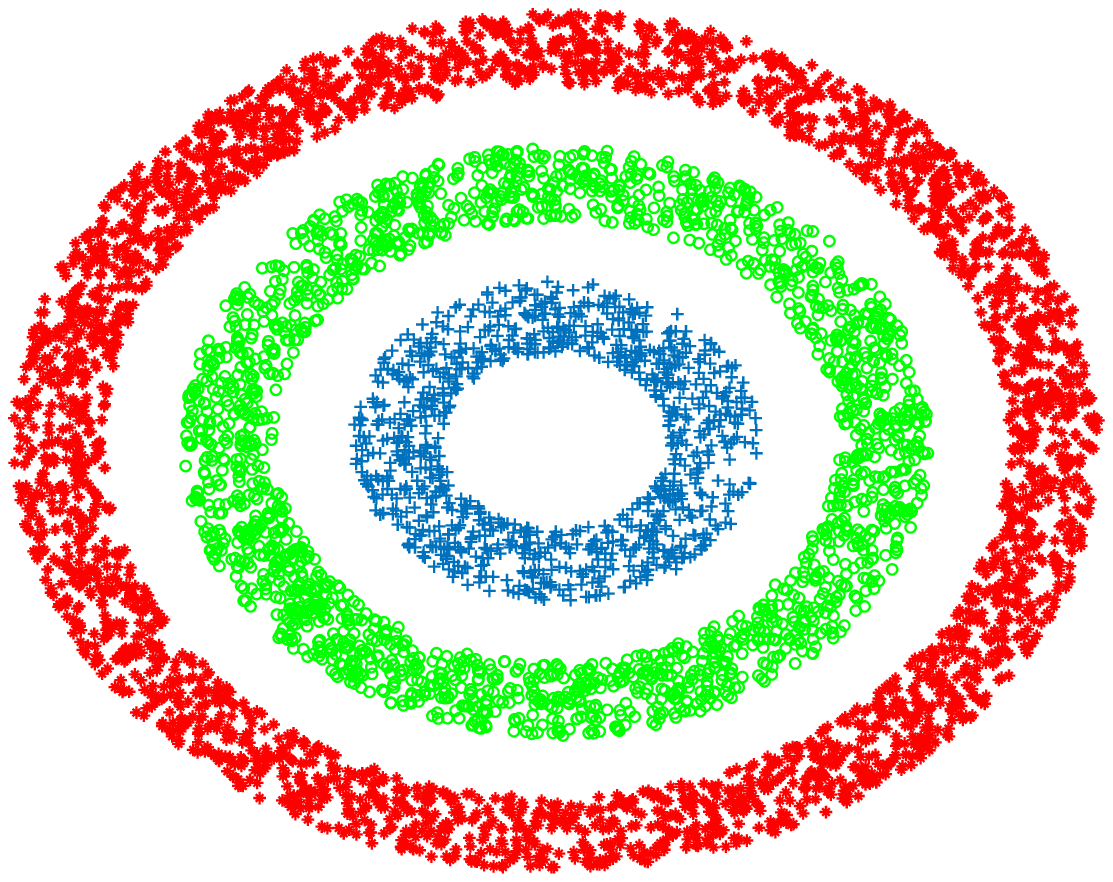}}               \\ \cline{2-17}
                                & RI      & ARI     & \multicolumn{1}{c|}{FS}      & RT    & RI       & ARI      & \multicolumn{1}{c|}{FS}         & RT
                                & RI      & ARI     & \multicolumn{1}{c|}{FS}      & RT    & RI       & ARI      & \multicolumn{1}{c|}{FS}         & RT   \\ \cline{2-17}
                                & 1       & 1       & \multicolumn{1}{c|}{1}      & 0.2617& 0.8686   & 1        & \multicolumn{1}{c|}{0.8520}   & 0.4633
                                & 0.9651  & 0.9969  & \multicolumn{1}{c|}{0.9639} & 0.3070& 0.9889   & 0.9889   & \multicolumn{1}{c|}{0.9876}   & 100.50  \\ \hline
\multirow{3}{*}{GOPC}           & \multicolumn{4}{c|}{\includegraphics[width=8cm,height=7cm]{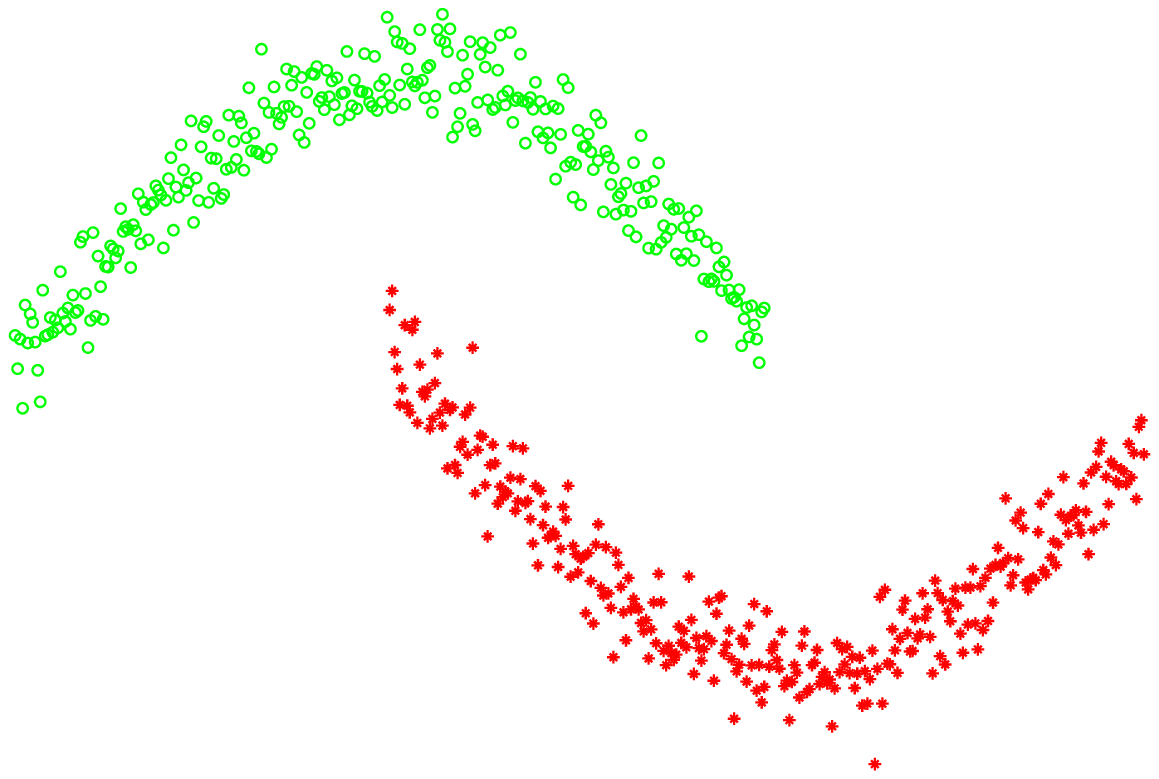}}
                                & \multicolumn{4}{c|}{\includegraphics[width=8cm,height=7cm]{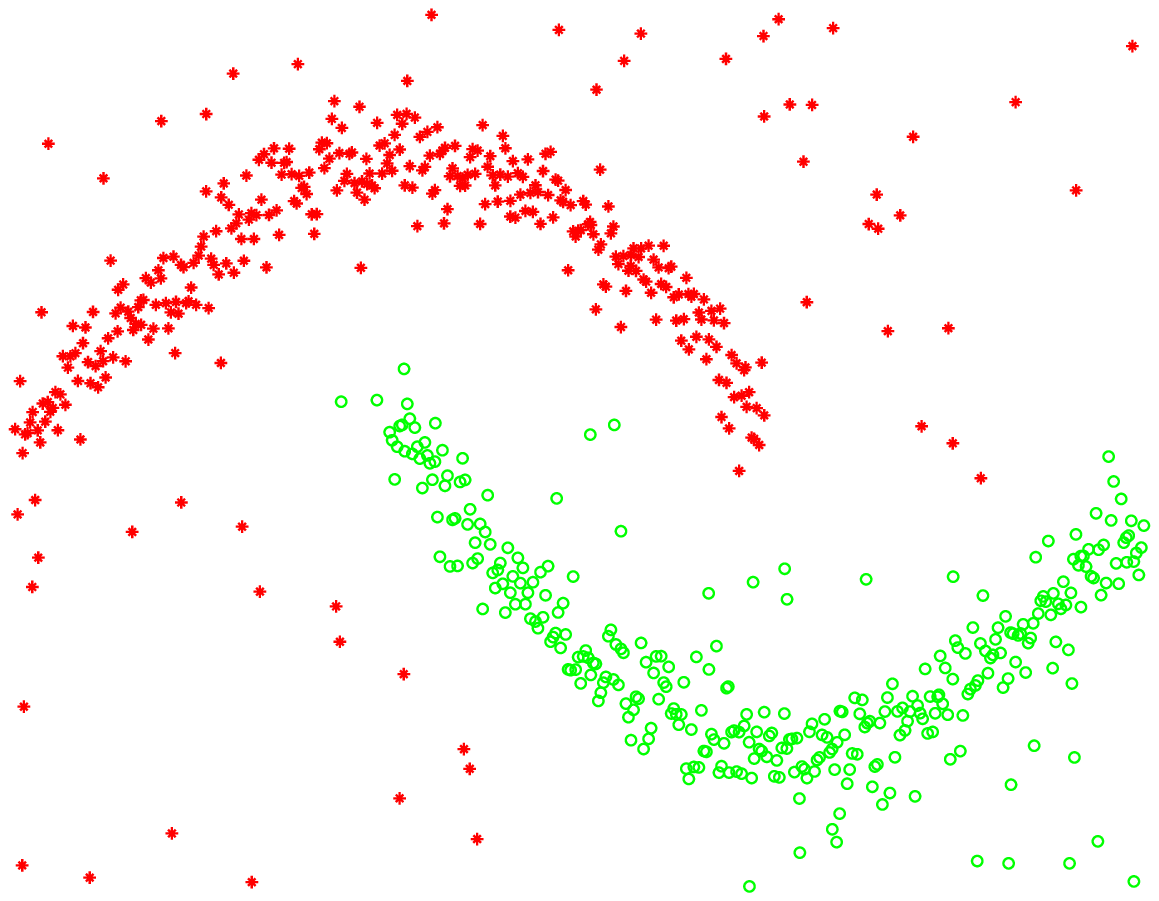}}
                                & \multicolumn{4}{c|}{\includegraphics[width=8cm,height=7cm]{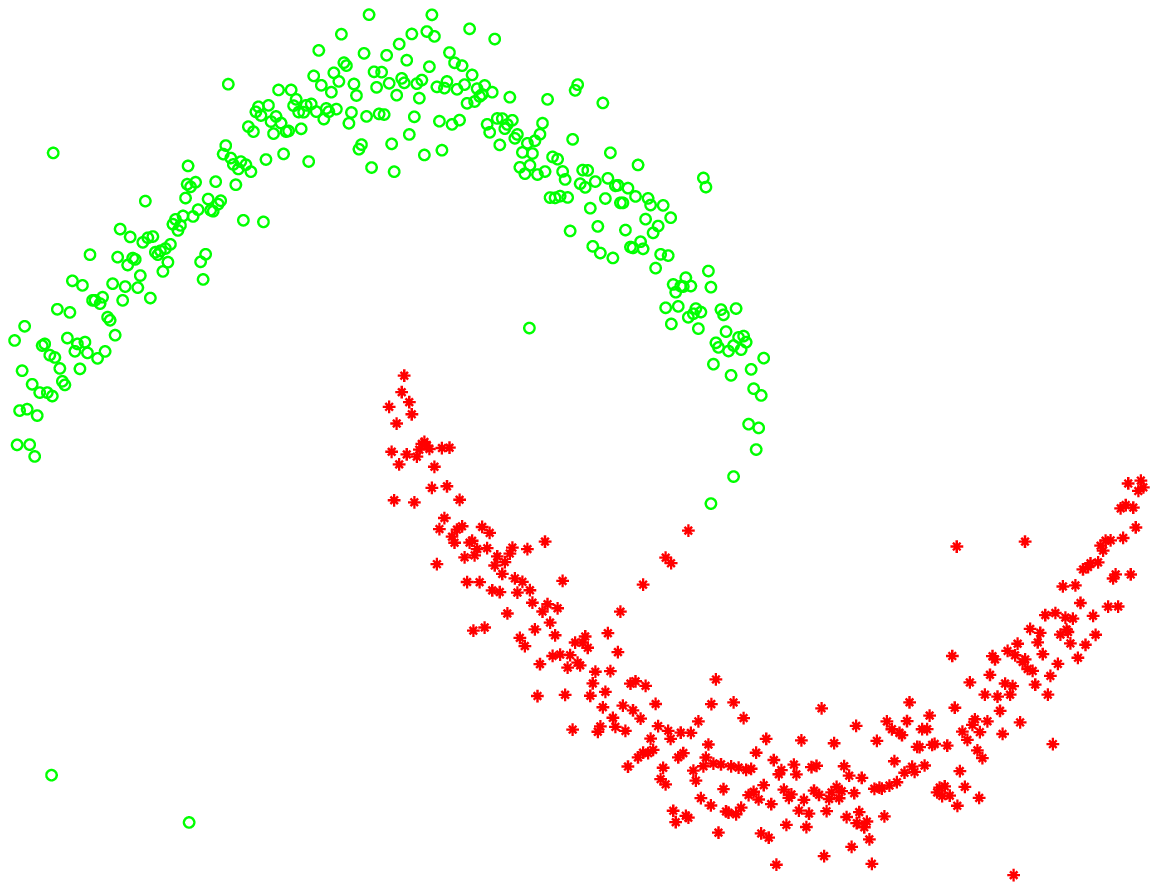}}
                                & \multicolumn{4}{c}{\includegraphics[width=8cm,height=7cm]{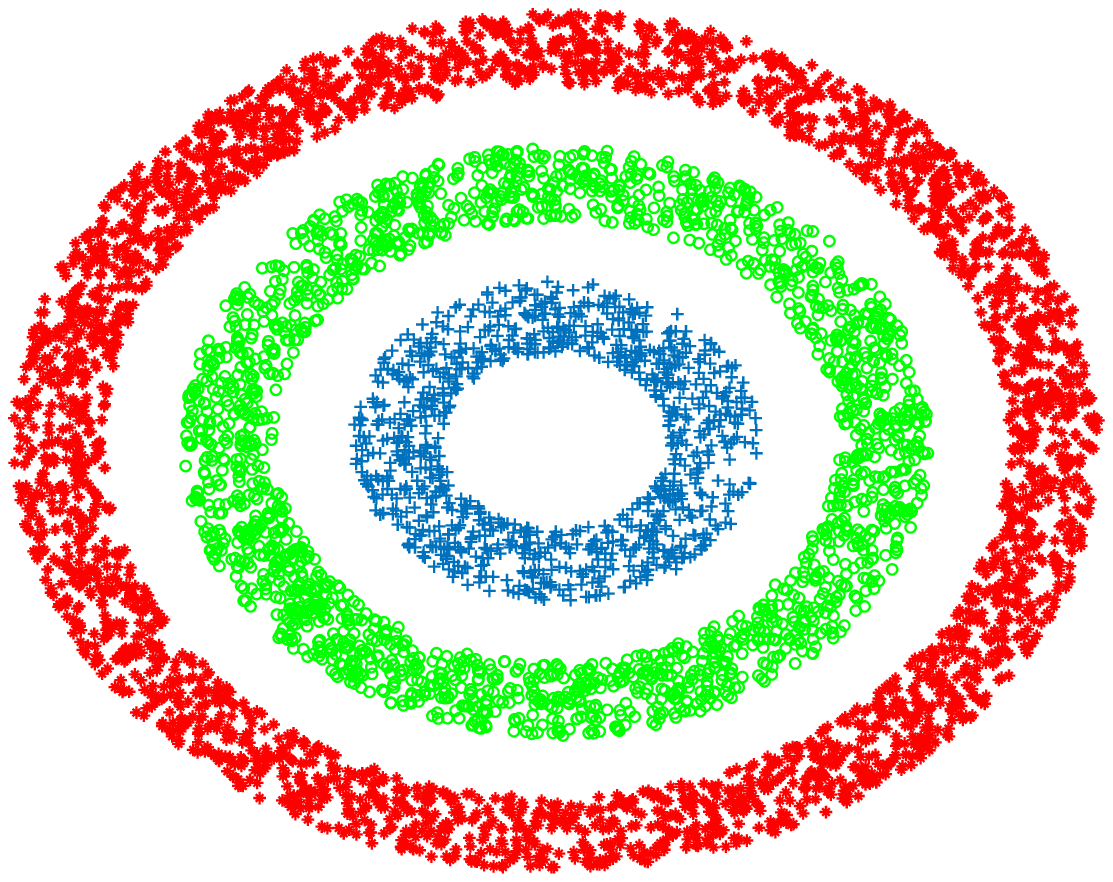}}                                 \\ \cline{2-17}
                                & RI      & ARI     & \multicolumn{1}{c|}{FS}      & RT    & RI        & ARI      & \multicolumn{1}{c|}{FS}         & RT
                                & RI      & ARI     & \multicolumn{1}{c|}{FS}      & RT   & RI        & ARI      & \multicolumn{1}{c|}{FS}         & RT              \\ \cline{2-17}
                                & 1       & 1       & \multicolumn{1}{c|}{1}     & 0.0516& 0.8685    & 1        & \multicolumn{1}{c|}{0.8518}   & 0.0617
                                & 0.9557  & 0.9876  & \multicolumn{1}{c|}{0.9543}& 0.0516& 1         & 1        &\multicolumn{1}{c|}{1}        & 4.9891           \\ \hline
\multirow{3}{*}{CutPC}           & \multicolumn{4}{c|}{\includegraphics[width=8cm,height=7cm]{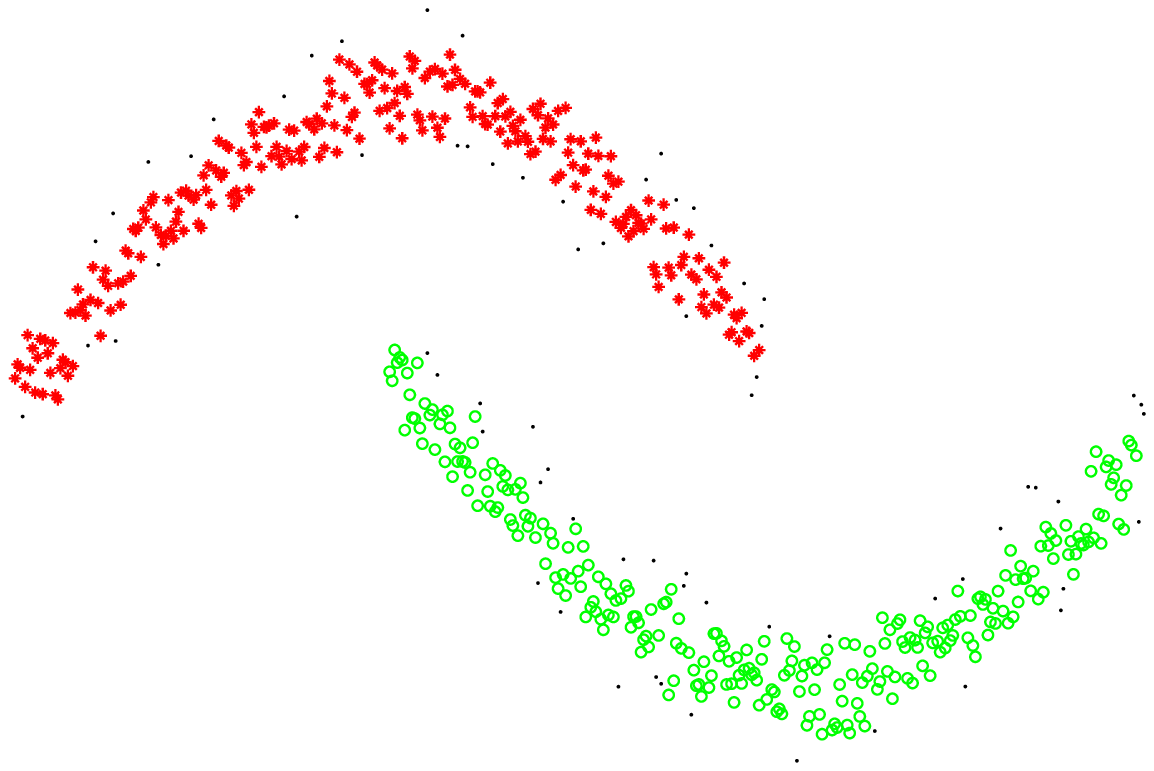}}
                                & \multicolumn{4}{c|}{\includegraphics[width=8cm,height=7cm]{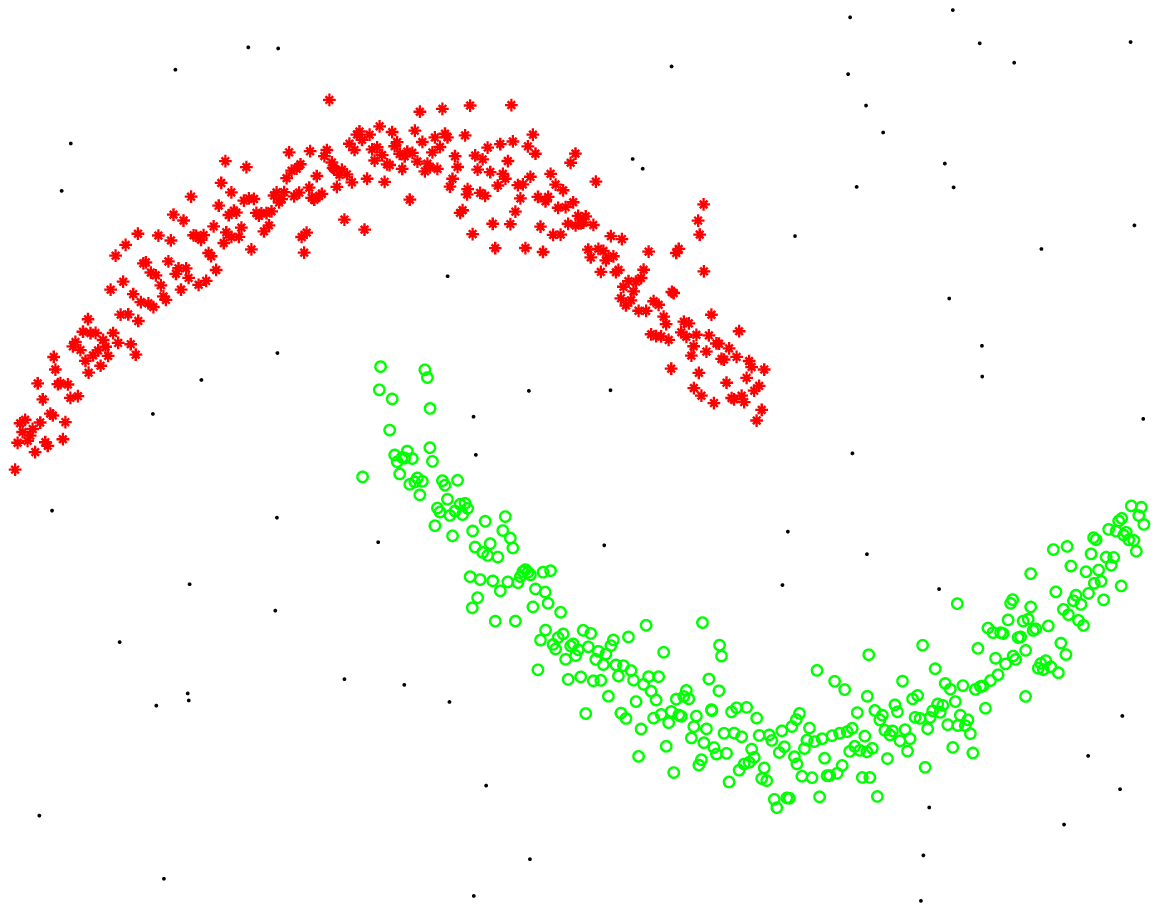}}
                                & \multicolumn{4}{c|}{\includegraphics[width=8cm,height=7cm]{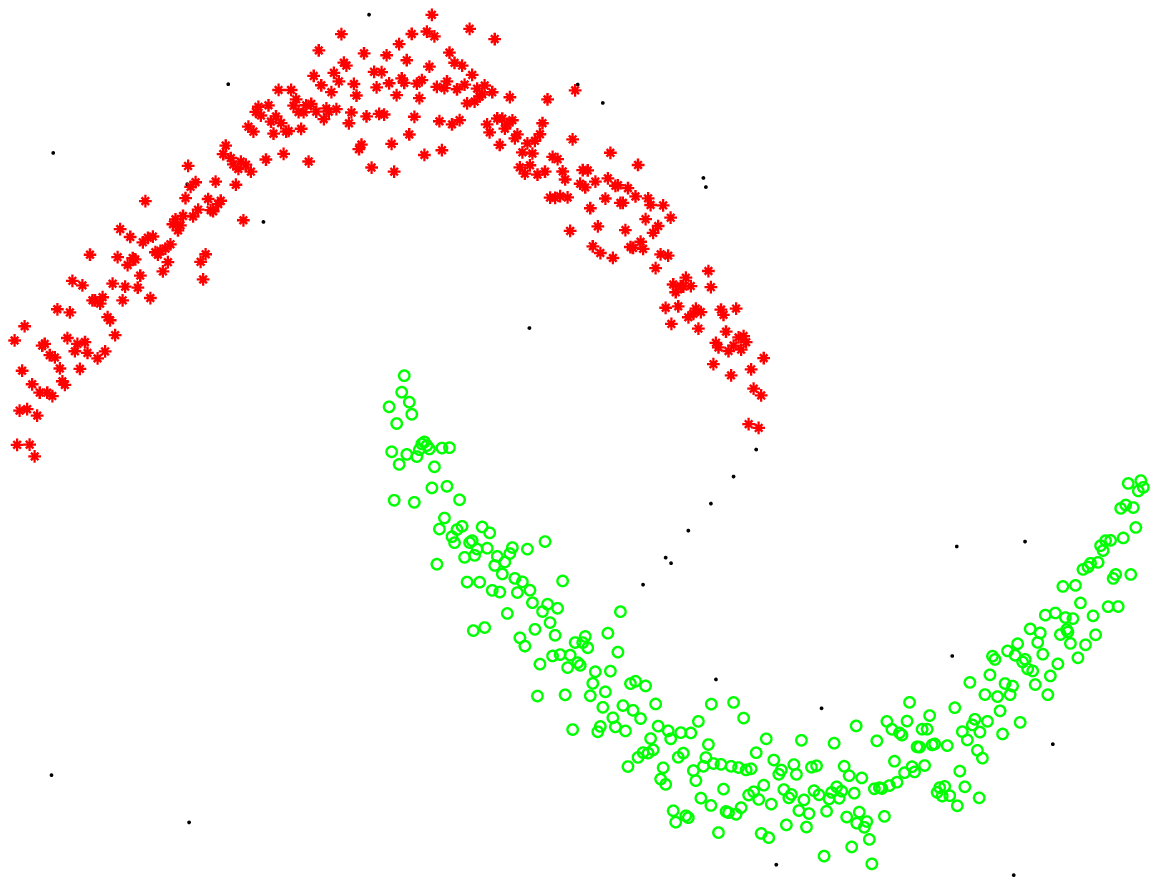}}
                                & \multicolumn{4}{c}{\includegraphics[width=8cm,height=7cm]{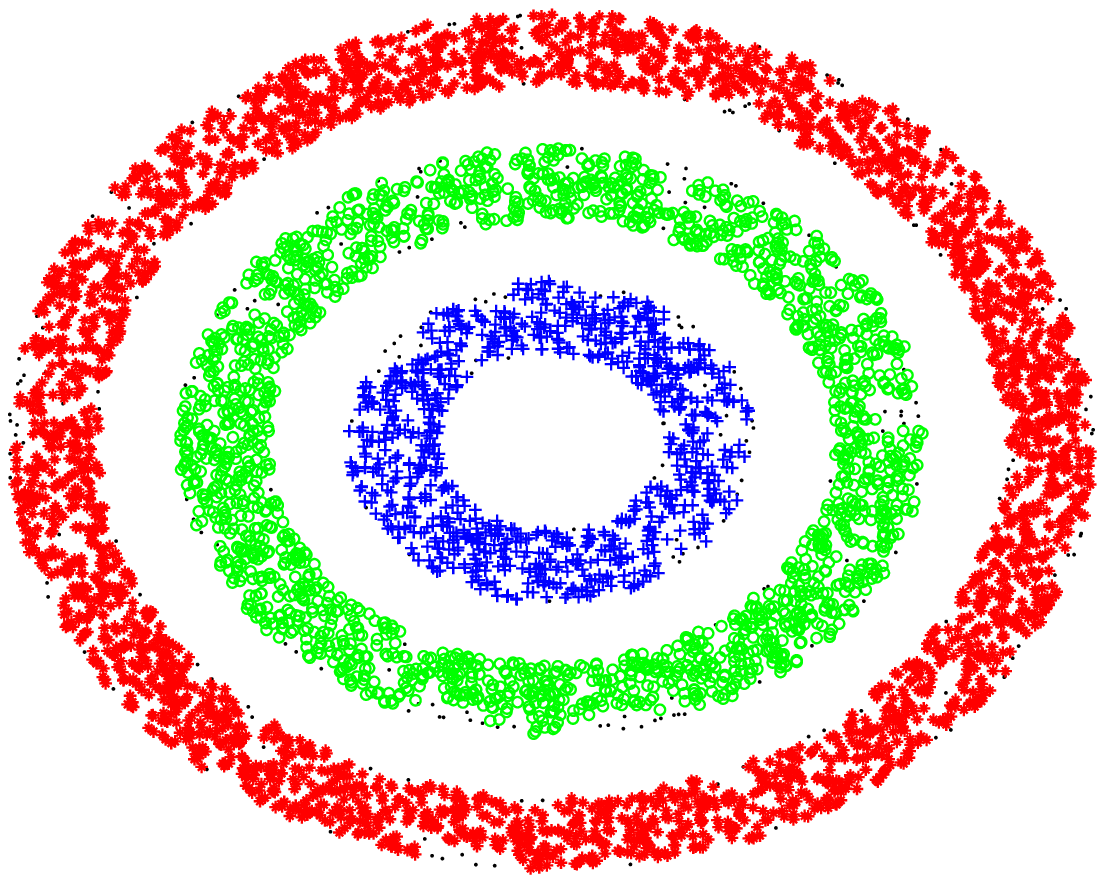}}                                 \\ \cline{2-17}
                                & RI      & ARI     & \multicolumn{1}{c|}{FS}      & RT    & RI        & ARI      & \multicolumn{1}{c|}{FS}         & RT
                                & RI      & ARI     & \multicolumn{1}{c|}{FS}      & RT   & RI        & ARI      & \multicolumn{1}{c|}{FS}         & RT              \\ \cline{2-17}
                                & 0.8900       & \textbf{1}       & \multicolumn{1}{c|}{0.8780}     & 0.1273 & 0.9451   & \textbf{1}        & \multicolumn{1}{c|}{0.9320}   & 0.1117
                                &  0.9662  & \textbf{0.9969}  & \multicolumn{1}{c|}{0.9636}& 0.0594 & 0.9697         & 1       &\multicolumn{1}{c|}{0.9595}        & 3.7156          \\ \hline
\end{tabular}
}
\end{table}

% \usepackage{multirow}
\begin{table}
\caption{Accuracy of the indicated algorithms and non-spherical datasets.(cont.)}
\centering
%\ContinuedFloat
\huge
\resizebox{0.9\linewidth}{0.65\linewidth}{%
\begin{tabular}{c|cccc|cccc|cccc|cccc}
\hline
\diagbox{Methods}{Performance}{Dataset}   & \multicolumn{4}{c|}{\emph{spiral} N=567} & \multicolumn{4}{c|}{\emph{t4.8k} N=8000} & \multicolumn{4}{c|}{\emph{atom} N=3000}         & \multicolumn{4}{c}{\emph{chains} N=600}        \\ \hline
\multirow{3}{*}{PaVa}           & \multicolumn{4}{c|}{\includegraphics[width=8cm,height=7cm]{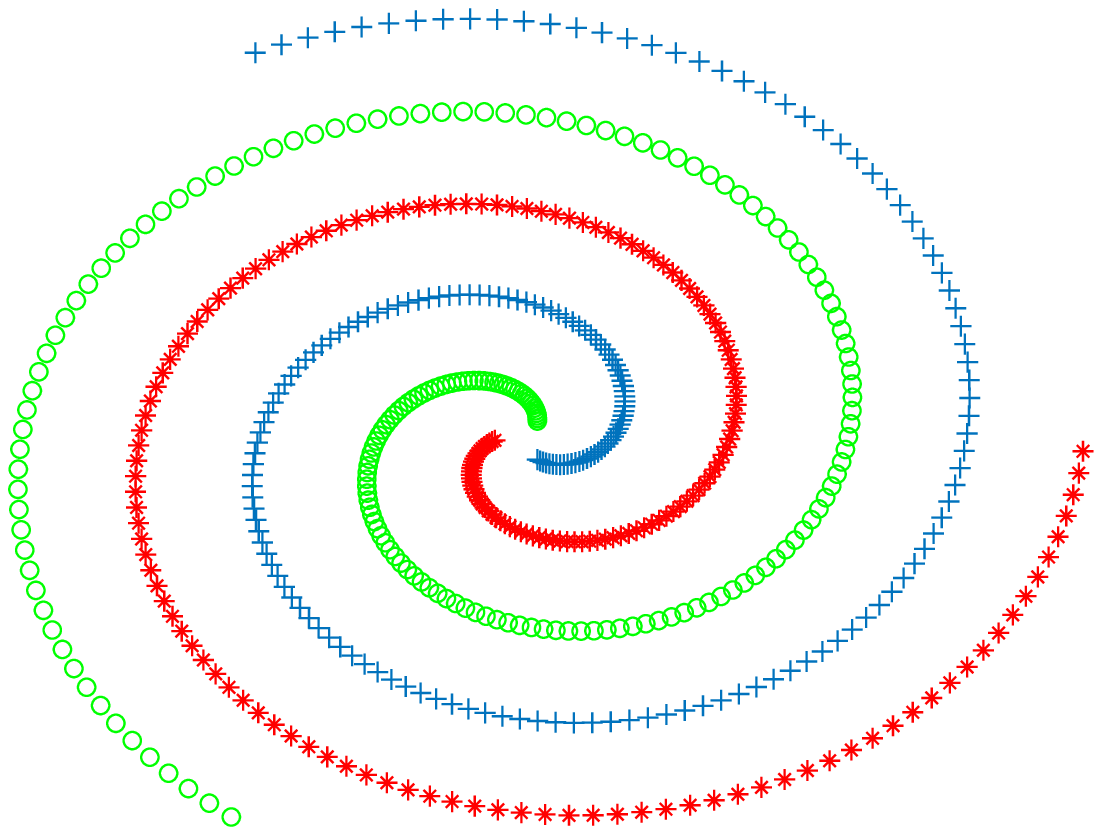}}
                                & \multicolumn{4}{c|}{\includegraphics[width=8cm,height=7cm]{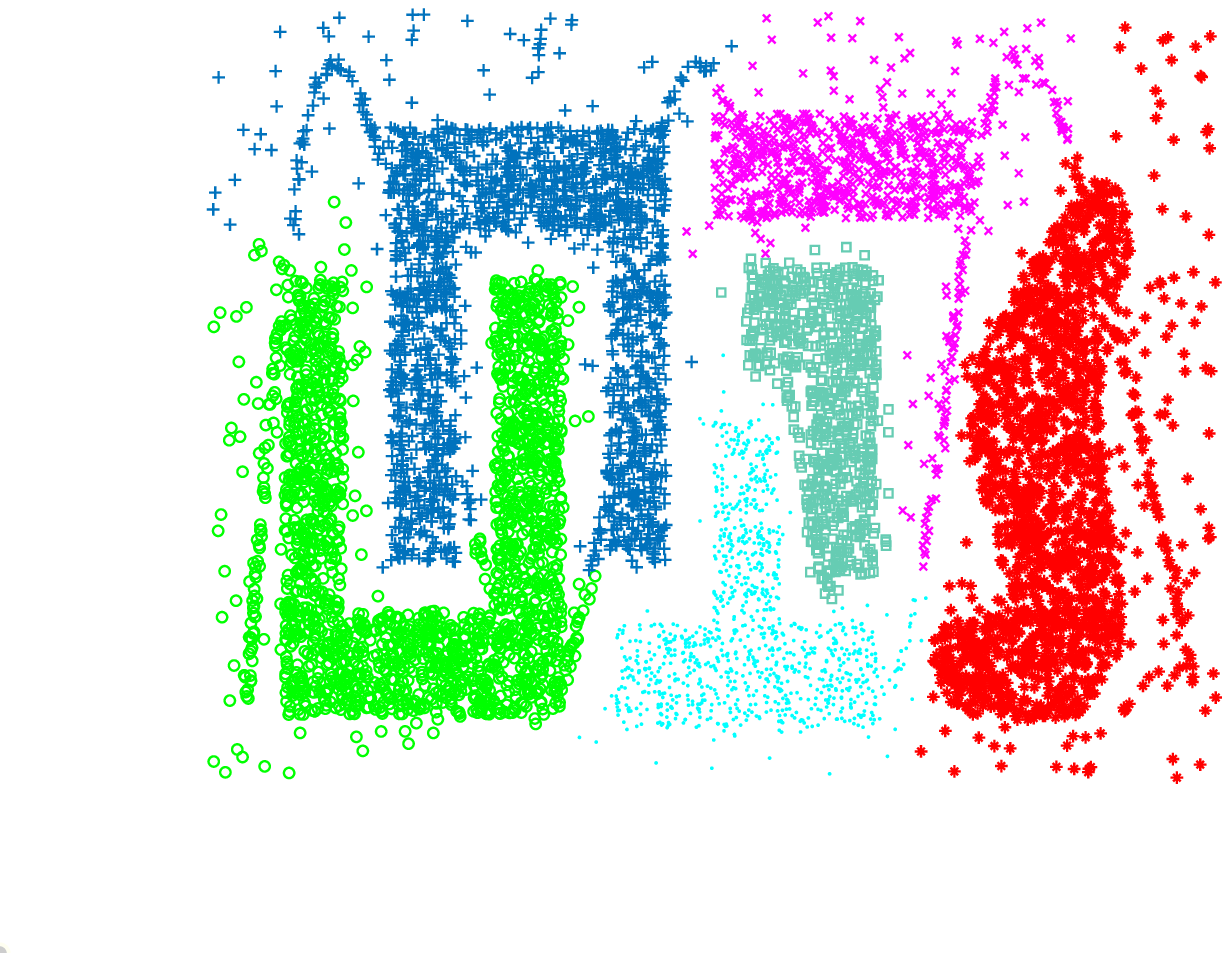}}
                                & \multicolumn{4}{c|}{\includegraphics[width=8cm,height=7cm]{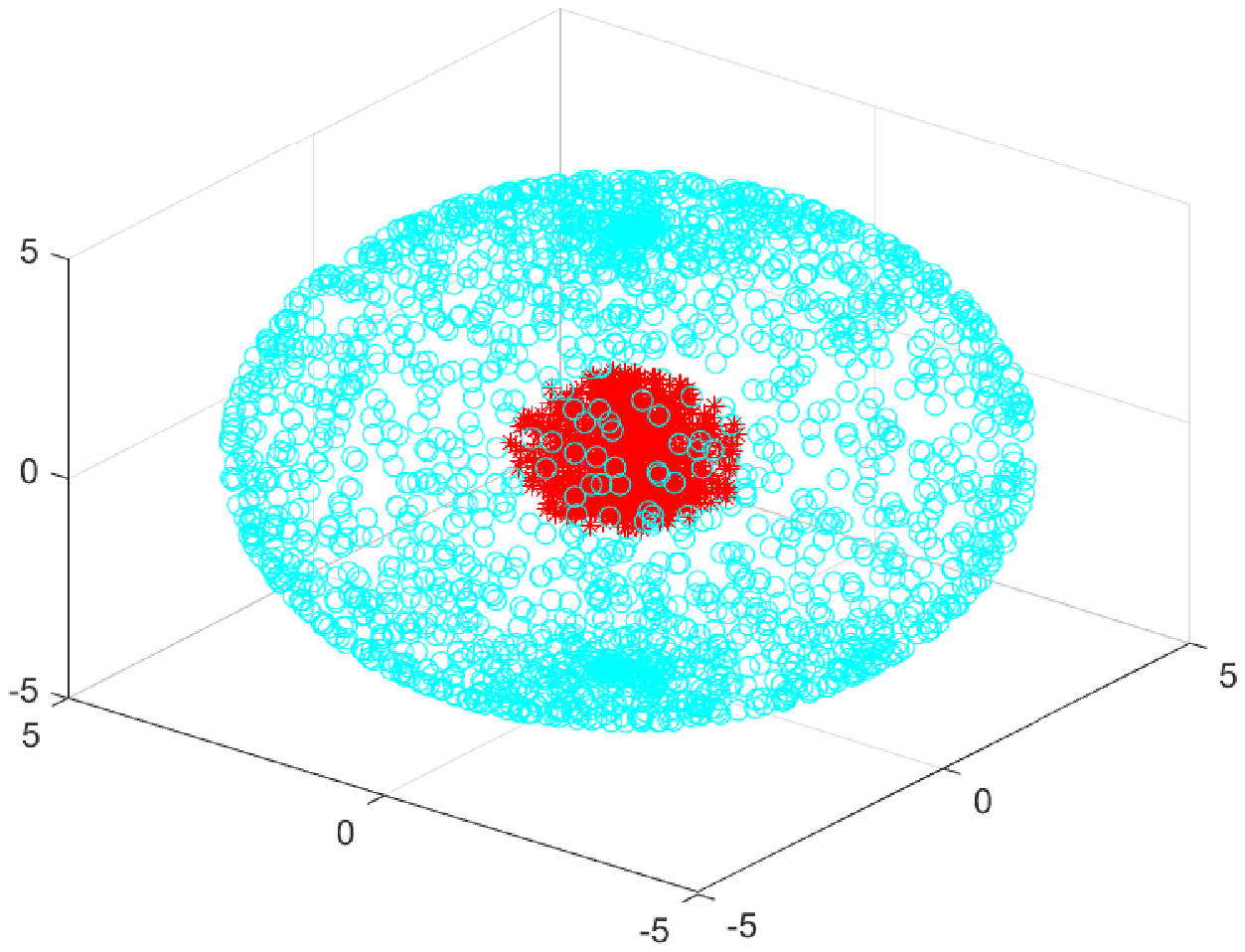}}
                                & \multicolumn{4}{c}{\includegraphics[width=8cm,height=7cm]{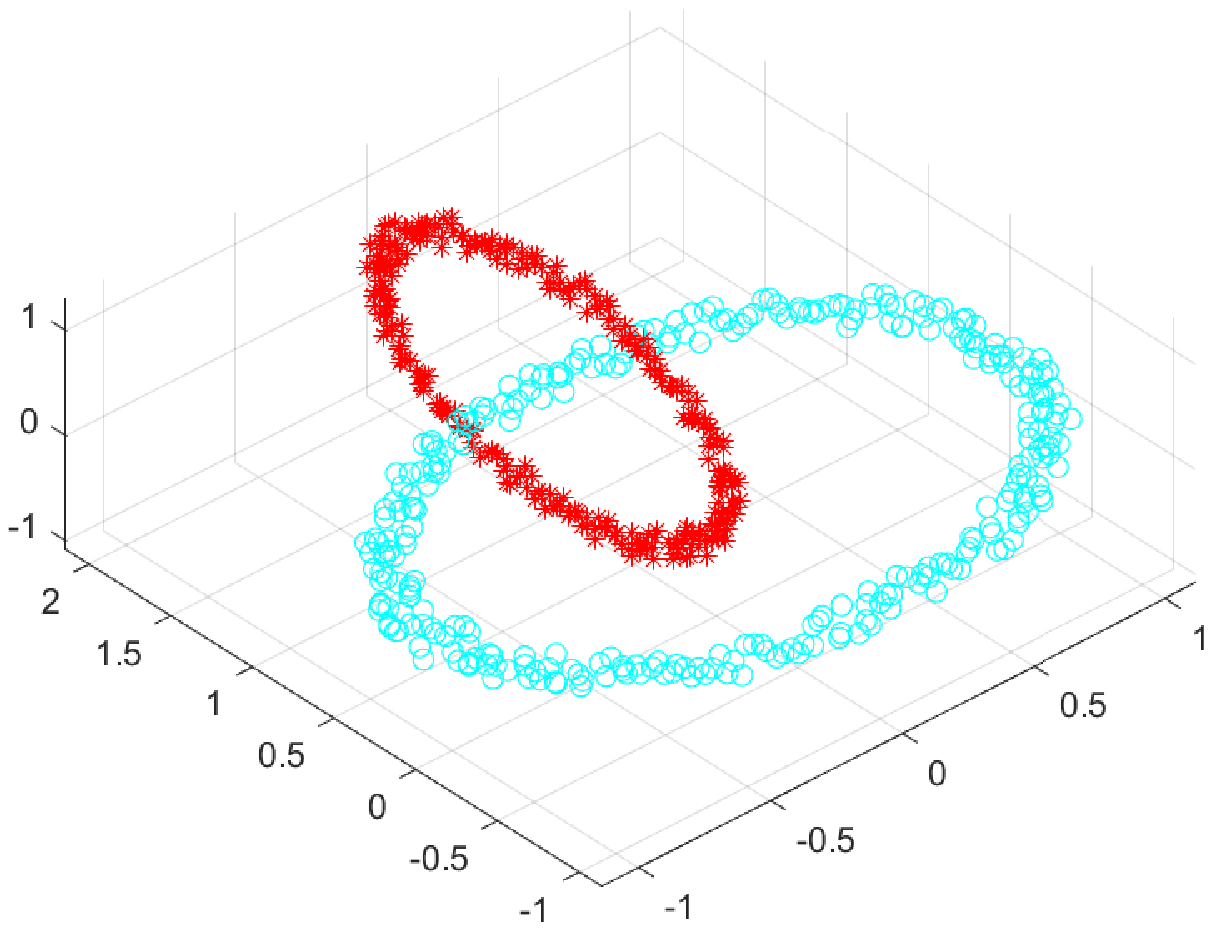}}              \\ \cline{2-17}
                                & RI     & ARI    & \multicolumn{1}{c|}{FS}   &RT       & RI     & ARI    &  \multicolumn{1}{c|}{FS}     &RT
                                & RI     & ARI    &  \multicolumn{1}{c|}{FS}  &RT       & RI     & ARI    &  \multicolumn{1}{c|}{FS}     &RT     \\ \cline{2-17}
                                & 1      & 1      &  \multicolumn{1}{c|}{1}  &0.0984   & N/A    & N/A    &  \multicolumn{1}{c|}{N/A}   &2.1960
                                & 1      & 1      &  \multicolumn{1}{c|}{1}  &0.3375   & 1      & 1      &  \multicolumn{1}{c|}{1}     &0.1172    \\ \hline
\multirow{3}{*}{Hierarchical}   & \multicolumn{4}{c|}{\includegraphics[width=8cm,height=7cm]{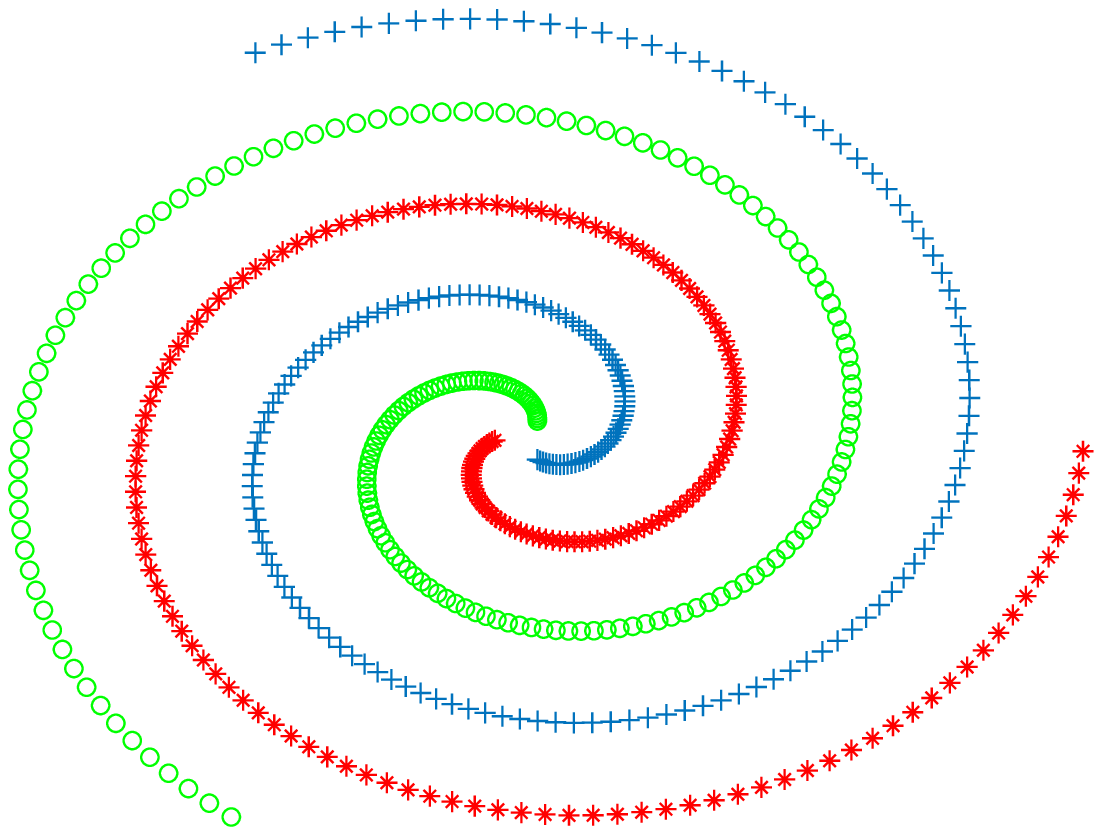}}
                                & \multicolumn{4}{c|}{\includegraphics[width=8cm,height=7cm]{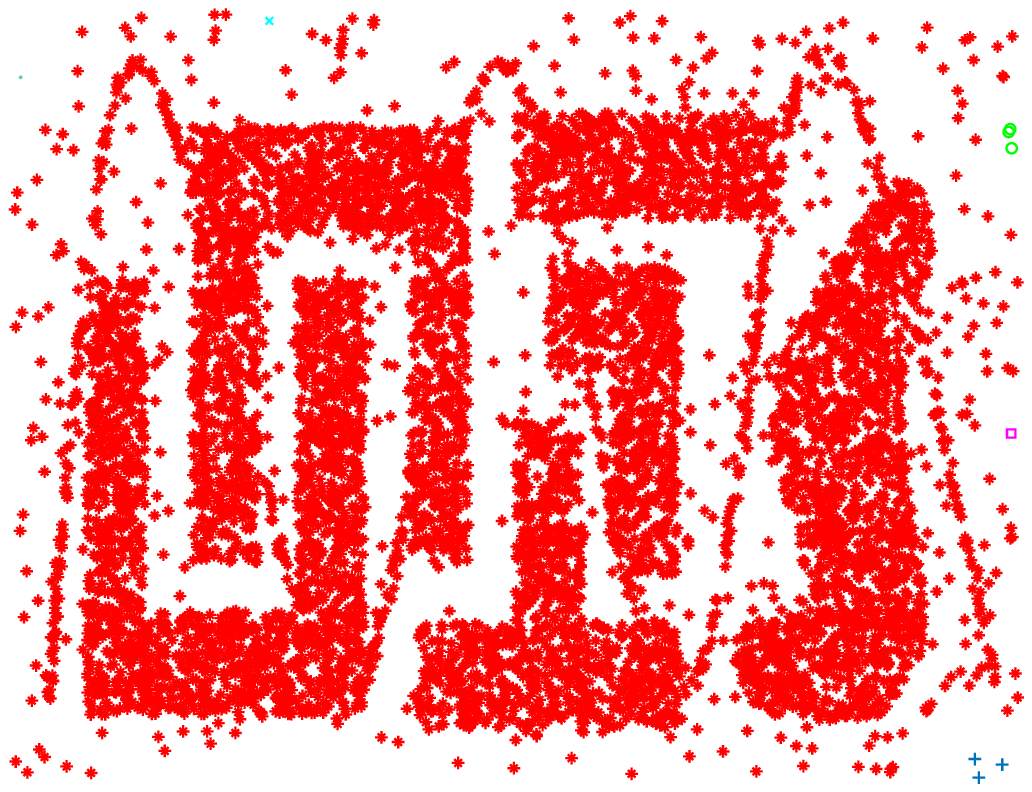}}
                                & \multicolumn{4}{c|}{\includegraphics[width=8cm,height=7cm]{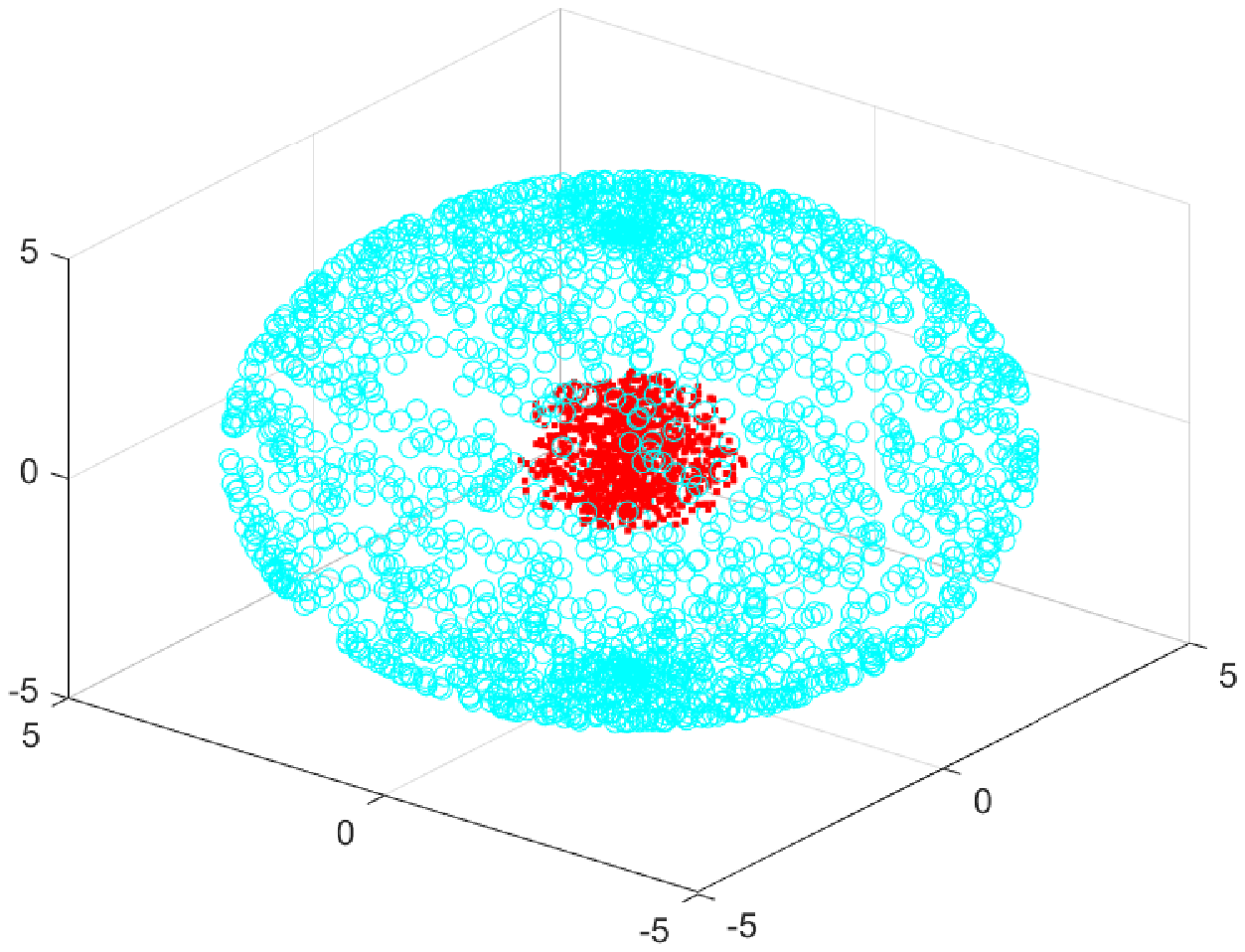}}
                                & \multicolumn{4}{c}{\includegraphics[width=8cm,height=7cm]{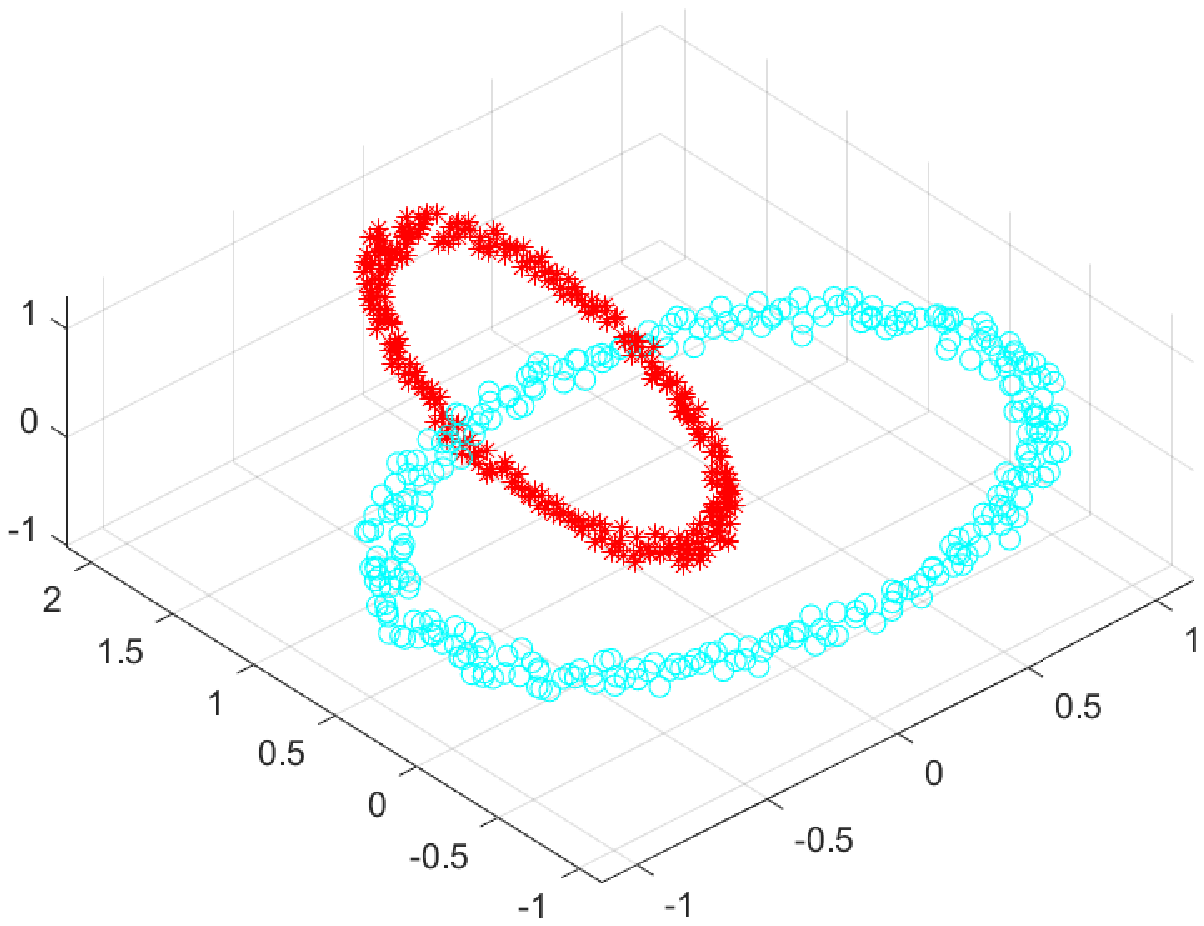}}              \\ \cline{2-17}
                                & RI     & ARI    & \multicolumn{1}{c|}{FS}   &RT        & RI     & ARI    & \multicolumn{1}{c|}{FS}     &RT
                                & RI     & ARI    & \multicolumn{1}{c|}{FS}   &RT        & RI     & ARI    & \multicolumn{1}{c|}{FS}     &RT      \\ \cline{2-17}
                                & 1      & 1      & \multicolumn{1}{c|}{1}   &0.0063    & N/A    & N/A    & \multicolumn{1}{c|}{N/A}   &1.5148
                                & 1      & 1      & \multicolumn{1}{c|}{1}   &0.1890    & 1      & 1      & \multicolumn{1}{c|}{1}     &0.0094       \\ \hline
\multirow{3}{*}{kernel k-means} & \multicolumn{4}{c|}{\includegraphics[width=8cm,height=7cm]{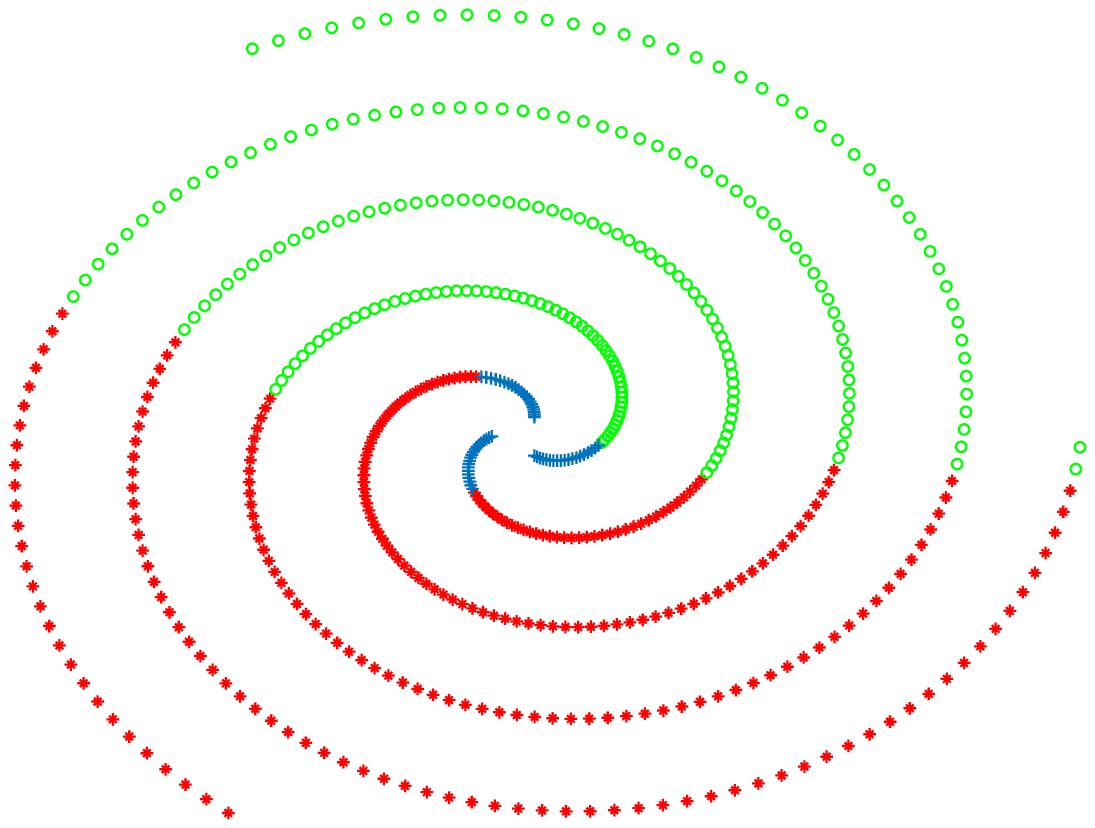}}
                                & \multicolumn{4}{c|}{\includegraphics[width=8cm,height=7cm]{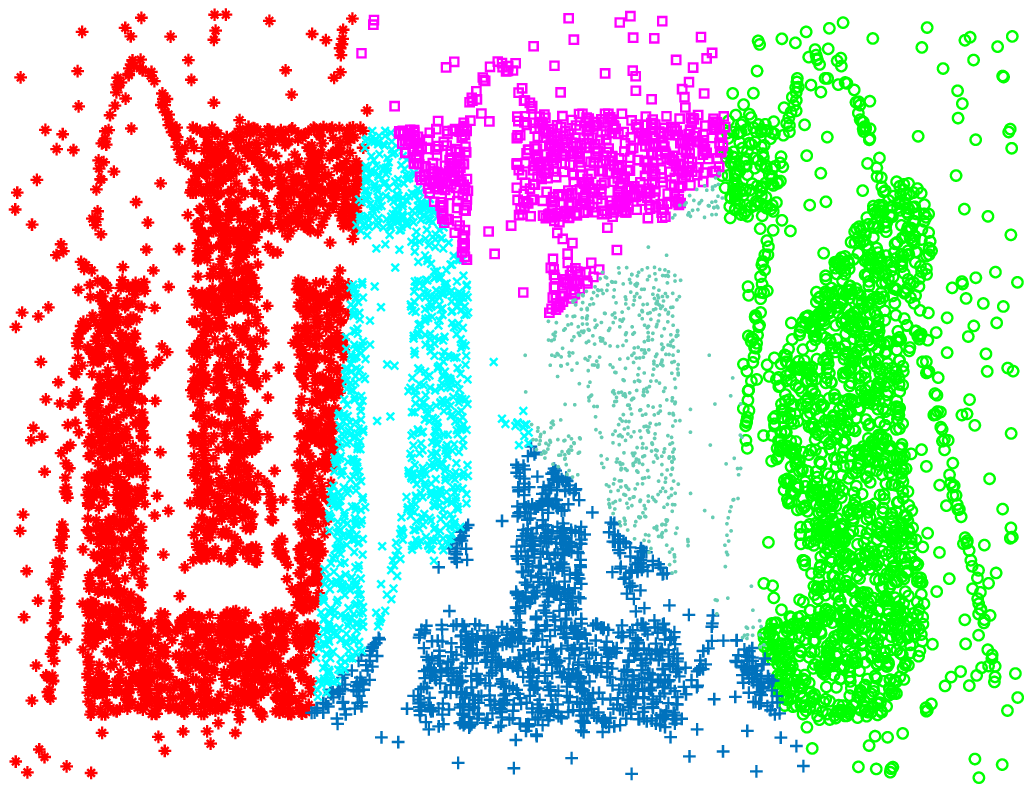}}
                                & \multicolumn{4}{c|}{\includegraphics[width=8cm,height=7cm]{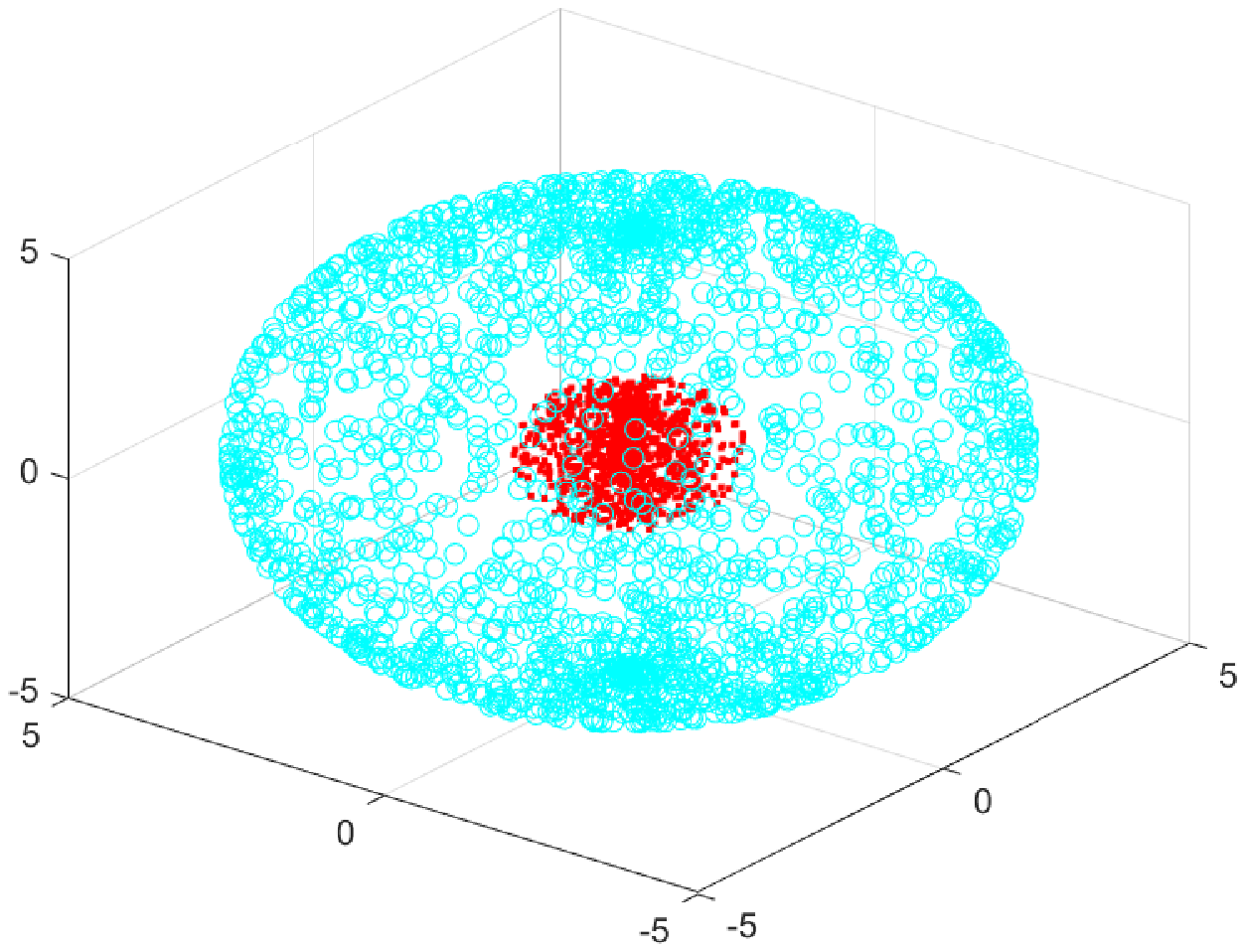}}
                                & \multicolumn{4}{c}{\includegraphics[width=8cm,height=7cm]{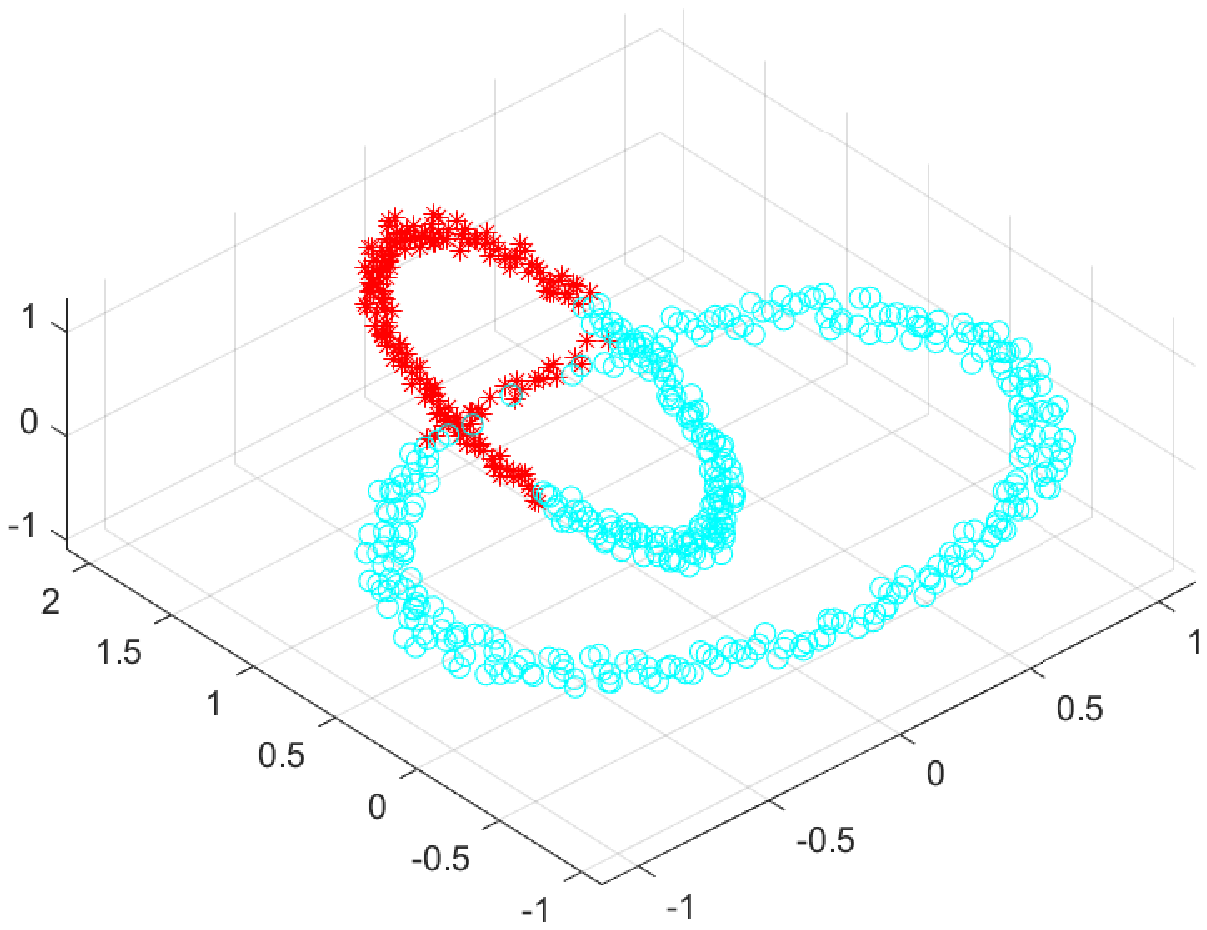}}              \\ \cline{2-17}
                                & RI     & ARI    & \multicolumn{1}{c|}{FS}        &RT     & RI     & ARI    & \multicolumn{1}{c|}{FS}      &RT
                                & RI     & ARI    & \multicolumn{1}{c|}{FS}        &RT     & RI     & ARI    & \multicolumn{1}{c|}{FS}      &RT     \\ \cline{2-17}
                                & 0.545  & 0.545  & \multicolumn{1}{c|}{0.3936}   &0.9773 & N/A    & N/A    & \multicolumn{1}{c|}{N/A}    &20.582
                                & 0.9752 & 0.9752 & \multicolumn{1}{c|}{0.9816}   &3.6352 & 0.6212 & 0.6212 & \multicolumn{1}{c|}{0.6255} &0.4258  \\ \hline
\multirow{3}{*}{DBSCAN}         & \multicolumn{4}{c|}{\includegraphics[width=8cm,height=7cm]{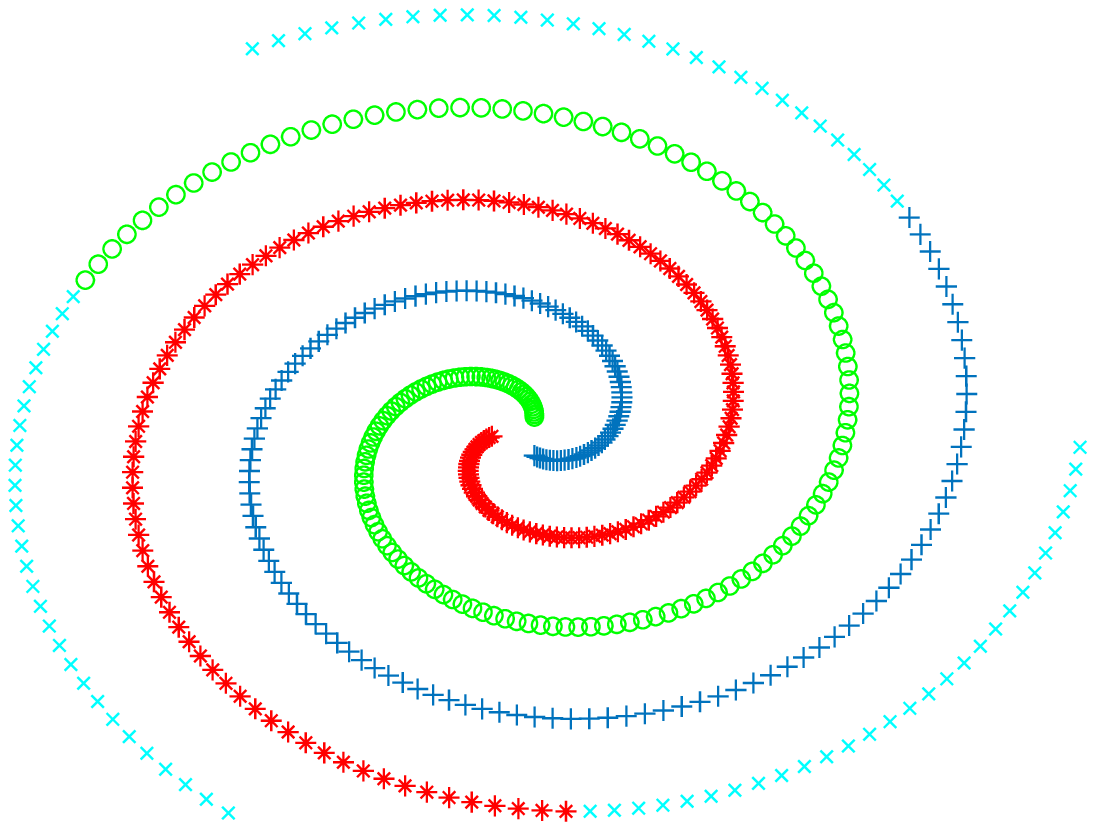}}
                                & \multicolumn{4}{c|}{\includegraphics[width=8cm,height=7cm]{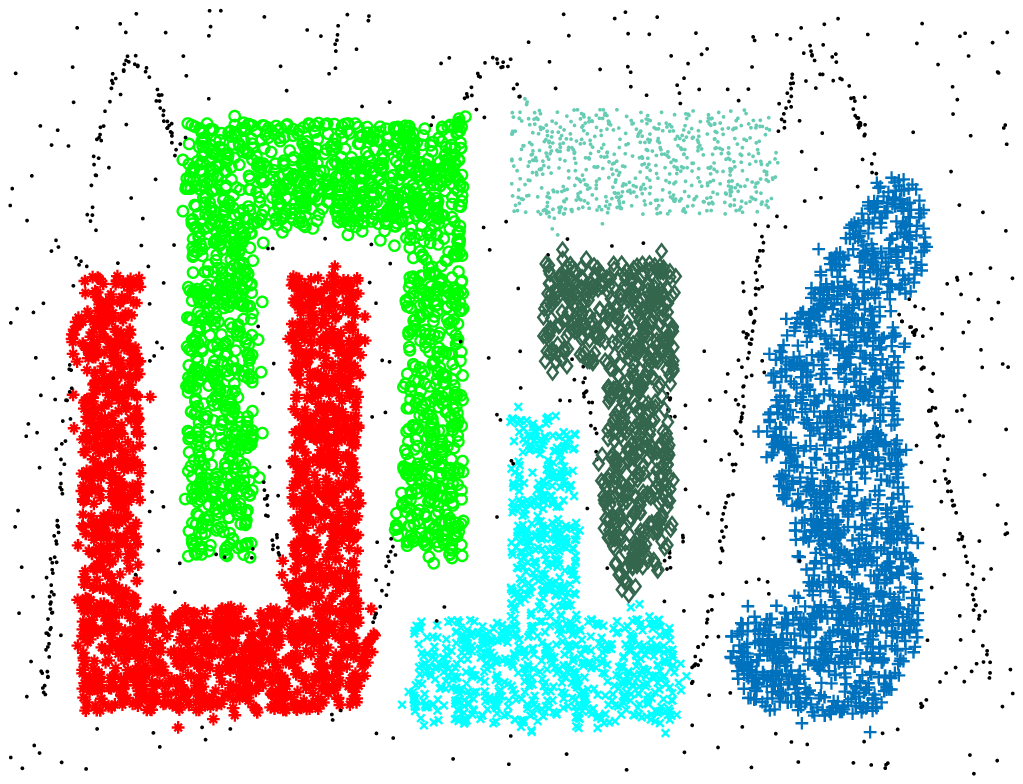}}
                                & \multicolumn{4}{c|}{\includegraphics[width=8cm,height=7cm]{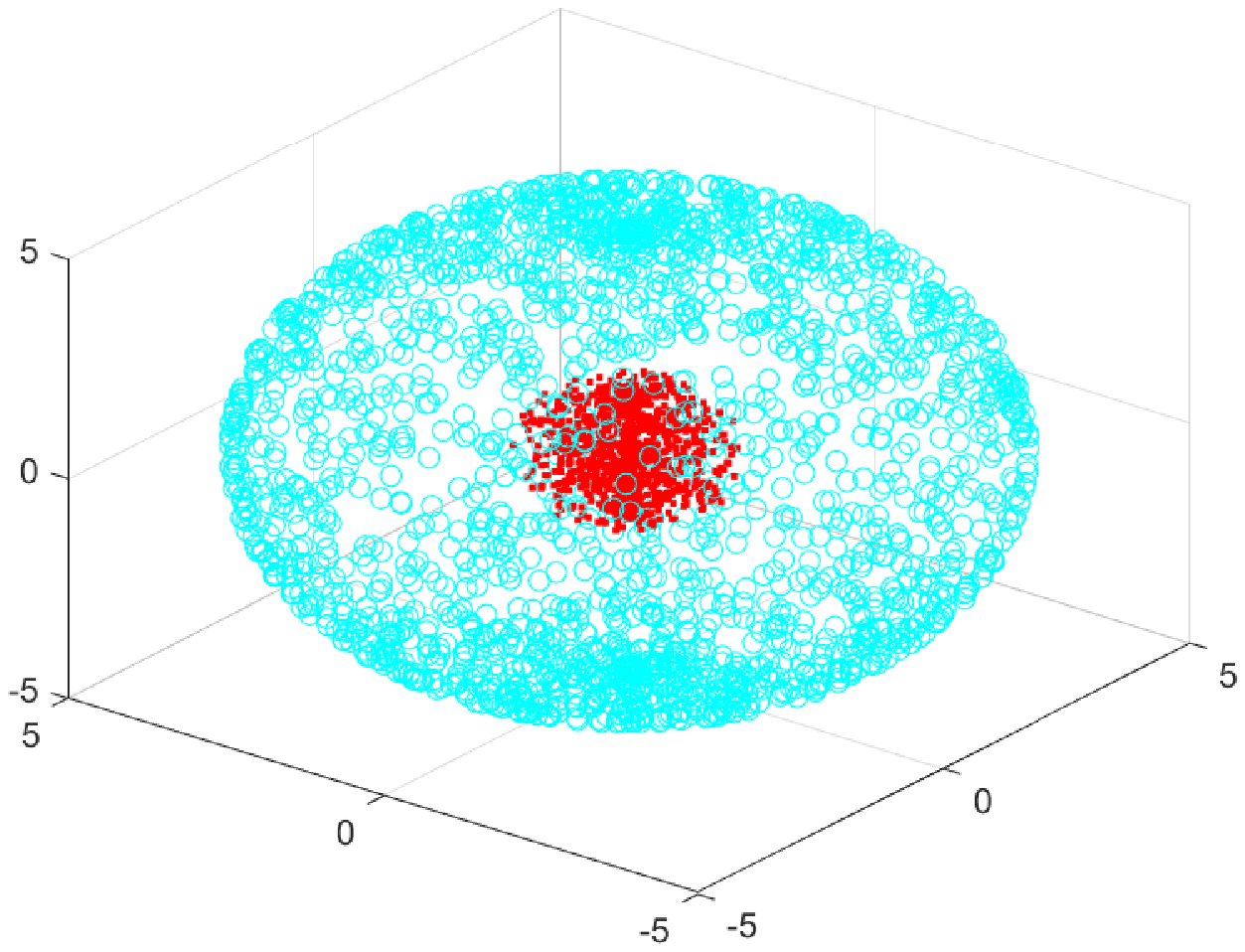}}
                                & \multicolumn{4}{c}{\includegraphics[width=8cm,height=7cm]{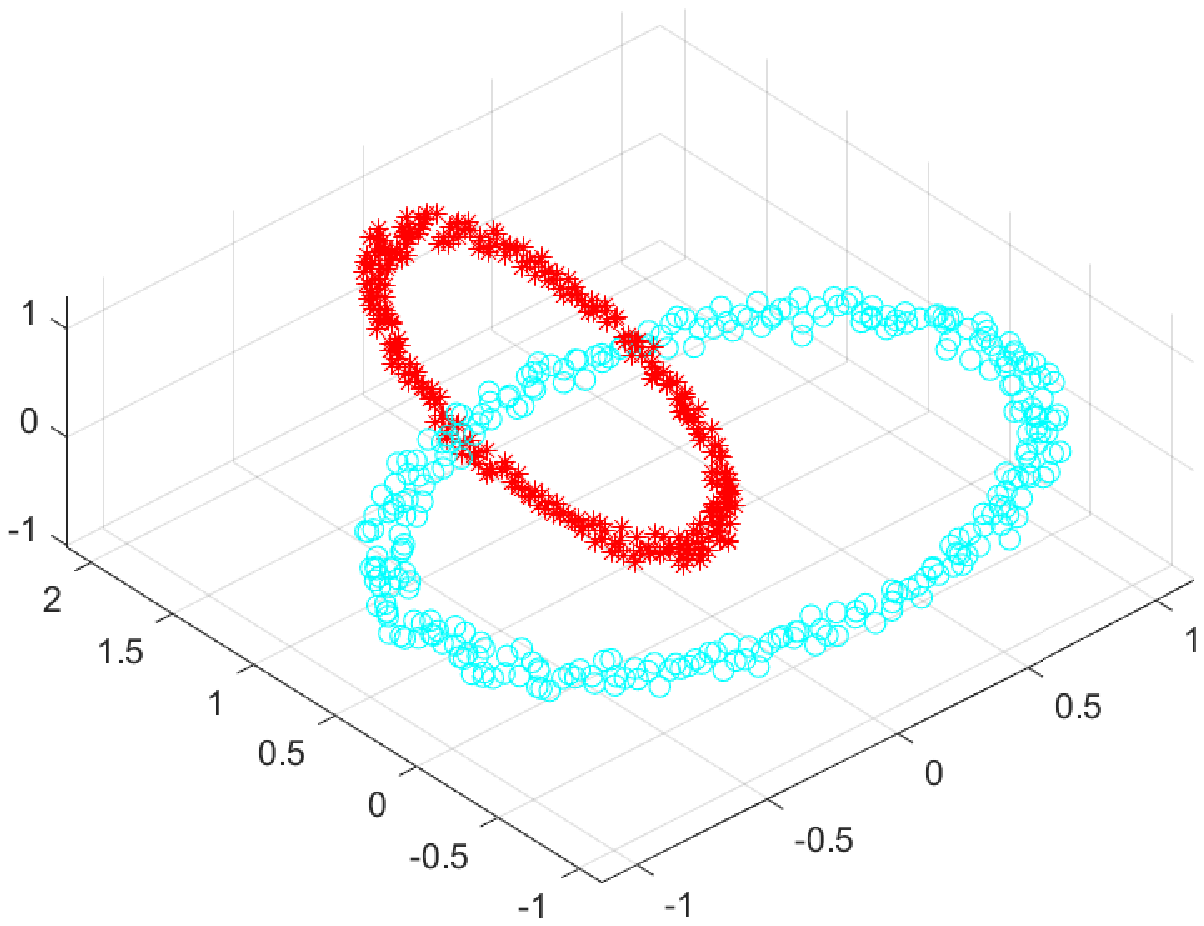}}              \\ \cline{2-17}
                                & RI     & ARI    & \multicolumn{1}{c|}{FS}      &RT      & RI     & ARI    & \multicolumn{1}{c|}{FS}     &RT
                                & RI     & ARI    & \multicolumn{1}{c|}{FS}      &RT      & RI     & ARI    & \multicolumn{1}{c|}{FS}     &RT     \\ \cline{2-17}
                                & 0.3322 & 0.3322 & \multicolumn{1}{c|}{0.4987} &0.0352  & N/A    & N/A    & \multicolumn{1}{c|}{N/A}   &6.1477
                                & 1      & 1      & \multicolumn{1}{c|}{1}      &0.7484  & 1      & 1      & \multicolumn{1}{c|}{1}     &0.0727       \\
                                \hline
\multirow{3}{*}{DC-SKCG}        & \multicolumn{4}{c|}{\includegraphics[width=8cm,height=7cm]{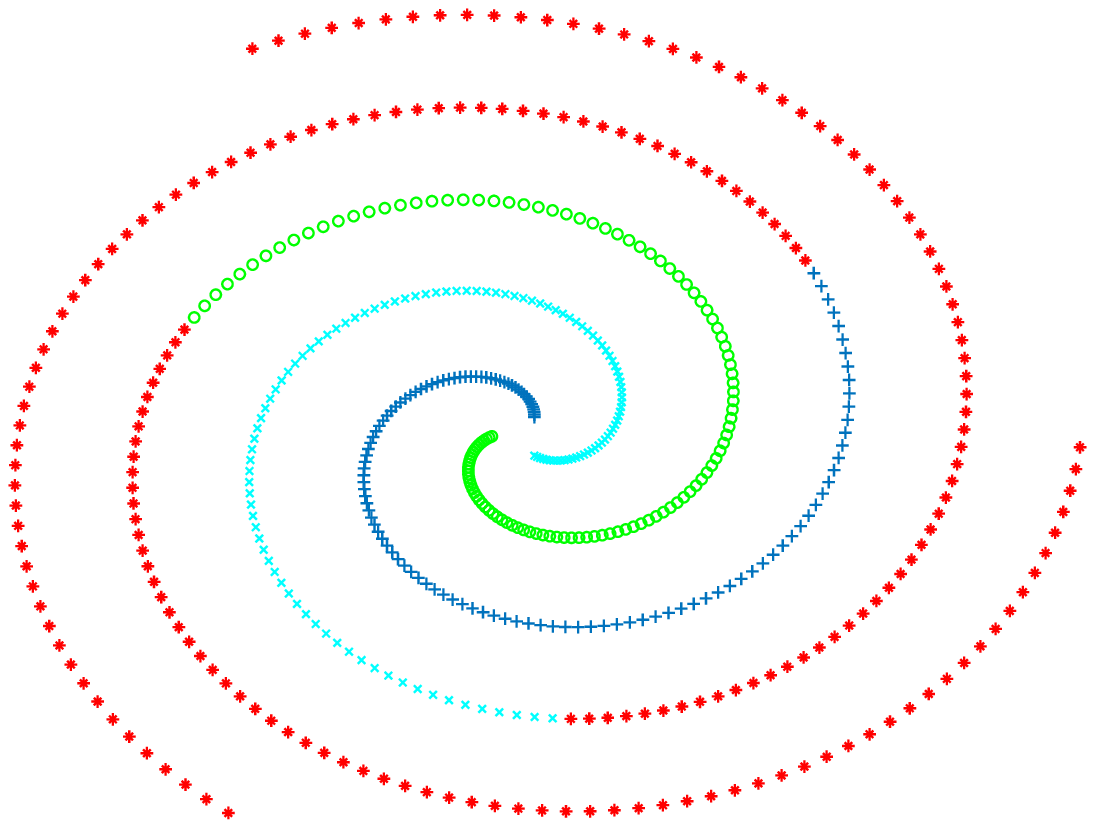}}
                                & \multicolumn{4}{c|}{\includegraphics[width=8cm,height=7cm]{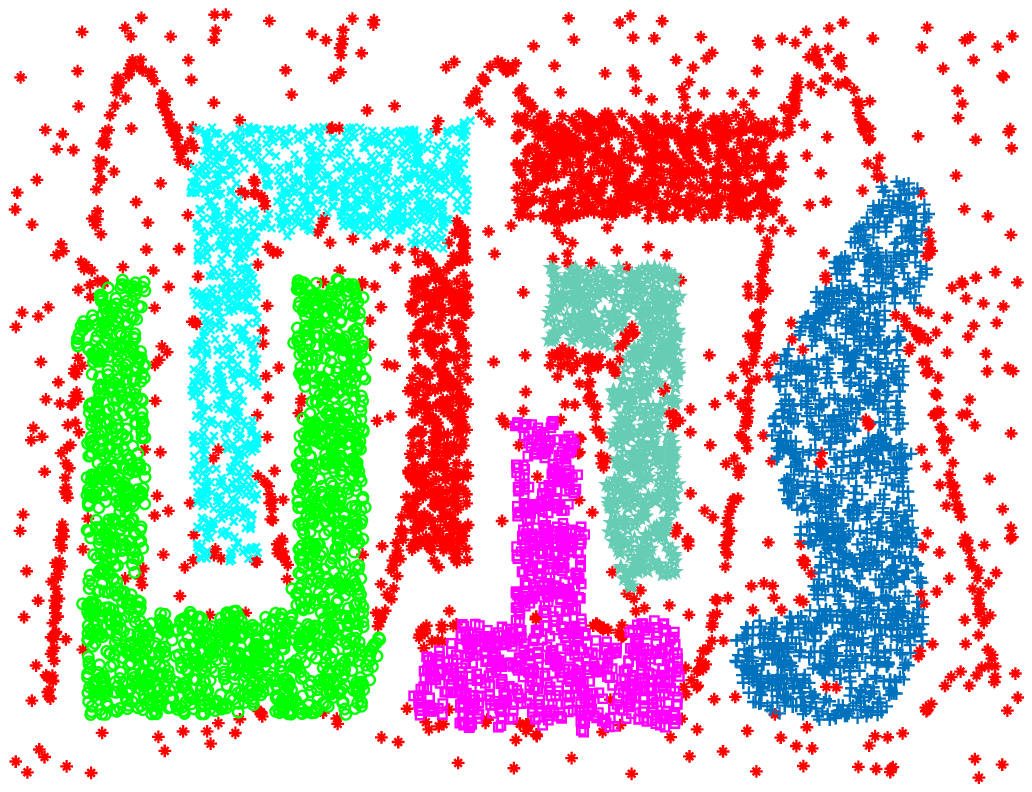}}
                                & \multicolumn{4}{c|}{\includegraphics[width=8cm,height=7cm]{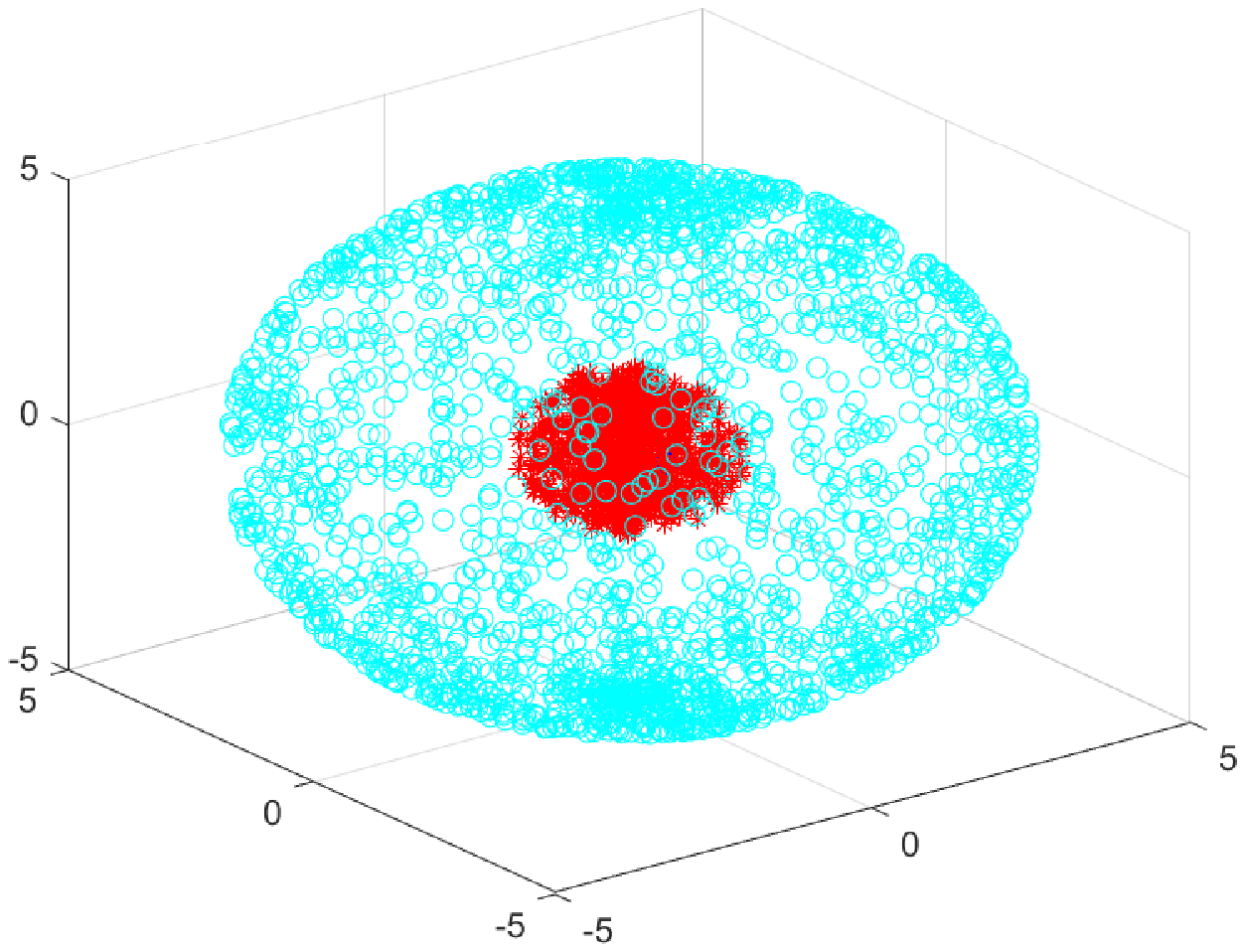}}
                                & \multicolumn{4}{c}{\includegraphics[width=8cm,height=7cm]{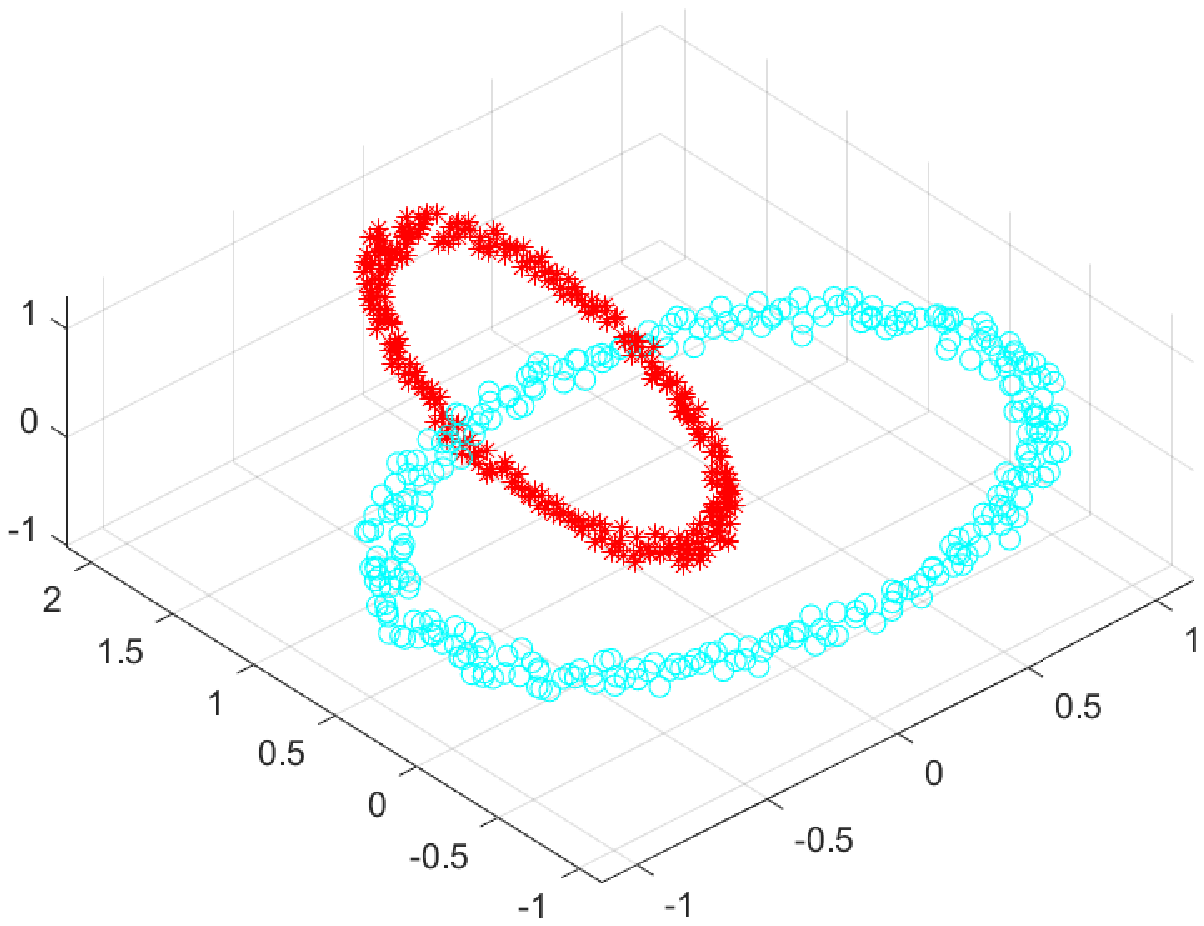}}              \\ \cline{2-17}
                                & RI     & ARI    & \multicolumn{1}{c|}{FS}      &RT      & RI     & ARI    & \multicolumn{1}{c|}{FS}    &RT
                                & RI     & ARI    & \multicolumn{1}{c|}{FS}      &RT      & RI     & ARI    & \multicolumn{1}{c|}{FS}    &RT     \\ \cline{2-17}
                                & 0.7491 & 1      & \multicolumn{1}{c|}{0.5832} &0.0492  & N/A    & N/A    & \multicolumn{1}{c|}{N/A}  &7.4800
                                & 0.9993 & 0.9993 & \multicolumn{1}{c|}{0.9994} & 2.2516 & 1      & 1      & \multicolumn{1}{c|}{1}    &0.0328     \\
                                \hline
\multirow{3}{*}{Spectral}       & \multicolumn{4}{c|}{\includegraphics[width=8cm,height=7cm]{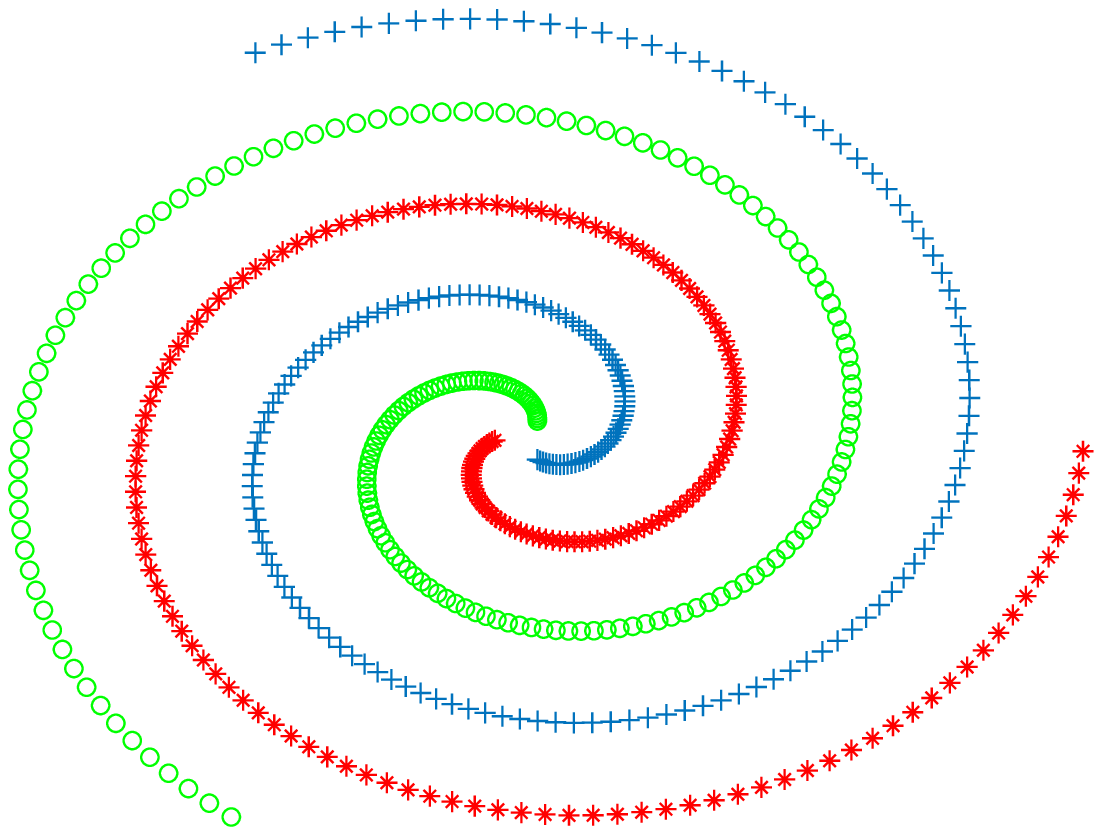}}
                                & \multicolumn{4}{c|}{\includegraphics[width=8cm,height=7cm]{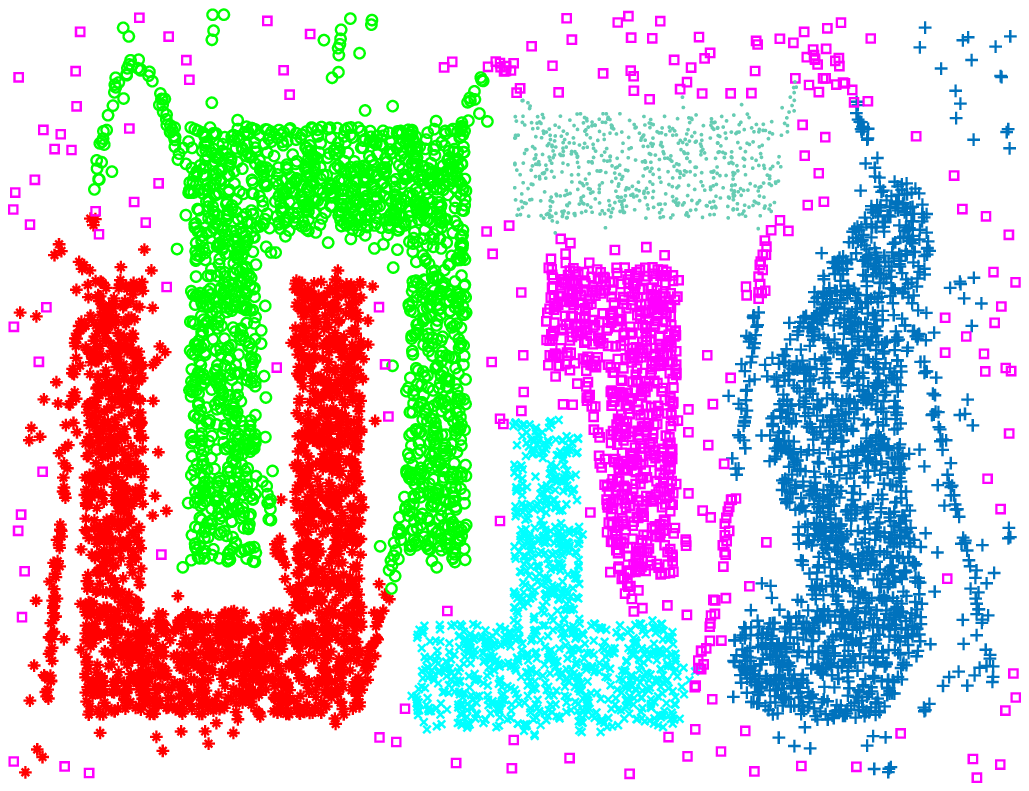}}
                                & \multicolumn{4}{c|}{\includegraphics[width=8cm,height=7cm]{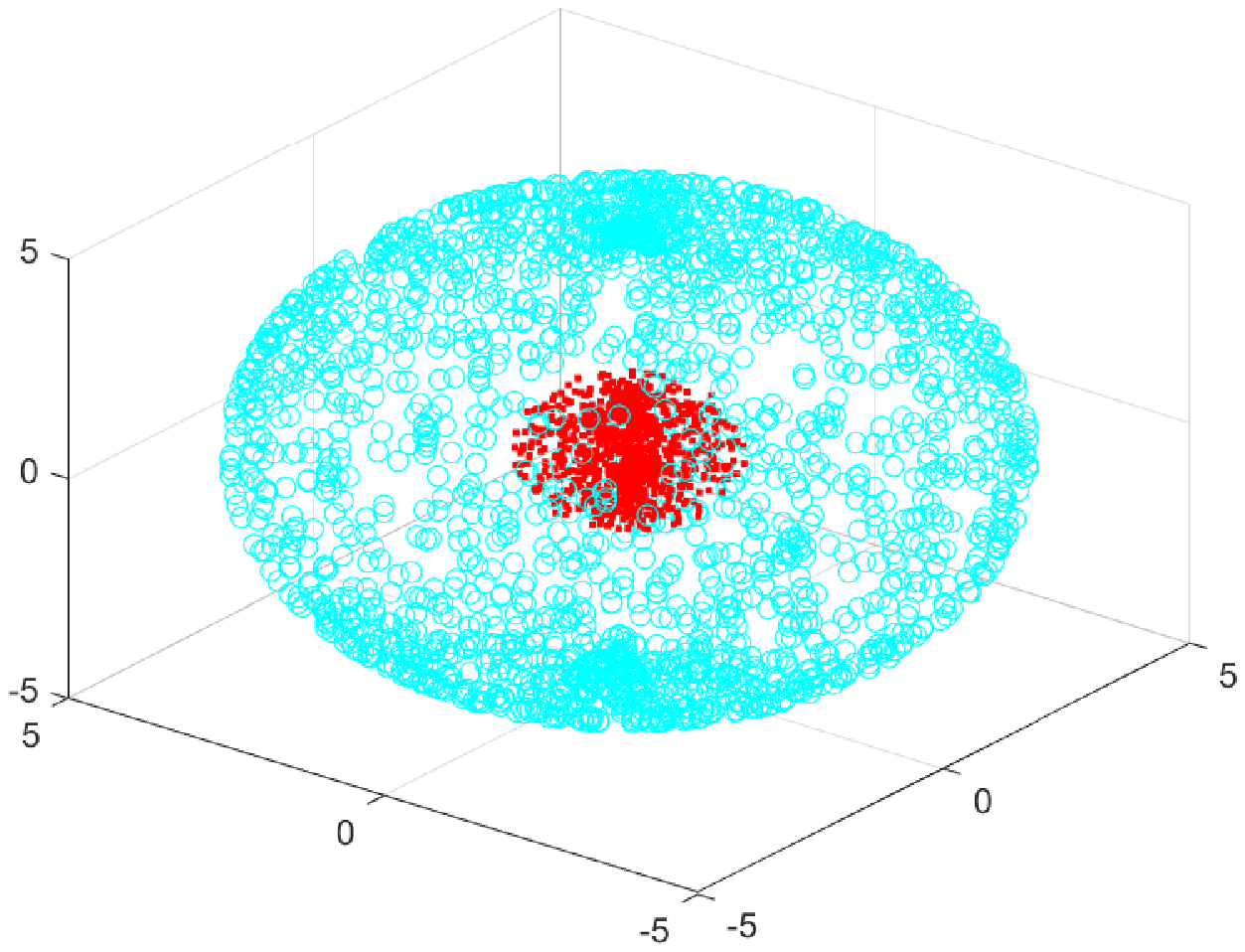}}
                                & \multicolumn{4}{c}{\includegraphics[width=8cm,height=7cm]{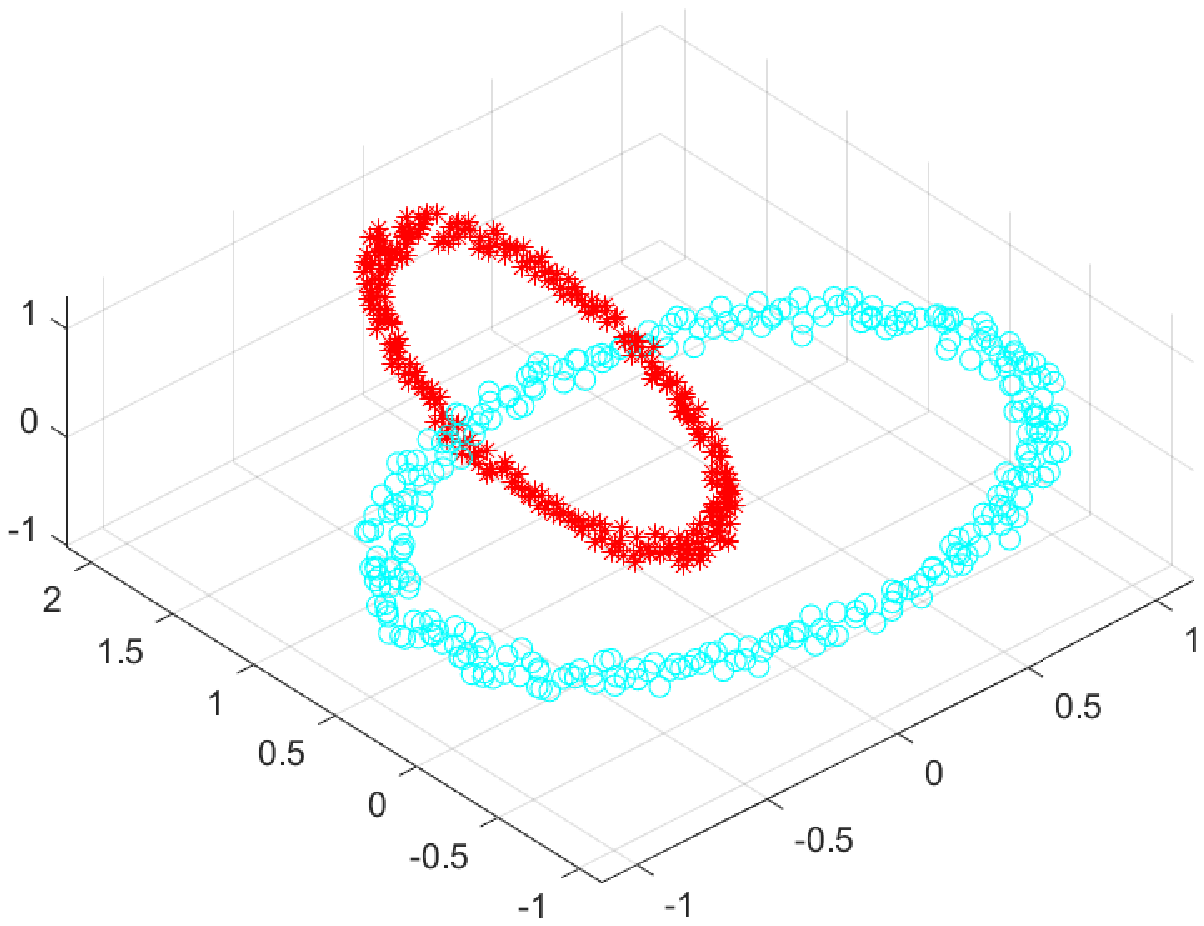}}              \\ \cline{2-17}
                                & RI     & ARI    & \multicolumn{1}{c|}{FS}       &RT     & RI      & ARI    & \multicolumn{1}{c|}{FS}      &RT
                                & RI     & ARI    & \multicolumn{1}{c|}{FS}       &RT     & RI      & ARI    & \multicolumn{1}{c|}{FS}      &RT       \\ \cline{2-17}
                                & 0.9866 & 0.9866 & \multicolumn{1}{c|}{0.9842}  &0.2734 & N/A     & N/A    & \multicolumn{1}{c|}{N/A}    &252.69
                                & 1      & 1      & \multicolumn{1}{c|}{1}       &13.310  & 0.9889 & 0.9889 & \multicolumn{1}{c|}{0.9876} &0.3148 \\ \hline
\multirow{3}{*}{GOPC}           & \multicolumn{4}{c|}{\includegraphics[width=8cm,height=7cm]{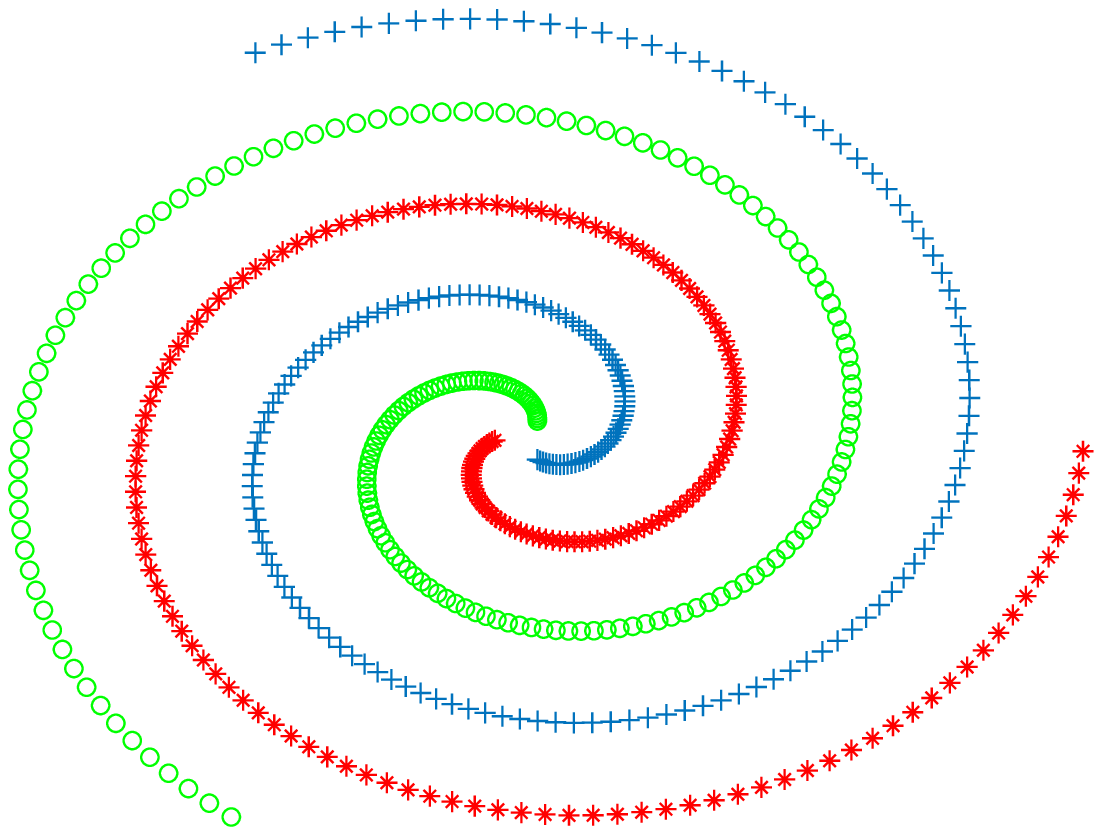}}
                                & \multicolumn{4}{c|}{\includegraphics[width=8cm,height=7cm]{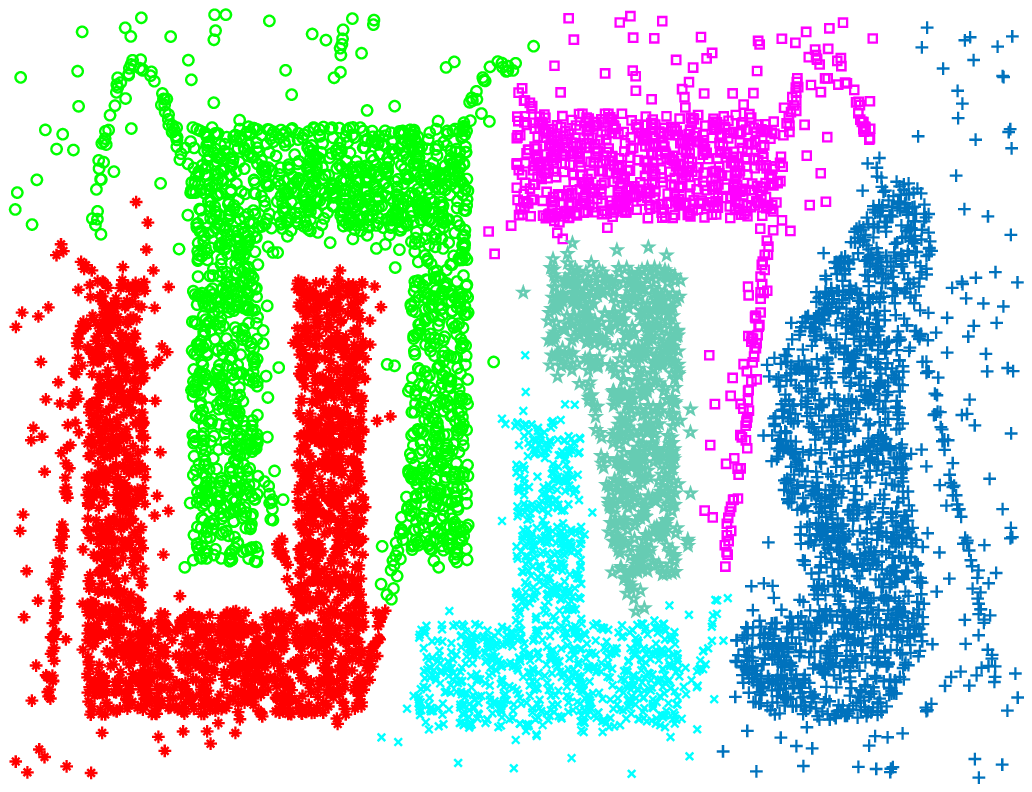}}
                                & \multicolumn{4}{c|}{\includegraphics[width=8cm,height=7cm]{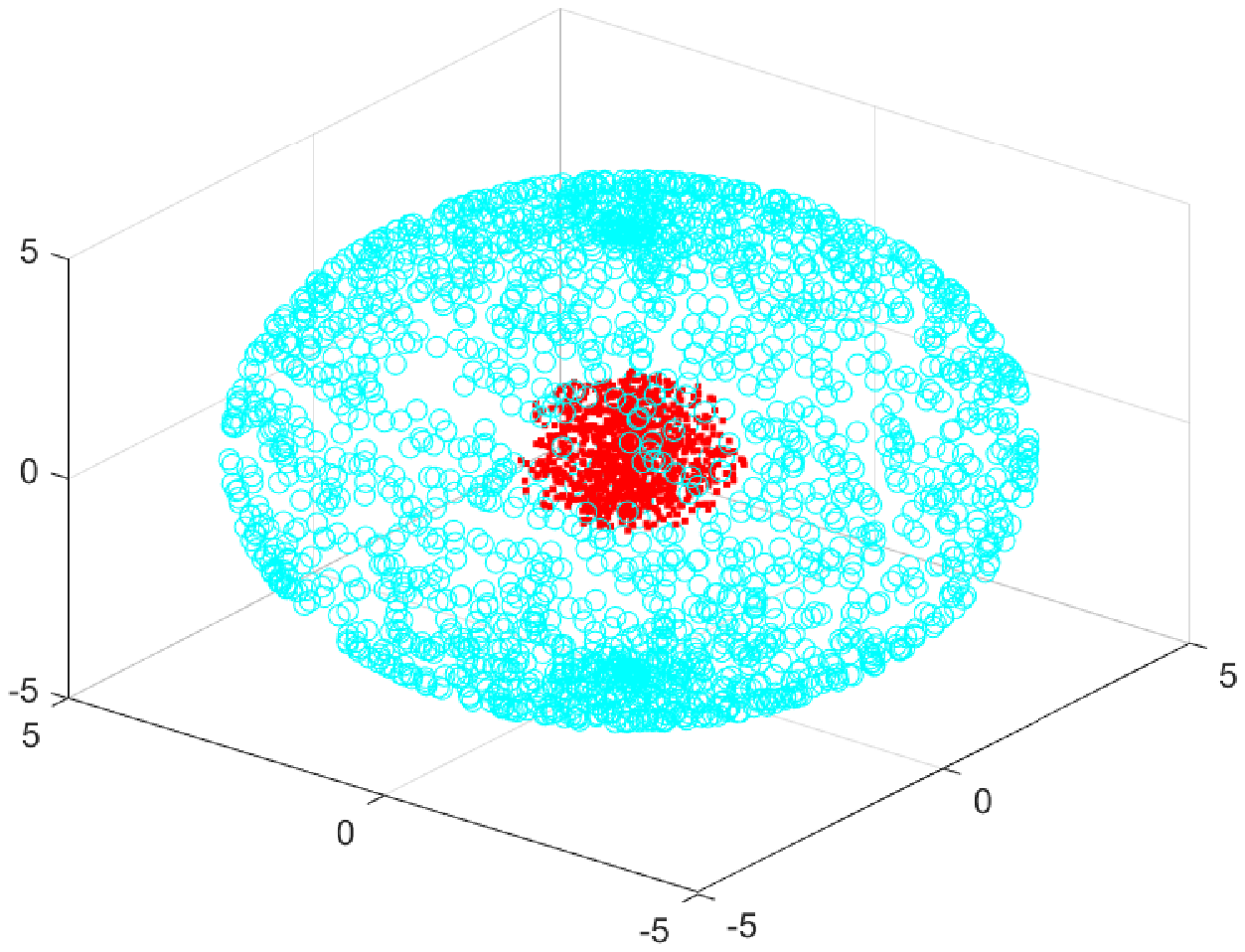}}
                                & \multicolumn{4}{c}{\includegraphics[width=8cm,height=7cm]{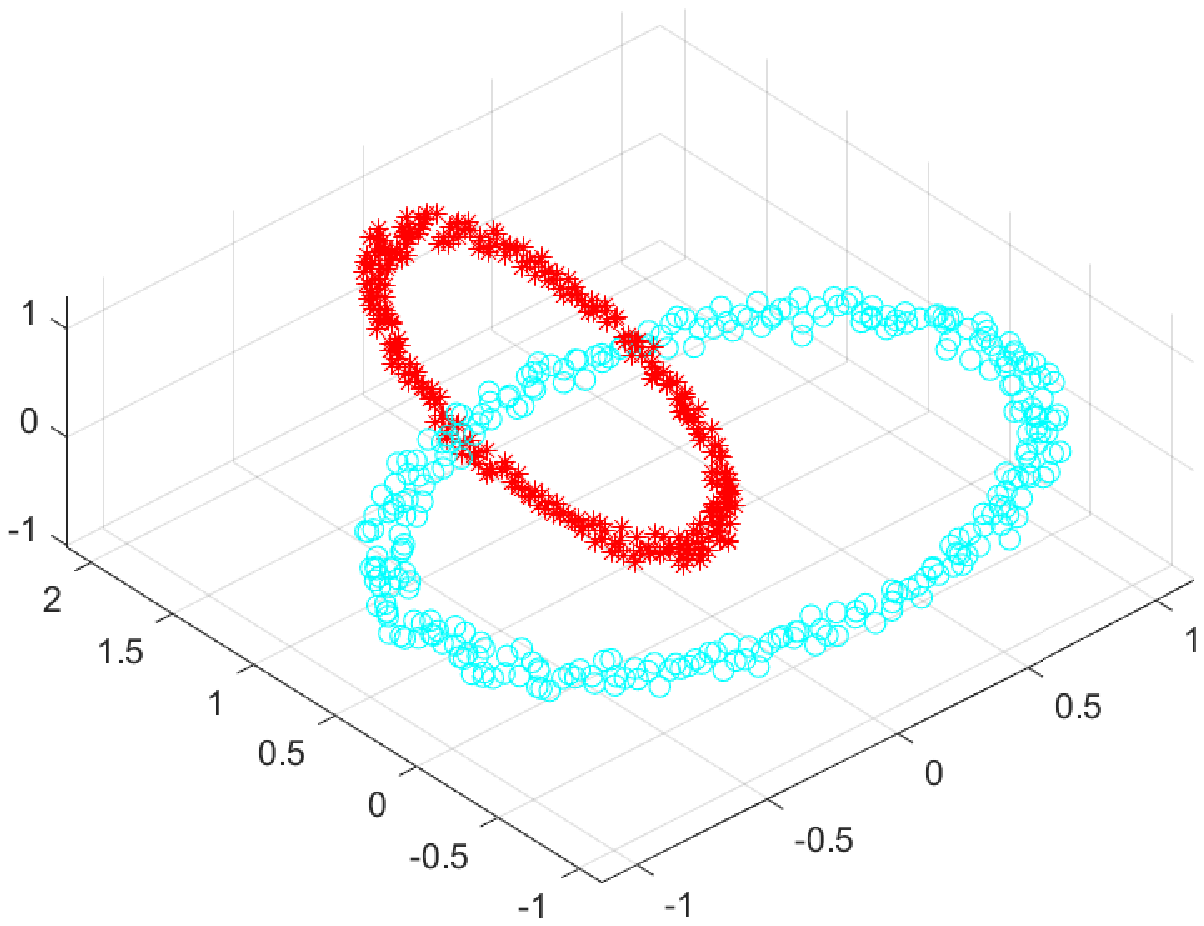}}
%  Fig below not found, Shuang
%  & \multicolumn{4}{c}{\includegraphics[width=8cm,height=7cm]{figure/experiment/GOPC_TwoRings3D.eps}}
\\ \cline{2-17}
                                & RI     & ARI    & \multicolumn{1}{c|}{FS}  &RT     & RI   & ARI  & \multicolumn{1}{c|}{FS}      &RT
                                & RI     & ARI    & \multicolumn{1}{c|}{FS}  &RT     & RI   & ARI  & \multicolumn{1}{c|}{FS}      &RT      \\ \cline{2-17}
                                & 1      & 1      & \multicolumn{1}{c|}{1}  &0.0555 & N/A  & N/A  & \multicolumn{1}{c|}{N/A}    &9.6797
                                & 1      & 1      & \multicolumn{1}{c|}{1}  &0.6398 & 1    & 1    & \multicolumn{1}{c|}{1}      &0.0539     \\ \hline
\multirow{3}{*}{CutPC}           & \multicolumn{4}{c|}{\includegraphics[width=8cm,height=7cm]{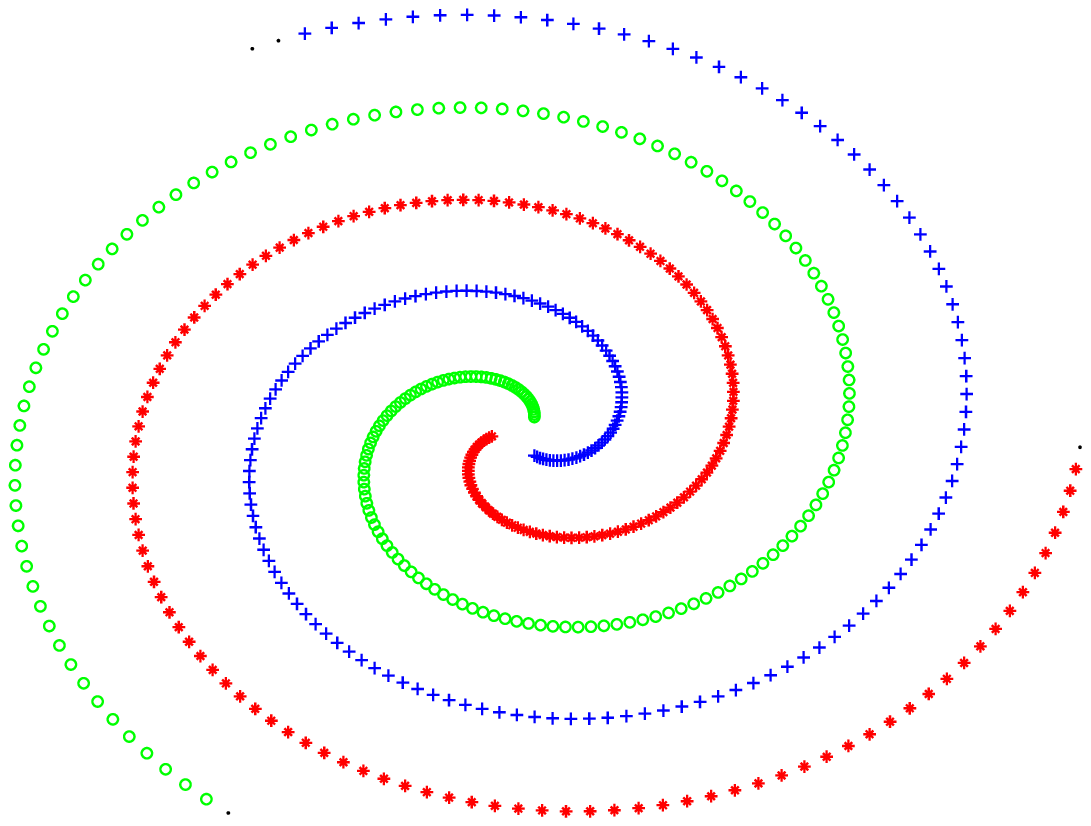}}
                                & \multicolumn{4}{c|}{\includegraphics[width=8cm,height=7cm]{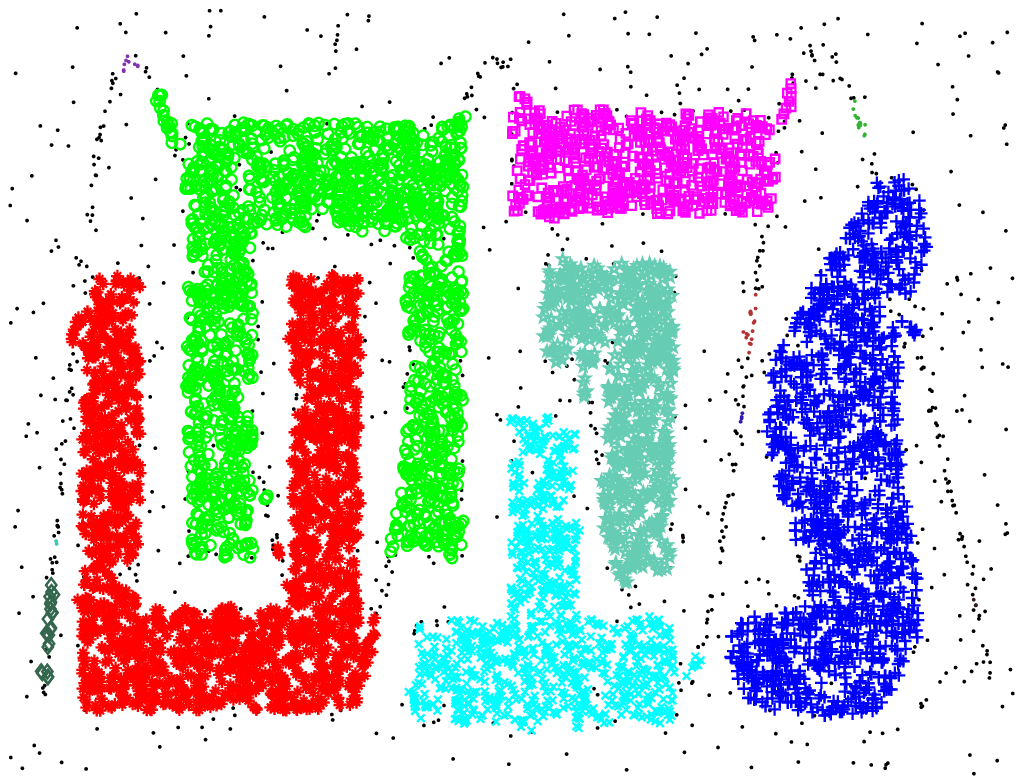}}
                                & \multicolumn{4}{c|}{\includegraphics[width=8cm,height=7cm]{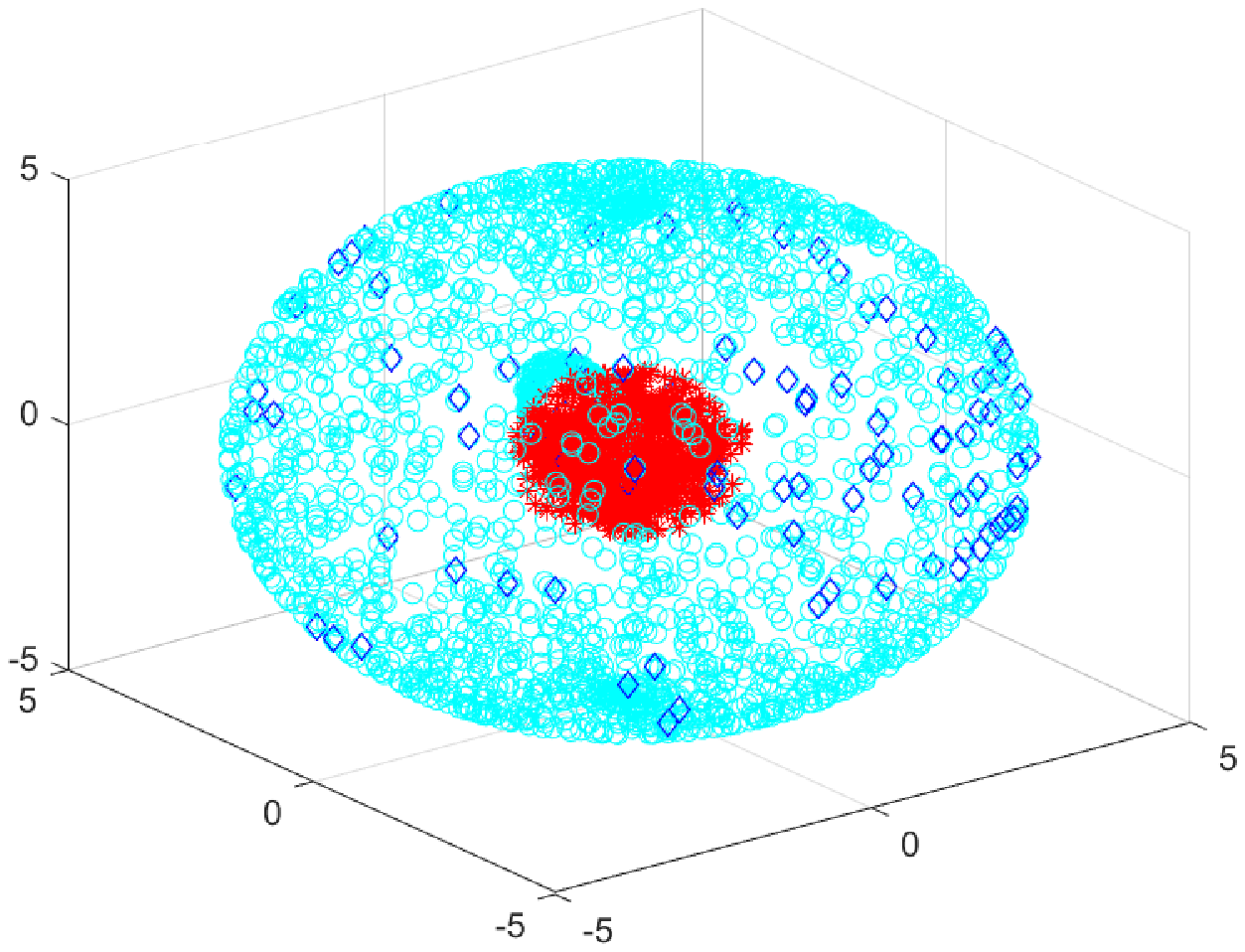}}
                                & \multicolumn{4}{c}{\includegraphics[width=8cm,height=7cm]{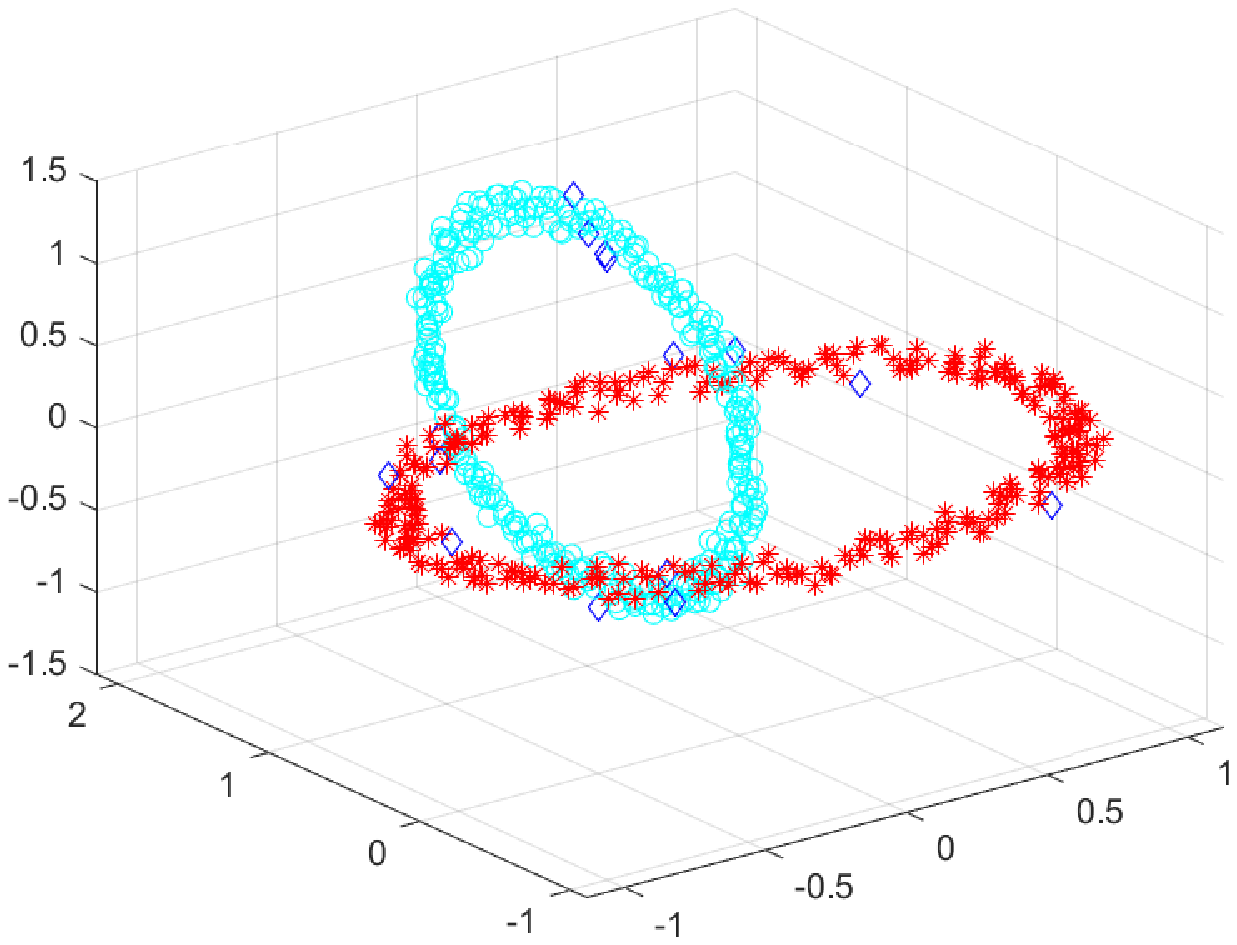}}
\\ \cline{2-17}
                                & RI     & ARI    & \multicolumn{1}{c|}{FS}  &RT     & RI   & ARI  & \multicolumn{1}{c|}{FS}      &RT
                                & RI     & ARI    & \multicolumn{1}{c|}{FS}  &RT     & RI   & ARI  & \multicolumn{1}{c|}{FS}      &RT      \\ \cline{2-17}
                                & 0.9953      & \textbf{1}      & \multicolumn{1}{c|}{0.9929}  & 0.0500 & N/A  & N/A  & \multicolumn{1}{c|}{N/A}    &9.6797
                                & 0.9624      & 1      & 0.9649  &1.0664 &  0.9754   & 1   & \multicolumn{1}{c|}{00.9747}      &0.0563    \\ \hline
\end{tabular}
}
\begin{tablenotes}
\tiny
\item[1] The symbol N/A in the table means the ground truth is not available.
\end{tablenotes}
\end{table}

% sec5.2
\subsection{Real world datasets}

The PaVa clustering algorithm is based on the dissimilarity of objects, so there is no doubt that it is operable as long as the adjacent matrix or dissimilarity matrix is given. Two real-world datasets are presented to demonstrate the generality of the PaVa clustering algorithm.

\begin{figure}[H]
  \centering
  \subcaptionbox{}{\includegraphics[width=7.1cm,height=6.4cm]{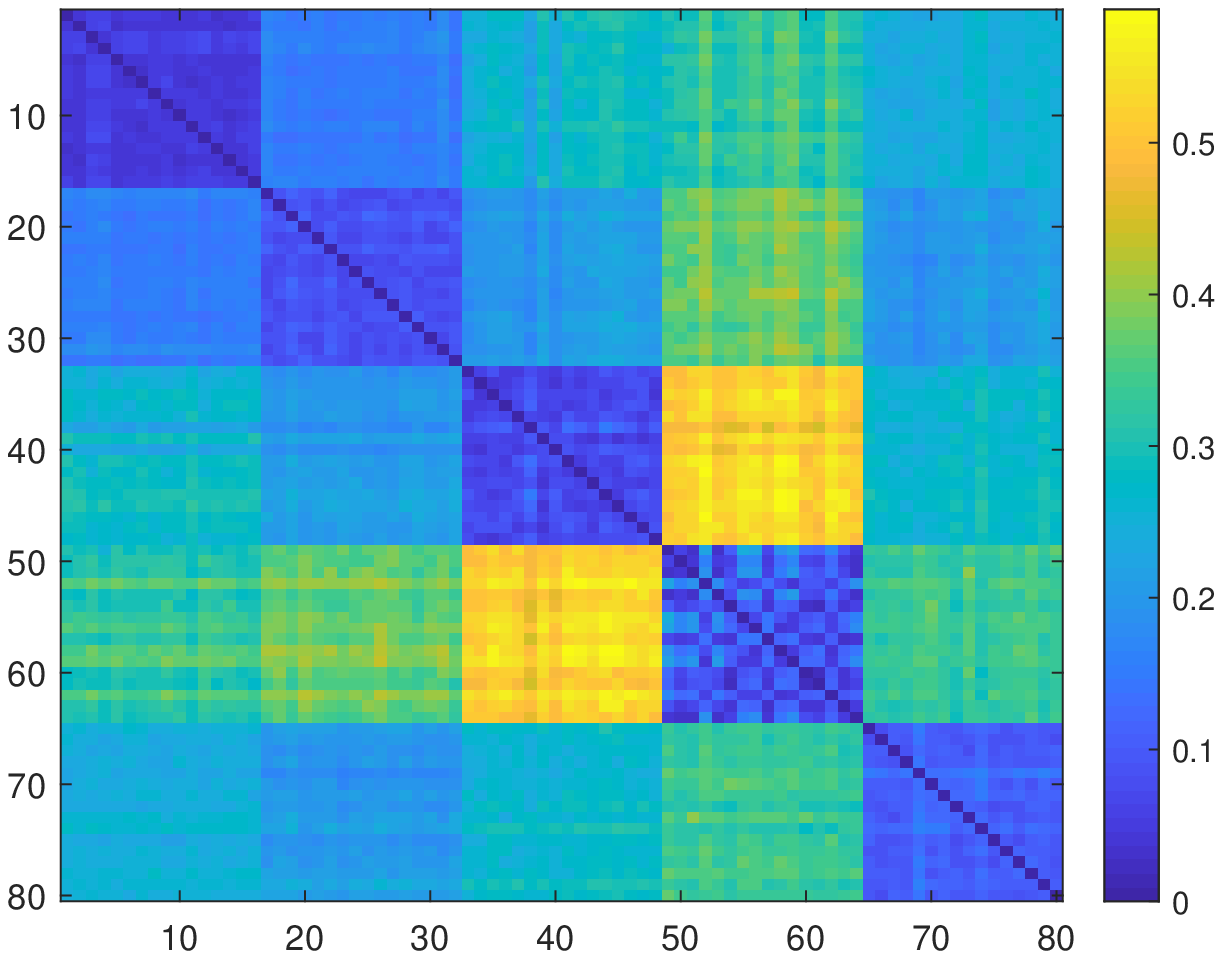}}
  \subcaptionbox{}{\includegraphics[width=6.4cm,height=6.4cm]{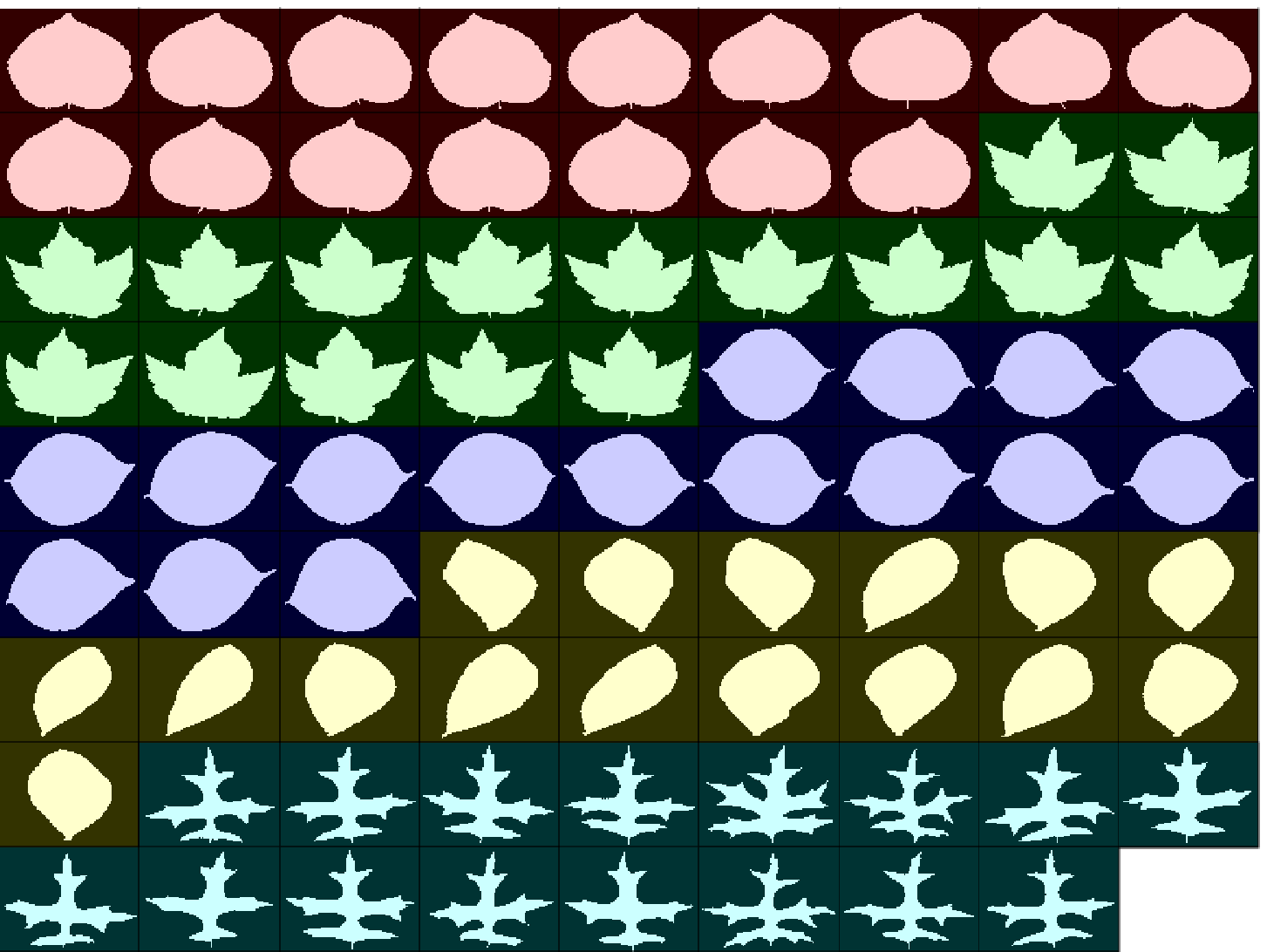}}
  \caption{\small{Results on dataset leaf: (a) Adjacent matrix. (b) Clustering result. The adjacent matrix shows the dissimilarity (shape context distance) of any pair of objects. After clustering, it takes the form of block diagonalization, which implies small intra-cluster dissimilarity and large inter-cluster dissimilarity.}}
  \label{figure:leaf}
\end{figure}

On the one hand, the dataset leaf presents the outline of five different leaves from \cite{mallah2013plant}. Hence, the shape context is used to measure the similarity of these outlines. The adjacent matrix is shown in Fig.\ref{figure:leaf}.(a). The result is shown in Fig.\ref{figure:leaf}.(b). The PaVa algorithm performs well in this balanced and well-separated dataset.

\begin{figure}[H]
  \centering
  \subcaptionbox{}{\includegraphics[width=6.8cm,height=6.4cm]{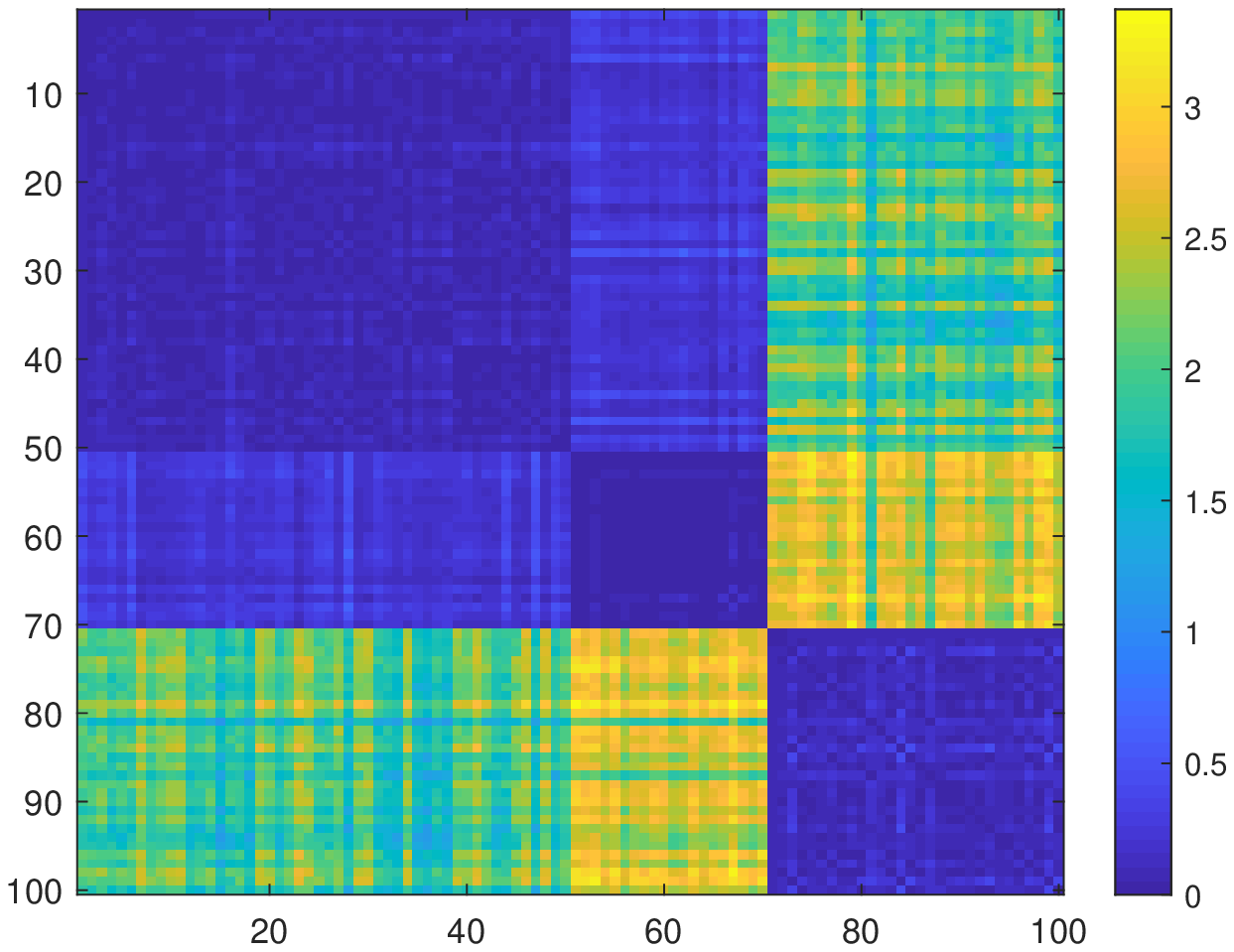}}
  \subcaptionbox{}{\includegraphics[width=6.8cm,height=6.4cm]{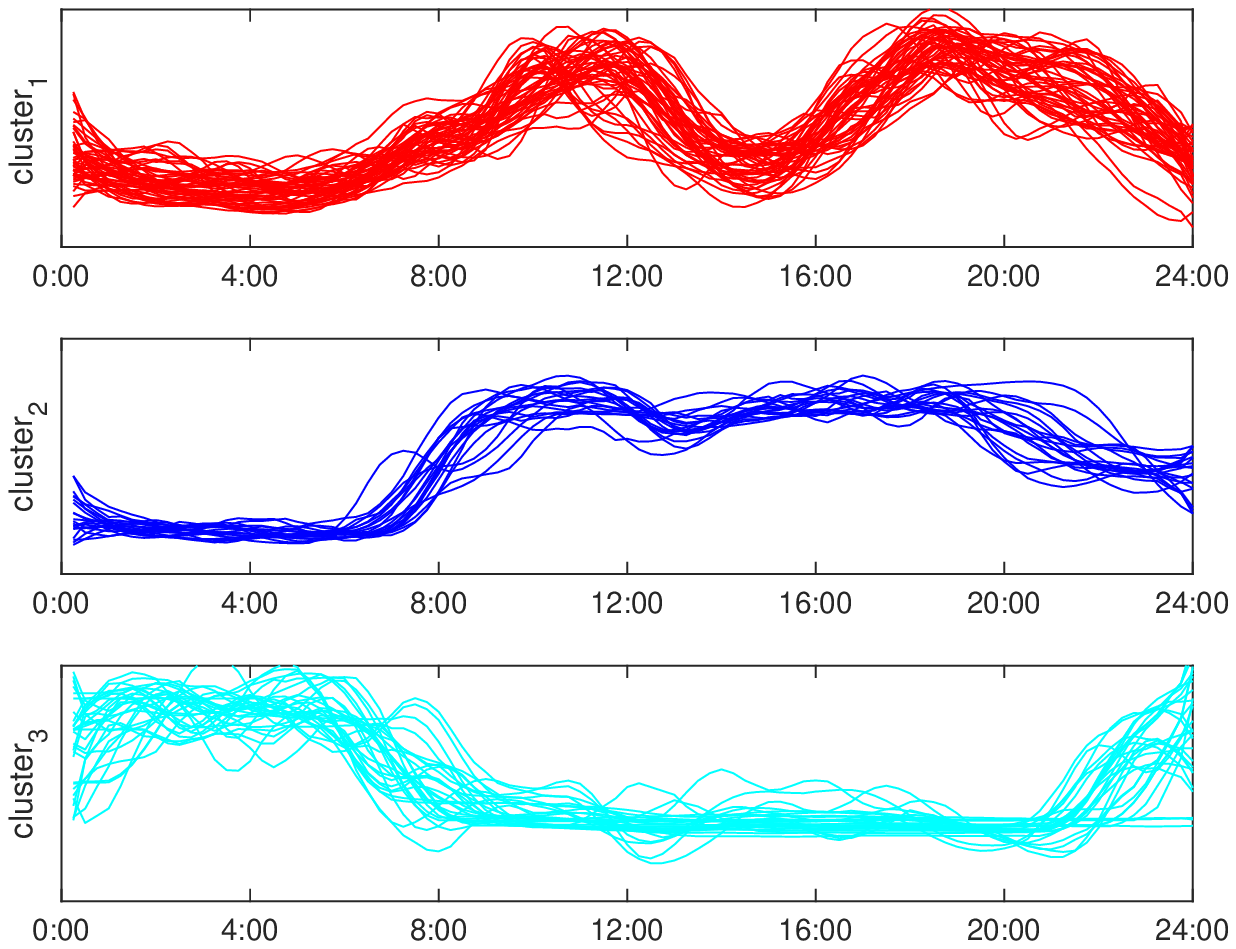}}
  \caption{\small{Results on dataset electric load: (a) Adjacent matrix. (b) Clustering result. The adjacency matrix takes the form of a heterogeneous block diagonalization, meaning that objects in different clusters have different numbers and varieties.}}
  \label{figure:electric_load}
\end{figure}

On the other hand, the dataset electric load is gathered from the smart meters in one industrial park gathered in \cite{li2022density}. After the electric load is normalized and smoothed to curve, the Dynamic Time Warp (DTW) with restriction is used to calculate the pairwise distances between any two curves so that the chronological order is carefully considered. The adjacent matrix is shown in Fig.\ref{figure:electric_load}.(a), and the result is shown in Fig.\ref{figure:electric_load}.(b). Compared with the dataset `leaf', the dataset electric load is unbalanced and heterogenous; the PaVa clustering algorithm still performs well on such a complex dataset. $cluster_1$  presents the load features of the service industry, such as restaurants or snack bars in this industrial park. There are two peaks in this group of load curves, corresponding with lunchtime and dinner time, respectively. Moreover, $cluster_2$ shows the features of the low energy consumption industry, which has regular working hours in the daytime and no activity at night. On the contrary, the load is low in the daytime and high at night in $cluster_3$; these load curves belong to the high energy consumption industry, which is more sensitive to the price of electricity. Therefore, the features of their load curves verify the effect of Time-of-Use.

In conclusion, the PaVa clustering algorithm can be applied to different domains with good results.

\subsection{Performance under different parameters with synthetic data}
The only parameter in the PaVa clustering algorithm is $k$ for calculating $k$-distance.

\begin{figure}[H]
  \centering
  \subcaptionbox{}{\includegraphics[width=6cm]{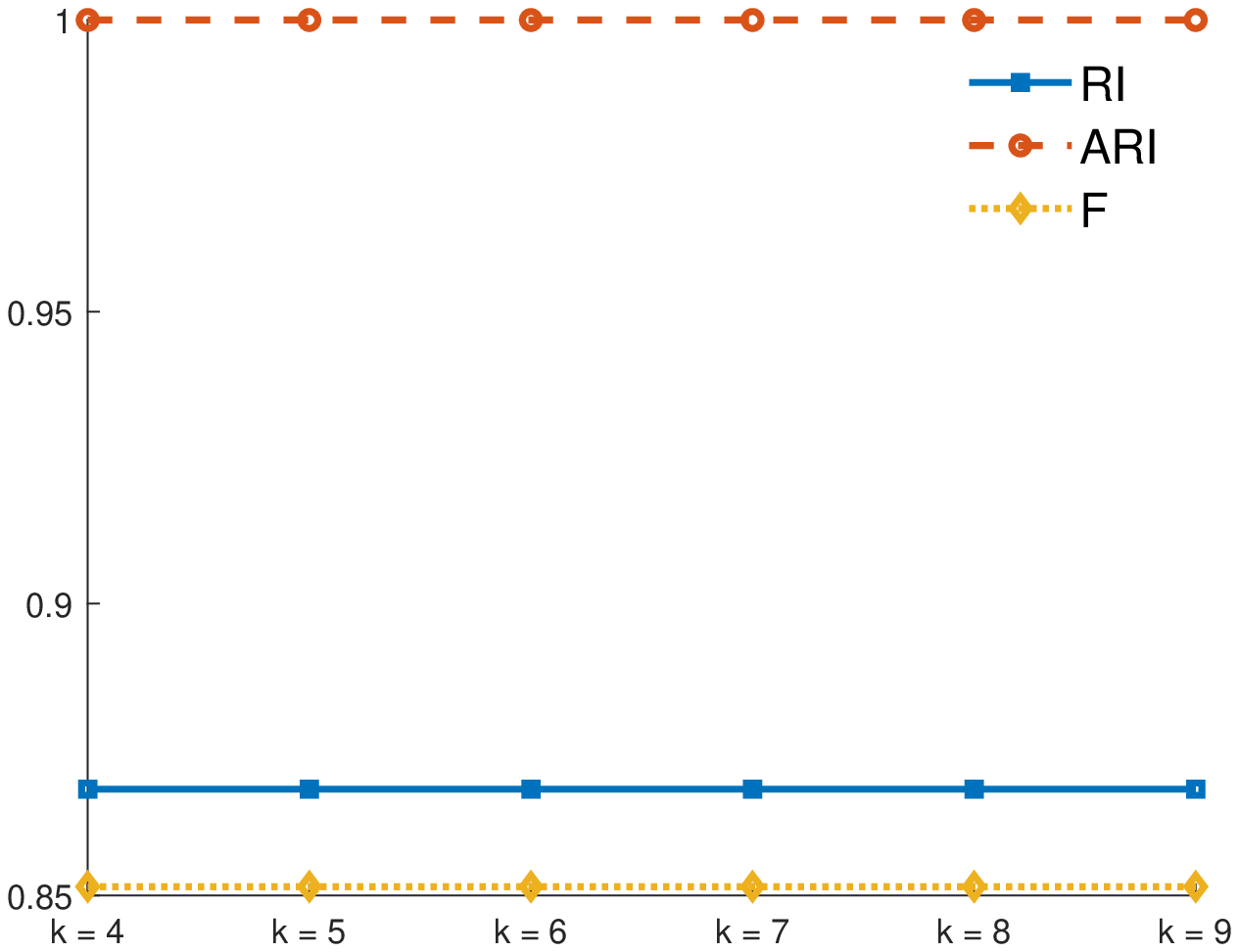}}
  \subcaptionbox{}{\includegraphics[width=6cm]{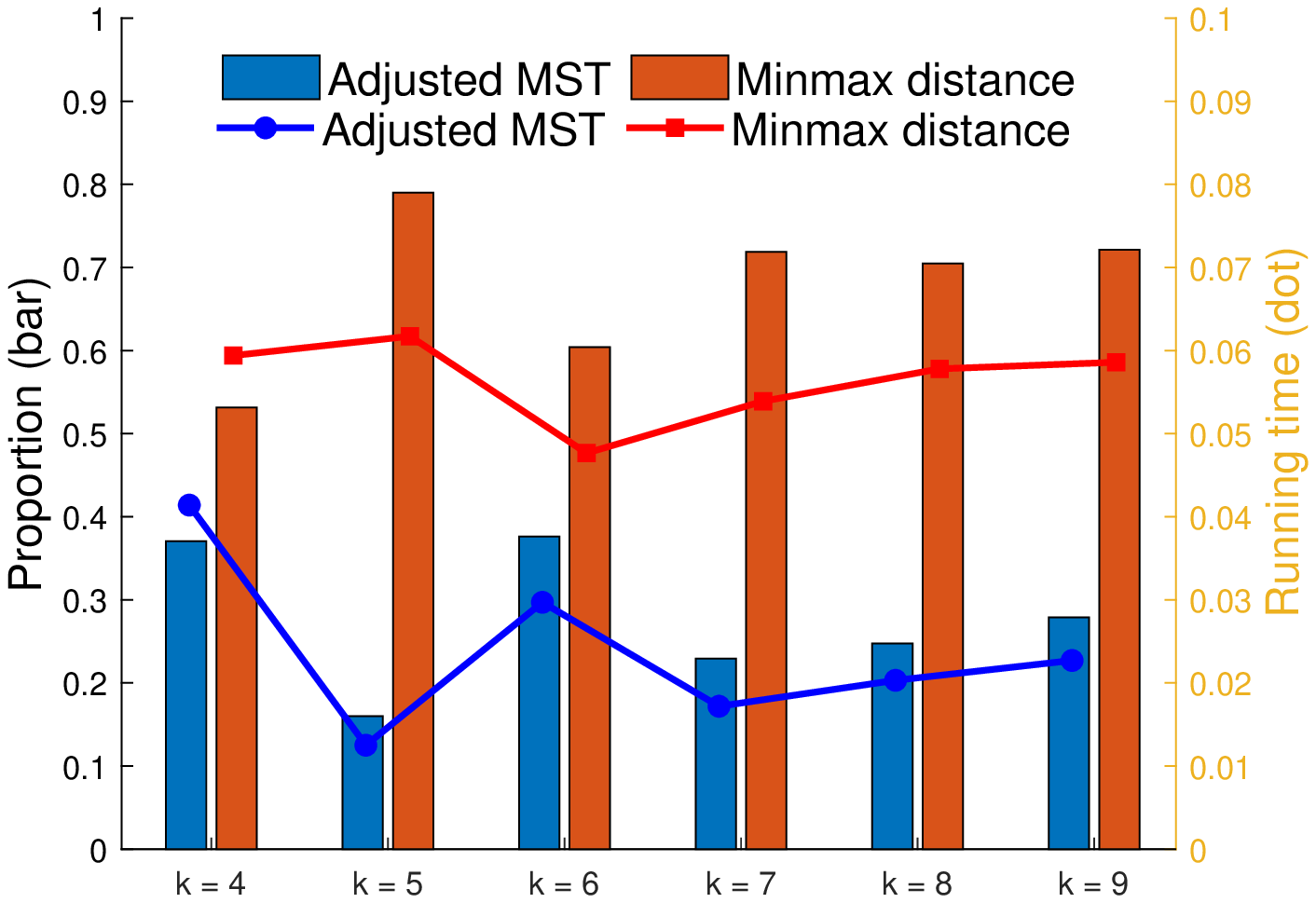}}

  \subcaptionbox{}{\includegraphics[width=6cm]{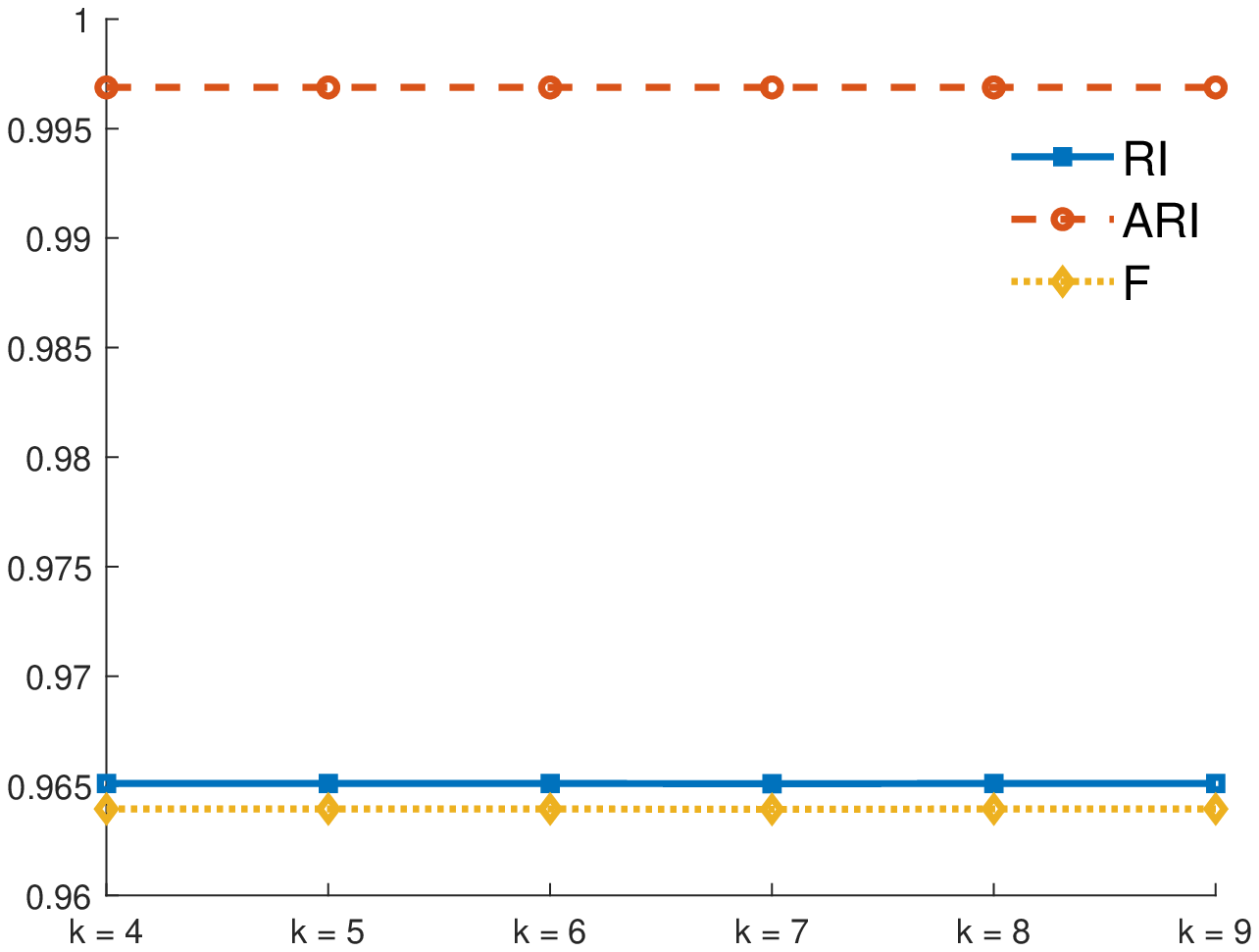}}
  \subcaptionbox{}{\includegraphics[width=6cm]{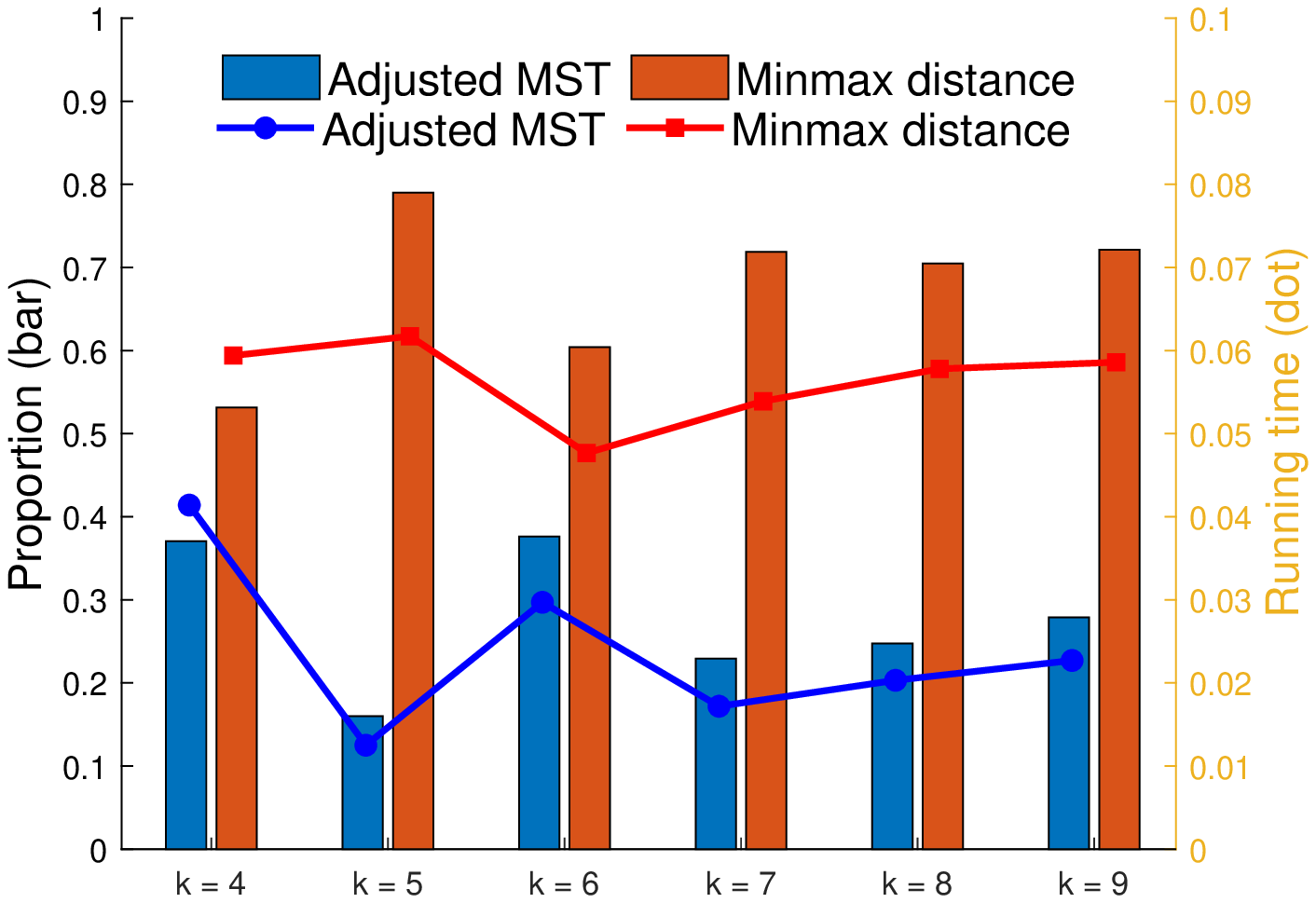}}
  \caption{\small{(a) Accuracy under different values of $k$ on dataset \emph{twomoons\_noise}. (b) Running times and proportions of two main modules under different values of parameters $k$ on dataset \emph{twomoons\_noise}.  (c) Accuracy under different values of $k$ on dataset \emph{twomoons\_bridge}. (d) Running times and proportions of two main modules under different values of parameters $k$ on dataset \emph{twomoons\_bridge}.}}
  \label{figure:paremater_effect}
\end{figure}

As mentioned before, the $k$-distance is a measurement for density and represents how the objects are scattered in an area. $k$ impacts the selection of generalized centers and adjustment of MST via the density measurement $k$-distance. Regardless of whether the value of $k$ is too large or too small, it will reduce the accuracy of density measurement. Meanwhile, a large $k$ leads to high computational costs. As a rule of thumb, $\lceil \log N \rceil $ is chosen for $k$.  To analyze the performance of the PaVa clustering algorithm under different values of $k$, we set $k$ to be different integers around $\log N$ on the mentioned dataset \emph{twomoons\_noise} and \emph{twomoons\_bridge}. Accuracy remains unchanged when $k$ varies within the specified range because the distribution of minmax distances is stable as long as the chosen centers are located at the main body of clusters, and the adjusted MST performs well under different types of noise, see Fig.\ref{figure:paremater_effect}.(a)(c).

Two modules occupy most of the running time, the construction of the adjusted MST, and the calculation of minmax distances between the center object and all the objects by the adjusted MST. The running times and proportions are shown in Fig.\ref{figure:paremater_effect}.(b)(d). Once constructed, the adapted MST can be used in all the following steps. By contrast, the calculation of minmax distances is run for $M$ times, where $M$ is the number of clusters. Thus, $k$ is only used in the first module, so its impact on running time is limited.

In conclusion, the PaVa clustering algorithm is robust to parameter $k$ from the perspective of accuracy and efficiency.

\section{Conclusion}\label{section_conclusion}

In this paper, we propose a novel Path-based Valley-seeking clustering algorithm which generalizes the clustering problem on spherical datasets to clustering problems on non-spherical datasets by replacing the Euclidean distance with minmax distance. Therefore, the valley among clusters is transformed from irregular boundaries to spherical shells. Moreover, adjusted MST is established to calculate a more proper minmax distance to enhance the stability under different types of noise. The PaVa clustering algorithm is applied to many different kinds of scenarios, including datasets composed of spherical clusters, arbitrarily shaped clusters, and two real-world datasets. The PaVa clustering algorithm performs well on all of the datasets regardless of whether they are homogeneous or heterogenous thanks to the data-driven approach which ascertains the centers and radii of spherical shells wrapping clusters. It also achieves good results on datasets, both in the presence of different types of noise and in the absence of noise.  The experiments show the generality of our algorithm and we believe that it can be applied to a wider range of scenarios.

% making the extraction process intuitive and efficient. Moreover, the PDF of distances determines the radii of shells, which ensures its ability to deal with heterogeneous clusters. As for the framework, the clusters are extracted one by one, which is user-friendly due to the absence of the number of clusters. Additionally, the adjusted MST is used to enhance its robustness to noise. Finally, the experiments show that the proposed method achieves good results on both synthetic datasets and real-world datasets from different domains.

% Even if determining radii of spherical shells by PDF of minmax distances are flexible in dealing with heterogeneous clusters, the discrete case for minmax distance would likely impact the accuracy of radii. Therefore, we look forward to finding a method to make minmax distance continuous. In addition, we would like to embed a hierarchical structure into our algorithm to approximate minmax distance so that it can be applied to datasets with huge volumes and more complex scenarios.

\section*{References}

%\bibliography{mybibfile}

\end{document}